\RequirePackage[2020-02-02]{latexrelease}

\documentclass{nlearxiv}
\usepackage{amsmath}
\usepackage{natbib}
\usepackage{enumitem}
\usepackage{array}
\usepackage{multirow}
\usepackage{tikz}
\usepackage{tikz-qtree}
\usepackage{varwidth}
\usepackage{microtype}
\usepackage{tabularx}
\usepackage[hidelinks]{hyperref}
\usepackage[perpage]{footmisc}
\usepackage{parskip}
\newcommand{\mrrt}[2]{\multirow{#1}{5mm}{\rotatebox{90}{#2}}}

\begin{document}
\label{firstpage}

\lefttitle{Natural Language Engineering}
\righttitle{E. Troiano \textit{et al}.}

\papertitle{Survey paper}

\jnlPage{1}{60}
\jnlDoiYr{2022}
\doival{This is a preprint. The published version can be found at
  \href{https://doi.org/10.1017/S1351324922000407}{doi:10.1017/S1351324922000407}. Please
only cite the published version.}

\title{From Theories on Styles to their Transfer in Text:\\Bridging the Gap with a Hierarchical Survey}

\begin{authgrp}
\author{Enrica Troiano$^{1,\dagger}$, Aswathy Velutharambath$^{1,2}$, and Roman Klinger$^{1,*}$}
\affiliation{$^1$Institut für Maschinelle Sprachverarbeitung, University
  of Stuttgart, Germany and 
  $^2$100 Worte Sprachanalyse GmbH, Heilbronn, Germany\\
  $*$ Corresponding author. E-mail: roman.klinger@ims.uni-stuttgart.de
}
\end{authgrp}

\begin{abstract}
  Humans are naturally endowed with the ability to write in a
  particular style.  They can, for instance, re-phrase a formal letter
  in an informal way, convey a literal message with the use of figures
  of speech or edit a novel by mimicking the style of some well-known
  authors.  Automating this form of creativity constitutes the goal of
  style transfer.  As a natural language generation task, style
  transfer aims at rewriting existing texts, and specifically, it
  creates paraphrases that exhibit some desired stylistic
  attributes. From a practical perspective, it envisions beneficial
  applications, like chatbots that modulate their communicative style
  to appear empathetic, or systems that automatically simplify
  technical articles for a non-expert audience.

  Several style-aware paraphrasing methods have attempted to tackle
  style transfer. A handful of surveys give a methodological overview
  of the field, but they do not support researchers to focus on
  specific styles.  With this paper, we aim at providing a
  comprehensive discussion of the styles that have received attention
  in the transfer task. We organize them in a hierarchy, highlighting
  the challenges for the definition of each of them, and pointing out
  gaps in the current research landscape. The hierarchy comprises two
  main groups. One encompasses styles that people modulate
  arbitrarily, along the lines of registers and genres. The other
  group corresponds to unintentionally expressed styles, due to an
  author's personal characteristics. Hence, our review shows how these
  groups relate to one another, and where specific styles, including
  some that have not yet been explored, belong in the hierarchy.
  Moreover, we summarize the methods employed for different stylistic
  families, hinting researchers towards those that would be the most
  fitting for future research.
\end{abstract}


\maketitle
\footnotetext{$^\dagger$ Enrica Troiano and Aswathy Velutharambath contributed equally.}

\section{Introduction}
\label{sec:introduction}

Communication comes in a style. Be it in language, visual arts or
music, the things that people express have a \textit{content} -- what
is to be conveyed, and a \textit{style} -- how that is done.  These
two concepts are evident in the Shakespearean verses ``\textit{By the
  pricking of my thumbs, Something wicked this way comes}'' (Macbeth,
Act 4, Scene 1.), where the content (i.e., the foreseeing of an evil
future) is encoded in the slant rhyme with peculiar rhythm and unusual
vocabulary choices.  Style is thus the form given to a core piece of
information, which collocates it into some distinctive communicative
categories. For instance, we perceive that the above example is a
poem, and specifically, one written in an old variety of English.

The binomial of content and style is interesting from a computational
perspective because content can be styled in a controlled manner. By
considering these two variables, many studies have dealt with the
automatic generation of texts \citep{gatt2018survey}, images
\citep{wu2017survey} and music \citep{briot2020deep} that display a
number of desired features. Works as such create content from scratch
and combine it with style, while a kin line of research transforms
styles starting from an already existing piece of content. The
rationale is: if style and content are two and separate, one can be
modified and the other kept unaltered.  This practice is pervasive
among humans as well. It can be observed, for instance, any time they
give an inventive twist to their utterances and creations (e.g., when
conveying a literal gist through a metaphor, or when painting by
imitating Van Gogh's singular brush strokes).  The field of vision has
achieved remarkable success in changing the styles of images
\citep{gatys2016image}, and following its footsteps, natural language
processing (NLP) has risen to the challenge of style transfer in text.

\subsection{Style transfer in text: task definition} The goal of
textual style transfer is to modify the style of texts while
maintaining their initial content (i.e., their main meaning).  More
precisely, style transfer requires the learning of $p(t'\mid s,t)$: a
text $t'$ has to be produced given the input $t$ and a desired
stylistic attribute $s$, where $s$ indicates either the presence or
the absence of such an attribute\footnote{We call ``attribute'' the
  value (e.g., presence, absence, degree) that a specific style (e.g.,
  formality) can take.}  with respect to $t$. For example, if $t$ is
written in a formal language, like the sentence ``\textit{Please, let
  us know of your needs}'', then $s$ may represent the opposite (i.e.,
\textit{in}formality), thus requiring $t'$ to shift towards a more
casual tone, such as ``\textit{What do you want?}''.  Therefore, style
transfer represents an effort towards conditioned language generation
and yet differs from this broader task fundamentally. While the latter
creates text and imposes constraints over its stylistic
characteristics alone, the style transfer constraints relate to both
style, which has to be different between input and output, and
content, which has to be similar between the two -- for some
definition of ``similar''.  In short, a successful style transfer
output checks three criteria. It should exhibit a different stylistic
attribute than the source text $t$, it needs to preserve its content,
and it has to read as a human production
\citep{mir-etal-2019-evaluating}.

\subsection{Applications and challenges} Style transfer lends itself
well for several applications.  For one thing, it supports automatic
linguistic creativity, which has a practical entertainment value.
Moreover, since it simulates humans' ability to switch between
different communicative styles, it can enable dialogue agents to
customize their textual responses for the users, and to pick the one
that is appropriate in the given situation
\citep{gao-etal-2019-structuring}. Systems capable of style transfer
could also improve the readability of texts by paraphrasing them in
simpler terms \citep{cao-etal-2020-expertise}, and help in this way
non-native speakers \citep{wang-etal-2019-harnessing}.

The transfer in text has been tackled with multiple styles (e.g.,
formality and sentiment) and different attributes thereof (e.g.,
formal vs.\ informal, sentiment gradations).  Nevertheless, advances
in these directions are currently hampered by a lack of appropriate
data.  Learning the task on human-written linguistic variations would
be ideal, but writers hardly produce parallel texts with similar
content and diverse attributes. If available, resources of this sort
might be unusable due to the mismatch between the vocabularies of the
source and target sides \citep{pang-2019-towards}, and constructing
them requires expensive annotation efforts
\citep{gong-etal-2019-reinforcement}.

The goal of style transfer seems particularly arduous to achieve per
se. Most of the time, meaning preservation comes at the cost of only
minimal changes in style \citep{wu-etal-2019-hierarchical-reinforced},
and bold stylistic shifts tend to sacrifice the readability of the
output \citep*{helbig-etal-2020-challenges}.  This problem is
exacerbated by a lack of standardized evaluation protocols, which
makes the adopted methods difficult to compare. In addition, automatic
metrics to assess \textit{content preservation} (i.e., if the input
semantics is preserved), \textit{transfer accuracy/strength} (i.e., if
the intended attribute is achieved through the transfer), and
\textit{fluency or naturalness} (i.e., if the generated text appears
natural) \citep{pang-gimpel-2019-unsupervised,
  mir-etal-2019-evaluating} often misrepresent the actual quality of
the output. As a consequence, expensive human-assisted evaluations
turn out inevitable
\citep{briakou-etal-2021-evaluating,briakou-etal-2021-review}.

\subsection{Purpose and scope of this survey} 
With the spurt of deep learning, style transfer has become a
collective enterprise in NLP \citep{hu2020text,jinreview}.  Much work
has explored techniques that separate style from content, and has
investigated the efficacy of different systems that share some basic
workflow components. Typically, a style transfer pipeline comprises an
encoder-decoder architecture inducing the target attribute on a latent
representation of the input, either directly
\citep{dai-etal-2019-style} or after the initial attribute has been
stripped away \citep{cheng-etal-2020-improving}. Different frameworks
have been formulated on top of this architecture, ranging from lexical
substitutions \citep{ijcai2019-maskinfill,li-etal-2018-delete}, to
machine translation
\citep{jin-etal-2019-imat,mishra-etal-2019-modular} and adversarial
techniques
\citep{pang-gimpel-2019-unsupervised,lai-etal-2019-multiple}.
Therefore, the time seems ripe for a survey of the task, and with this
paper, we contribute to organizing the existing body of knowledge
around it.

The recurring approaches to style transfer make it reasonable to
review its methods, but there already exist three surveys that do so
\citep{hu2020text,jinreview,toshevska2021review}. They take a
technical perspective and focus on the \textit{methods} used to
transfer styles.  Automatic metrics and evaluation practices have been
discussed as well in previous publications
\citep{briakou-etal-2021-evaluating,briakou-etal-2021-review}.  We
move to a different and complementary angle which puts focus on the
\textit{styles} to be transferred. Our leading motive is a question
that is rooted in the field but is rarely faced: \textit{Can all
  textual styles be changed or transferred?}

Current publications in the field see style transfer by and large from
an engineering angle, aiming at acceptable scores for the three style
transfer criteria, and comparing their numerical results in a limited
fashion: they neglect the peculiarities of the styles that they are
transferring. In our view, each style requires robust understanding in
itself, as a pre-requisite for the applied transfer models' choice and
success.  We thus provide a detailed look into both well-established
styles, and those that remain under-explored in the literature.
Instead of asking \textit{Is that method advantageous for style
  transfer?}, we are interested in questions like \textit{How well
  does it perform when dealing with a particular style?}  and
\textit{Is finding a balance between naturalness, transfer, and
  content preservation equally difficult for all styles?}  In this
vein, we propose a hierarchy of styles that showcases how they relate
to each other. We not only characterize them separately and by tapping
on some insights coming from humanity-related
disciplines\footnote{Also \cite{jinreview} compare various styles and
  their respective definitions, but in a data-driven approach, as
  features that vary across datasets.}, but we also illustrate how
they have been handled in the context of style transfer, covering the
challenges that they pose (e.g., lack of data), their potential
applications, and the methods that have been employed for each of
them.  Further, we observe if such models have been evaluated in
different ways (some of which could fit a style more than others), and
lastly, we consider how well styles have been transferred with respect
to the three style transfer criteria.  Our hierarchy incorporates a
selection of papers published from 2008 to September 2021 that we
found relevant because of their use or development of datasets for the
task at hand, for their proposal of methods that later became
well-established in the field, or alternatively, for their proposed
evaluation measures. A few of these studies tackle Chinese
\citep{su-etal-2017-rephrasing, shang-etal-2019-semi}, a handful of
them deal with multilingual style transfer
\citep{niu-etal-2018-multi-task-formality, briakou-etal-2021-ola}, but
most works address style transfer for English.

The paper is structured as follows.
Section~\ref{sec:style-transfer-methods} summarizes the technical
approaches to this task, covering also some recurring evaluation
techniques. Our main contribution, organizing styles in a hierarchy,
is outlined in Section~\ref{sec:style-hierarchy} (with details in
Sections~\ref{sec:unintented} and \ref{sec:intended}). These
discussions include descriptions of data, methods, as well as the
evaluations employed for their transfer performance.
Section~\ref{sec:conclusions} concludes this work and indicates
possible directions for future research.

\subsection{Intended audience}
This survey is addressed to the reader seeking an overview of the
state of affairs for different styles that undergo
transfer. Specifically, we aim for the following.

\textit{Readers needing a sharp focus on a specific style.} We revise
what has been done within the scope of each style, which could hardly
be found in works with a more methodological flavour.

\textit{Readers preparing for upcoming style transfer studies},
interested in the research gaps within the style transfer
landscape. On the one hand, this review can help researchers
categorize future work among the massive amount produced in this
field, indicating similar works to which they can compare their
own. This can eventually guide researchers to decide on 
the appropriate models for their specific case. On the
other hand, we suggest possible ``new'' styles that were not
treated yet but which have an affinity to the existing ones.

\textit{Readers questioning the relationship between content and
  style.}  NLP has fallen short in asking what textual features can be
taken as a style, and has directly focused on applying transfer
procedures -- often generating not too satisfying output texts.
Without embarking on the ambitious goal of defining the concept of
``style'', we systematize those present in NLP along some
theoretically-motivated coordinates.

\section{Style transfer methods and evaluation}
\label{sec:style-transfer-methods}
Our survey focuses on styles and relations among them. To connect the
theoretical discussion with the methodological approaches to transfer,
we now briefly describe the field from a technical perspective. We
point the readers to \citet{jinreview}, \citet{hu2020text} and
\cite{toshevska2021review} for a comprehensive clustering and review
of the existing methods, and to \cite{prabhumoye-etal-2020-exploring}
for a high-level overview of the techniques employed in controlled
text generation, style transfer included.
 
\begin{figure}
  \centering
  \scalebox{0.7}{%
    \begin{tikzpicture}[grow'=right,level distance=2in,sibling distance=.001in]
	\tikzset{edge from parent/.style={draw, ->}}
	\tikzset{every tree node/.style={align=center}}
	
	\Tree [.{ST}
	[.{Parallel \\ Data } ]
	[.{Non-Parallel \\ Data} 
	[.{Explicit Style-Content \\ Disentanglement} ]
	[.{Implicit Style-Content \\ Disentanglement} [.{\\\begin{varwidth}{\textwidth}\begin{itemize}\item Backtranslation \item Adversarial  Learning \item Attribute Controlled  Generation \item Other methods \end{itemize}\end{varwidth}} ]]
	[.{Without \\ Disentanglement} [.{\\\begin{varwidth}{\textwidth}\begin{itemize}\item Entangled Latent Representation Editing \item Reinforcement Learning \item Attribute Controlled Generation \item Probabilistic Modelling  \end{itemize}\end{varwidth}} ]]]
	]
	]
      \end{tikzpicture}
    }\\[7mm]
  \caption{Methods discussed in previous style transfer surveys,
    adapted from \cite{hu2020text}. In contrast, our contribution is
    the inspection of styles depicted in Figure~\ref{fig:hierarchy}.}
  \label{fig:methods}
\end{figure}
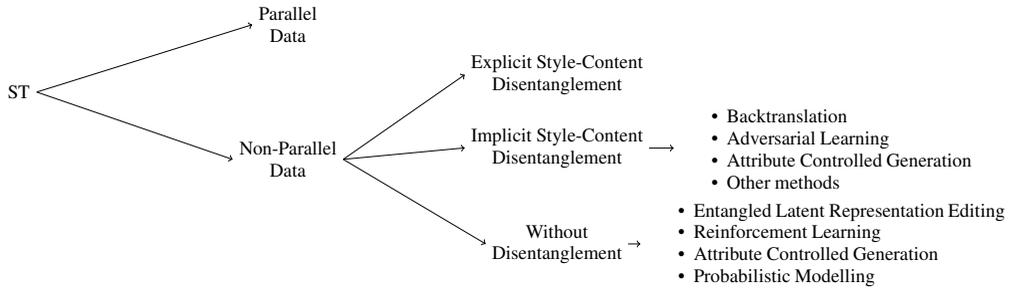

Methodological choices typically depend on what data is available.  In
the ideal scenario, the transfer system can directly observe the
linguistic realization of different stylistic attributes on parallel
data. However, parallel data cannot be easily found or created for all
styles. On the other hand, mono-style corpora that are representative
of the attributes of concern might be accessible (e.g., datasets of
texts written for children and datasets of scholarly papers), but they
might have little content overlap -- thus making the learning of
content preservation particularly challenging
\citep{romanov-etal-2019-adversarial}.  Therefore, we group style
transfer methods according to these types of corpora, that is,
parallel resources (either ready to use
\citep[i.a.]{xu-etal-2012-paraphrasing, rao-tetreault-2018-dear} or
created via data augmentation strategies
\cite[i.a.]{zhang-etal-2020-parallel}), and mono-style datasets
\cite[i.a.]{shen2017style, li-etal-2018-delete,
  john-etal-2019-disentangled}.  As illustrated in
Figure~\ref{fig:methods}, which adapts the taxonomy of methods
presented in \cite{hu2020text}, the two groups are further divided
into sub-categories with respect to the training techniques adopted to
learn the task.

Throughout the paper, such methods are reported to organize the
literature in Table~\ref{tab:unintended},
Table~\ref{tab:unintendeddynamic}, Table~\ref{tab:intendedtargeted},
Table~\ref{tab:intendednontargetedcircumstancial} and
Table~\ref{tab:intendednontargetedconventional}, which inform the
reader about the approach that each study has taken for a given style,
the approaches that have not yet been leveraged for it (i.e., no author
is reported in a cell of a table), and those that have been
indiscriminately applied for multiple styles (e.g., the same authors
appear more than once in a table, or appear in many of them).

\subsection{Parallel Data}
A parallel corpus for transfer would contain texts with a particular
stylistic attribute on one side (e.g., formal texts) and paraphrases
with a different attribute on the other (e.g., informal texts).  When
such datasets exist, style transfer can be approached as a translation
problem that maps one attribute into the other. Using a corpus of
Shakespearean texts and their modern English equivalents,
\cite{xu-etal-2012-paraphrasing} demonstrated the feasibility of
style-conditioned paraphrasing with phrase-based machine translation.
Later, neural models started to be trained to capture fine stylistic
differences between the source and the target sentences, one instance
at a time.  \cite{jhamtani-etal-2017-shakespearizing}, for example,
improved the transfer performance on the Shakespearean dataset by
training a sequence-to-sequence architecture with a pointer network
that copies some words from the input.  \cite{rao-tetreault-2018-dear}
corroborated that machine translation techniques are a strong baseline
for style transfer on the Grammarly's Yahoo Answers Formality Corpus,
a parallel corpus for formality transfer which turned out to drive the
majority of the style transfer research on parallel data
\citep[leveraged by][among others]{niu-etal-2018-multi-task-formality,
  wang-etal-2019-harnessing, xu2019formality}.

Sequence-to-sequence models achieved remarkable results in conjunction
with different style controlling strategies, like multi-task learning
\citep{niu-etal-2018-multi-task-formality, xu2019formality}, rule
harnessing \citep{wang-etal-2019-harnessing}, post-editing with
grammatical error correction \citep{ge-etal-2019-automatic}, and
latent space sharing with matching losses
\citep{wang-etal-2020-formality}.  Parallel resources, however, are
scarce or limited in size. This has triggered a number of attempts to
synthesize parallel examples.  \cite{zhang-etal-2020-parallel} and
\cite{jin-etal-2019-imat} exemplify this effort.  While the former
augmented data with translation techniques (i.e., backtranslation and
backtranslation with a style discriminator) and a multi-task transfer
framework, \cite{jin-etal-2019-imat} derived a pseudo-parallel corpus
from mono-style corpora in an iterative procedure, by aligning
sentences which are semantically similar, training a translation model
to learn the transfer, and using such translations to refine the
alignments in return.

\subsection{Non-parallel data}
The paucity of parallel resources also encouraged transfer strategies
to develop on mono-style corpora. This research line mainly approached
the task intending to disentangle style and content, either by
focusing the paraphrasing edits on the style-bearing portions of the
input texts, or by reducing the presence of stylistic information into
the texts' latent representations. On the other hand, a few studies
claimed that such disentanglement can be avoided.  Therefore, methods
working with non-parallel data can be divided into those which do
style transfer with an explicit or implicit style-to-content
separation and those which operate no separation.

\subsubsection{Explicit Style-Content Disentanglement.}
Some styles have specific markers in text: expressions like
``\textit{could you please}'' or ``\textit{kindly}'' are more typical
of a formal text than an informal one. This observation motivated a
spurt of studies to alter texts at the level of explicit markers --
which are replaced in the generated sentences by the markers of a
different attribute.  The first step of many such studies is to find a
comprehensive inventory of style-bearing words. Strategies devised
with this goal include frequency statistics-based methods
\citep{li-etal-2018-delete, madaan-etal-2020-politeness}, lexica
\citep{wen-etal-2020-decode}, attention scores of a style classifier
\citep{xu-etal-2018-unpaired, sudhakar-etal-2019-transforming,
  helbig-etal-2020-challenges, reid-zhong-2021-lewis}, or combinations
of them \citep{ijcai2019-maskinfill, lee-2020-stable}. As an
alternative, \cite{malmi-etal-2020-unsupervised} identified spans of
text on which masked language models \citep{devlin-etal-2019-bert},
trained on source and target domains, disagree in terms of likelihood:
these would be the portions of a sentence responsible for its style,
and their removal would produce a style-agnostic representation for
the input.

Candidate expressions are then retrieved to replace the source markers
with expressions of the target attribute. Distance metrics used to
this end are (weighted) word overlap \citep{li-etal-2018-delete},
Euclidean distance \citep{li-etal-2018-delete} and cosine similarity
between sentence representations like content embeddings
\citep{li-etal-2018-delete}, weighted TF-IDF vectors and averaged
GloVe vectors over all tokens \citep{sudhakar-etal-2019-transforming}.
Some studies resorted instead to WordNet-based retrievals
\citep{helbig-etal-2020-challenges}.

In the last step, (mostly) neural models combine the retrieved tokens
with the style-devoid representation of the input, thus obtaining an
output with the intended attribute. There are also approaches that skip this
step and directly train a generator to produce sentences in the target
attribute based on a template \citep[i.a.]{lee-2020-stable}. Similar
techniques for explicit keyword replacements are relatively easy to
train, and are more explainable than many other methods, like adversarial
ones \citep{madaan-etal-2020-politeness}.

\subsubsection{Implicit style-content disentanglement}
\label{methods-implicit}
Approaches for explicit disentanglement cannot be extended to all
styles because many of them are too complex and nuanced to be reduced
to keyword-level markers. Methods for implicit disentanglement
overcome this issue.  Their idea is to strip the input style away by
operating on the latent representations (rather than at the text
level). This usually involves an encoder-decoder architecture.
The encoder produces the latent representation of the input and the
decoder, which generates text, is guided by training losses
controlling for the style and content of the output.

\paragraph{Adversarial learning} Implicit disentanglement has been
instantiated by adversarial learning in several ways.  To ensure that
the representation found by the encoder is devoid of any style-related
information, \cite{fu2018} trained a style classifier adversarially,
making it unable to recognize the input attribute, while
\cite{lin-etal-2020-learning} applied adversarial techniques to
decompose the latent representation into a style code and a content
code, demonstrating the feasibility of a one-to-many framework (i.e.,
one input, many variants). \cite{john-etal-2019-disentangled} inferred
embeddings for both content and style from the data, with the help of
adversarial loss terms that deterred the content space and the
style space from containing information about one another, and with a
generator that reconstructed input sentences after the words
carrying style were manually removed. Note that, since
\cite{john-etal-2019-disentangled} approximated content with words that
do not bear sentiment information, they could also fit under the group
of Explicit Style-Content Disentanglement. We include them here
because the authors themselves noted that ignoring sentiment words 
can boost the transfer, but is
not essential.

\paragraph{Backtranslation} A whole wave of research banked on the
observation that backtranslation washes out some stylistic traits of
texts \citep{rabinovich-etal-2017-personalized} and followed the work
of \cite{prabhumoye-etal-2018-style}. There, input sentences were
translated into a pivot language and back as a way to manipulate their
attributes: the target values were imposed in the backward direction,
namely, when decoding the latent representation of the (pivot
language) text, thus generating styled paraphrases of the input (in
the source language).
	 
\paragraph{Attribute controlled generation} Attribute control proved
to be handy to produce style-less representations of the content while
learning a code for the stylistic attribute. This emerges, for
instance, in \cite{pmlr-v70-hu17e}, who leveraged a variational
auto-encoder and some style discriminators to isolate the latent
representation and the style codes, which were then fed into a
decoder. While the discriminators elicited the disentanglement, the
constraint that the representation of source and target sentence
should remain close to each other favored content preservation.
	
\paragraph{Other Methods} An alternative path to disentanglement stems
from information theory. \cite{cheng-etal-2020-improving} defined an
objective based on the concepts of mutual information and variation of
information as ways to measure the dependency between two random
variables (i.e., style and content). On the one hand, the authors
minimized the mutual information upper bound between content and style
to reduce their interdependency; on the other, they maximized the
mutual information between latent embeddings and input sentences,
ensuring that sufficient textual information was preserved.

\subsubsection{Without disentanglement} 
By abandoning the disentanglement venture, some studies argued that
separating the style of a text from its content is not only difficult
to achieve -- given the fuzzy boundary between the two, but also
superfluous \citep{lample2018multiple}. This observation became the
core of a wave of research that can be categorized as follows.

\paragraph{Entangled latent representation editing} Some works
edited the latent representations of the input texts
learned by an auto-encoder. A common practice in this direction is to
jointly train a style classifier and iteratively update the
auto-encoder latent representation by maximizing the confidence on the
classification of the target attribute \citep{pmlr-v70-mueller17a,
  Liu_Fu_Zhang_Pal_Lv_2020}.  Another approach trained a multi-task
learning model on a summarization and an auto-encoding task, and it
employed layer normalization and a style-guided encoder attention
using the transformer architecture \citep{wang_2019_latent_edit}.

\paragraph{Attribute controlled generation} 
Proven successful by disentanglement-based studies, methods for
learning attribute codes were also applied without the
content-vs.-style separation. \cite{lample2018multiple}, for instance,
employed a denoising auto-encoder together with backtranslation and an
averaged attribute embedding vector, which controlled for the presence
of the target attribute during generation.  Instead of averaging the
one-hot encoding for individual attribute values, \cite{smith2019zero} used
supervised distributed embeddings to leverage similarities between
different attributes and perform zero-shot transfer.
	
\paragraph{Reinforcement learning} 
Multiple training loss terms have been defined in style
transfer to endow the output texts with the
three desiderata of content preservation, transfer accuracy and text
naturalness -- often referred to as ``fluency''. 
The dependency on differentiable objectives can be bypassed
with reinforcement learning, which uses carefully designed training
rewards \citep[i.a.,]{luo-etal-2019-towards}. Generally, rewards that
cope with the presence of the target attribute are based on some style
classifiers or discriminators, those pertaining to naturalness
rely on language
models; and those related to content preservation use \textsc{Bleu} or
similar metrics that compare an output text against some reference.

\cite{gong-etal-2019-reinforcement} worked in a generator-evaluator
setup. There, the generator's output was probed by an evaluator
module, whose feedback helped improve the output attribute, semantics and
fluency. Two building blocks can also be found in \cite{Luo19DualRL}.
They approached style transfer as a dual task (i.e., source-to-target
and target-to-source mappings) in which, to warm-up the reinforcement
learning training, a model was initially trained on a pseudo-parallel
corpus.  \cite{wu-etal-2019-hierarchical-reinforced}, instead,
explored a sequence operation method called Point-Then-Operate, with a
high-level agent dictating the text position where the operations
should be done and a low-level agent performing them. Their
policy-based training algorithm employed extrinsic and intrinsic
rewards, as well as a self-supervised loss to model the three transfer
desiderata. The model turned out relatively interpretable thanks to
these explicitly defined operation steps. Tuning their number, in
addition, allowed to control the trade-off between the presence of the
initial content and of the target attribute.
	
An exception among reinforcement learning studies is the cycled
reinforcement learning of \cite{xu-etal-2018-unpaired}, which fall
within the disentangling picture.

\paragraph{Probabilistic modelling} Despite being a common practice in
unsupervised learning, the definition of task-specific losses can lead
to training instability. These objectives are empirically determined
among a vast number of possible alternatives. To overcome the issue,
\cite{he2020a} formulated a probabilistic generative strategy that
follows objectives defined by some principles of probabilistic
inference, and which makes clear assumptions about the data. This
approach allowed them to reason transparently about their system
design, and to outperform many works choosing ad-hoc
training objectives.

\subsection{Evaluation}
\label{sec:style-transfer-evaluation}

The methods presented above are usually assessed with metrics that
quantify content preservation, transfer accuracy/intensity and
generation of natural-sounding paraphrases.  A detailed discussion of
the evaluation methods can be found in
\cite{mir-etal-2019-evaluating}, \cite{pang2019daunting},
\cite{briakou-etal-2021-evaluating} and
\cite{briakou-etal-2021-review}, with the latter focusing on human
evaluation settings.  As they appear in most style transfer
publications, we briefly introduce them here and will refer back to
them throughout the paper.

Content preservation, i.e., the degree to which an output retains the
content of the input, is usually gauged with measures that originated
in machine translation. They compute the overlap between the words of
the generation system and some reference texts, under the assumption
that the two should share much lexical material.  Among them are
\textsc{Bleu} \citep{papineni2002bleu} and \textsc{Meteor}
\citep{banerjee2005meteor}, often complemented with \textsc{Rouge}
\citep{lin2004rouge}, initially a measure for automatic
summaries. Transfer accuracy, i.e., the efficacy of the models in
varying stylistic attributes, is usually scored by classifiers:
trained on a dataset characterized by the style in question, a
classifier can tell if an output text has the target attribute or not.
Applied on a large scale, this second criterion can be quantified as
the percentage of texts that exhibit the desired attribute. Last comes
the naturalness or fluency of the variants that have been changed in
style. This is typically estimated with the perplexity of language
models, indicating the degree to which a sequence of words in
a paraphrase is predictable --
hence grammatical.

Focusing on automatic content preservation,
\cite{tikhonov-etal-2019-style} advocated that \textsc{Bleu} should be
used with some caution in style transfer. They argued that the
entanglement between semantics and style in natural language is
reflected in the entanglement between the \textsc{Bleu} score measured
between input and output and the transfer accuracy.  Indeed, they
provided evidence that such measures can be easily manipulated: the
outputs that a classifier in the generative architecture
indicates as having the incorrect attribute could be replaced with
sentences which are most similar to the input in their surface form --
thus boosting both the reported accuracy and \textsc{Bleu}.
Human-written reformulations are necessary in their view for upcoming
experiments, as current style transfer architectures become more
sophisticated, and therefore, accuracy and \textsc{Bleu} might be too
naive metrics to estimate their performance.  Going in a similar
direction, the extensive meta-analysis of
\cite{briakou-etal-2021-evaluating} discusses the pitfalls of
automatic methods and the need for standardized evaluation practices
(including human evaluation) to boost advance in this field.

\section{Style hierarchy}
\label{sec:style-hierarchy}

\begin{figure}
\centering
\scalebox{0.7}{\begin{tikzpicture}[grow'=right,level distance=1.4in,sibling distance=.18in]\hspace{-.7cm}
    \tikzset{edge from parent/.style={draw, ->}}
    \tikzset{every tree node/.style={align=center}}
    
    \Tree [.Styles
    [.{Unintended} 
    [.{Persona} [.{\\\begin{varwidth}{\textwidth}\begin{itemize}\item Gender and Age \item Personality Traits\item Background\\(Country and Ethnicity,\\ Education and Culture)\end{itemize}\end{varwidth}} ]]
    [.{Dynamic\\States} [.{\\\begin{varwidth}{\textwidth}\begin{itemize}\item Writing Time \item Subjective Bias\end{itemize}\end{varwidth}} ]]]
    [.{Intended} 
    [.{Targeted} [.{\\\begin{varwidth}{\textwidth}\begin{itemize}\item Emotion State (*)\item Sentiment (*)\item Sarcasm  \item Political Slant (*)\end{itemize}\end{varwidth}} ]]
    [.{Non-\\targeted} [.{Circumstantial\\Registers} [.{\\\begin{varwidth}{\textwidth}\begin{itemize}\item Formality\item Politeness  \item Humor \item Offensiveness \item Literality\end{itemize}\end{varwidth}} ]][.{Conventional\\Genres} [.{\\\begin{varwidth}{\textwidth}\begin{itemize}\item News \item Technical Language\item Literature \item Song Lyrics \end{itemize}\end{varwidth}} ]]]
    ]
    ]
  \end{tikzpicture}
}\\[14mm]
\caption{The hierarchy of styles guiding our discussion. Each branch defines different challenges for style transfer and illustrates how styles relate to one another. Asterisks (*) mark the nodes on the fence between
content and style, since altering their attributes brings substantial content loss -- they are included in the hierarchy nevertheless, because they have been leveraged for the transfer goal.}
\label{fig:hierarchy}
\end{figure}
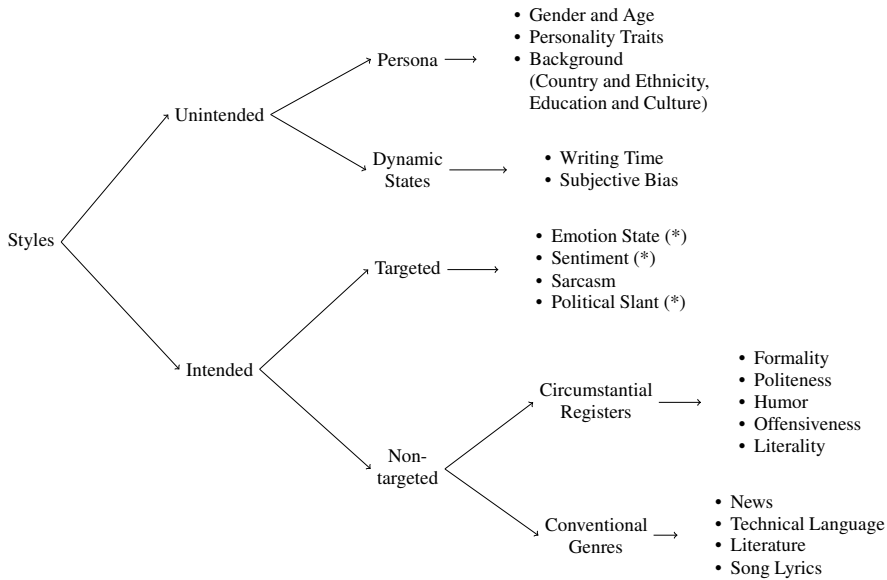

Style transfer relies on a conceptual distinction between meaning and form
\citep[e.g.,][]{saussure}, but what is this form? It is a
dimension of sociolinguistic variation that manifests in
syntactic and lexical patterns, that can be correlated with
independent variables and that, according to \cite{bell}, we shift in
order to fit an audience. Bell's characterization emphasizes the
intentionality of language variation, accounting only for the styles
ingrained in texts out of purpose. Yet, many others emerge as a
fingerprint of the authors' identities, for instance from specific markers 
of people's personality and internal states
\citep{brennan2012adversarial}. This already suggests that
different styles
have diverse characteristics. However,
their peculiar challenges have received little attention in the 
literature.  As a remedy for the lacuna, we bring style transfer closer to
the linguistic and sociological theories on the phenomenon it targets.
We propose a hierarchy of styles in which we place the relevant
body of NLP research.

A recent study by \cite{kang-hovy-2021-style} actually groups styles
into a handful of categories (\textit{personal},
\textit{interpersonal}, \textit{figurative} and \textit{affective})
based on some social goals achieved through communication. Their work
did not investigate specific styles. It rather intended to fertilize
research towards a cross-style direction, by combining existing
corpora into an overarching collection of 15 styles.\footnote{Their
  data, part of which is annotated with 15 styles, is available at
  \url{https://github.com/dykang/xslue}.} By contrast, our hierarchy
concentrates on the peculiarities of styles separately, while
indicating the methods that have been used and those that have been
dismissed for each of them.

To unify the above-mentioned theoretical insights, we make a first,
coarse separation between accidental and voluntary styles, structuring
them into the \textit{unintended} and \textit{intended}
families.\footnote{We do not include works on the transfer of syntax,
  such as \cite{bao-etal-2019-generating},
  \cite{balasubramanian-etal-2021-polarized} and
  \cite{lyu-etal-2021-styleptb}.} The former group copes with the
self. It corresponds to the personal characteristics of the authors, which
we split into factors that define \textit{between-persons} and
\textit{within-person} language variations. Namely, there are stable
traits defining systematic differences between writers and short-term
internal changes within an individual subject which, in response to
situations, do not persist over time \citep{beckmann2017dynamic}.  We
call them \textit{persona} and \textit{dynamic states} respectively.
The other category of styles is \textit{intended}, as it covers
deliberate linguistic choices with which authors adapt to their
communicative purpose or environment. Style transfer publications that
fall within this group echo what is known as ``palimpsest'' in
literary theories, i.e., the subversion of a text into a pastiche or a
parody to imitate an author, degrade a text, or amplify its content
\citep{genette1997palimpsests}.  Among these are styles used to
express how one feels about the topic of discussion: a speaker/writer
can have a positive sentiment \textit{on} a certain matter, be angry
or sad \textit{at} it, be sarcastic \textit{about} it, etc. Of this
type are styles \textit{targeted} towards a topic, while others, the
\textit{non-targeted} subset, are more independent of it.  Some
(\textit{circumstantial registers}) are rather dependent on the
context in which they are deployed, and they convey a general attitude
of the writers, a tone in which they talk or a social posture -- an
example being formality, that speakers increase if they perceive their
interlocutor as socially superior \citep{vanecek1975bericht}.  Other
styles are socially coded. They can be thought of as
\textit{conventional} writing styles tailored to the ideal addressee
of the message rather than an actual one, and are typically employed
in mass communication, such as scientific, literary, and technical
productions.

These categories subsume a number of individual styles. For instance,
\textit{persona} branches out into \textit{personality traits},
\textit{gender and age}, and \textit{background}, which in turn
encompasses \textit{country and ethnicity}, \textit{education and
  culture}. Note that the leaves in our hierarchy are the major styles
that have been addressed so far by automatic systems, but many others
can be identified and explored in future work. We include some in our
discussions.  Furthermore, we acknowledge that a few styles
pertain to both the \textit{unintended} and \textit{intended}
branches. Our motivation to insert them under one rather than the
other is due to the type of data on which the transfer was made (e.g.,
\textit{emotion state}) or to how the problem was phrased
by the corresponding studies (e.g., \textit{literature}).

The remainder of this paper follows the structure of our hierarchy.
We provide a top-down discussion of the nodes, starting from the
high-level ones, which are presented from a theoretical perspective, and
proceeding towards the leaves of the branches, which is where the
concrete style transfer works are examined in relation to the data,
the methods and the evaluation procedures that they used.

\section{Unintended styles}
\label{sec:unintented}
Writers leave traces of their personal data. Information like one's
mental disposition, biological and social status are revealed by
stylometric cues present in a text.  These cues might be produced
unknowingly, and because of that, they could help to combat
plagiarism, foster forensics and support humanities.  On the other
hand, accessing knowledge about writers could breach people's privacy
and exacerbate demographic discrimination.  Hence, while
classification-based studies leveraged such latent information to
profile people's age and gender
\citep{rosenthal-mckeown-2011-age,nguyen2013old,sarawgi2011gender,fink2012inferring},
geolocation and personality
\citep{eisenstein-etal-2010-latent,verhoeven-etal-2016-twisty,plank-hovy-2015-personality},
the attempt to defeat authorship recognition moved research towards
the transfer of such unintended styles -- i.e., age, gender, etc.

Arguably the first work to address this problem is that of
\cite{brennan2012adversarial}, who tried to confound stylometric
analyses by backtranslating existing texts with available translation
services, such as Google Translate and Bing Translator.  Their
preliminary results did not prove successful, as the writer's identity
remained recognizable through the translation passages from source to
targets and back, but follow-up research provided evidence that
automatic profilers can be effectively fooled
\cite[i.a.,]{kacmarcik-gamon-2006-obfuscating,emmery-etal-2018-style,shetty2018a4nt,bo2020authorship}.

Successive style transfer studies narrowed down the considered authors' traits.
They tackled stable features that are a proxy for the writers'
biography, which we subsume under the category of \textit{persona}, or
more \textit{dynamic states} that characterize writers at a specific
place and time. It should be noticed that such works rely on a tacit
assumption about writers' authenticity:
writers express themselves spontaneously and do not attempt to mask
their own traits \citep{brennan2012adversarial}.

We illustrate the methods used to 
transfer \textit{unintended} styles in Table~\ref{tab:unintended}.

\begin{table}
  \centering\small
  \caption{Style transfer methods and
    the \textit{unintended} styles of \textit{persona} (Pers.:
    \textit{Personality Traits}, Co.: \textit{Country}, Variety: \textit{English Variety}, Et.:
    \textit{Ethnicity}, Edu.: \textit{Education}).}
  \label{tab:unintended}
  \begin{tabular}{lp{0.2\linewidth}p{0.2\linewidth}p{0.2\linewidth}p{0.2\linewidth}}
    \toprule[1pt]	
    & \multicolumn{1}{c}{Parallel} & \multicolumn{3}{c}{Non-parallel}
    \\
    \cmidrule(r){2-2}\cmidrule(l){3-5} 
    & & Exp. Disent. & Imp. Disent. & No Disent.
    \\ 
   \cmidrule(lr){3-3}\cmidrule(lr){4-4}\cmidrule(l){5-5}
    \mrrt{3}{Gender}
    & Kang \citeyear{kang-etal-2019-male}
                    & Reddy \citeyear{reddy2016obfuscating}\par
                      Sudhakar\citeyear{sudhakar-etal-2019-transforming}\par
                      Madaan \citeyear{madaan-etal-2020-politeness}
    & Prabhumoye \citeyear{prabhumoye-etal-2018-multilingual-back-translation}\par
      Prabhumoye \citeyear{prabhumoye-etal-2018-style}\par
      Nangi \citeyear{nangi-etal-2021-counterfactuals}    
      & Preotiuc-Pietro \citeyear{preotiuc2016discovering}\par
        Lample \citeyear{lample2018multiple}\par
        Liu \citeyear{liu2020revision}
    \\
    \cmidrule(r){2-2}\cmidrule(lr){3-3}\cmidrule(lr){4-4}\cmidrule(l){5-5}
    \mrrt{1}{Age}
    &Kang \citeyear{kang-etal-2019-male}
                    & \texttt{-{}-}
    &\texttt{-{}-}
      & Lample \citeyear{lample2018multiple}
    \\
    \cmidrule(r){2-2}\cmidrule(lr){3-3}\cmidrule(lr){4-4}\cmidrule(l){5-5}
    \mrrt{1}{Pers.}
    & Bujnowski \citeyear{bujnowski-etal-2020-empirical}
                    &\texttt{-{}-}
    & Cheng \citeyear{cheng-etal-2020-improving}
      &\texttt{-{}-}
    \\
    \cmidrule(r){2-2}\cmidrule(lr){3-3}\cmidrule(lr){4-4}\cmidrule(l){5-5}
    \mrrt{1}{Co.}
    & Kang \citeyear{kang-etal-2019-male}
                    &\texttt{-{}-}
    &\texttt{-{}-}
      & Riley \citeyear{riley-etal-2021-textsettr}
    \\
    \cmidrule(r){2-2}\cmidrule(lr){3-3}\cmidrule(lr){4-4}\cmidrule(l){5-5}
    \mrrt{2}{Variety}
    &\texttt{-{}-}
                    &\texttt{-{}-}
    & Logeswaran \citeyear{logeswaran2018content}
      & Lee \citeyear{lee-etal-2019-neural} \par
        Krisha \citeyear{krishna-etal-2020-reformulating}
    \\
    \cmidrule(r){2-2}\cmidrule(lr){3-3}\cmidrule(lr){4-4}\cmidrule(l){5-5}
    \mrrt{1}{Et.}
    &  Kang \citeyear{kang-etal-2019-male}
                    &\texttt{-{}-}
    &\texttt{-{}-}
      & Krishna \citeyear{krishna-etal-2020-reformulating}
    \\ 
    \cmidrule(r){2-2}\cmidrule(lr){3-3}\cmidrule(lr){4-4}\cmidrule(l){5-5}
    \mrrt{1}{Edu.}
    & Kang \citeyear{kang-etal-2019-male} & \texttt{-{}-} & \texttt{-{}-} & \texttt{-{}-} \\[5pt]
    \bottomrule[1pt]	
  \end{tabular}
\end{table}

\subsection{Persona}
\label{persona}
\textit{Persona} includes biographic attributes coping with
personality and people's social identity.  Individuals construct
themselves ``as girls or boys, women or men – but also as, e.g., Asian
American'' \citep{eckert_ginet_1999}, that is, they often form an idea
of the self as belonging to a group with a shared enterprise or
interest \citep{tajfel1974social}.  The interaction within such a group
also affects their linguistic habits \citep{lave1991situated} as they
develop a similar way of talking.  In this sense, linguistic style is
a key component of one's identity \citep{giles1987ethnolinguistic}. It
manifests some traits unique to a specific person or
community (\citet{mendoza1993} provide insights on the topic with
respect to the Asian-American English speech).

At least to a degree, \textit{persona} styles are implicit in the way
people express themselves.  As opposed to the \textit{intended} branch
of our hierarchy, they are not communicative strategies consciously
set in place by the writers, but they are spontaneous indicators
of other variables.  For instance, it has been shown that women tend
to use paralinguistic signals more often than men
\citep{carli1990gender}, that speakers' vocabulary becomes more
positively connotated and less self-referenced in older ages
\citep{pennebaker2003words}, and that sub-cultures express themselves
with a specific slang \citep{bucholtz2006word}.

The transfer of \textit{persona} aims to go from one attribute to the
other (e.g., young to old for the style of age), and its main challenge is
that different styles are closely intertwined. Age and gender, for
instance, can imply each other because ``the appropriate age for
cultural events often differs for males and females''
\citep{eckert1997}, and therefore, one may not be changed without
altering the other.  Moreover, there is still a number of styles
dealing with people's communicative behaviours and skills which are
left unexplored. Future studies could focus on those, like the pairs
playful vs.\ aggressive, talkative vs.\ minimally responsive,
cooperative vs.\ antagonist, dominant vs.\ subject, attentive
vs.\ careless, charismatic vs.\ uninteresting, native vs.\ L2 speaker,
curious vs.\ uninterested, avoidant vs.\ involved.

\subsubsection{Gender and age}
Style transfer treats gender and age as biological facts.
The transfer usually includes a mapping between discrete labels: from
male to female or vice versa, and from young to old or the other way
around (see some examples in Table~\ref{persona-examples}).  It should
be noticed that such labels disregard the fluidity of one's gender
experience and performance, which would be better described along a
spectrum \citep{eckert2003language}, and they represent age as a
chronological variable rather than a social one depending on peoples'
personal experiences \citep{eckert1997}.  This simplification is not
made by style transfer specifically, but it is common to many studies
focused on authors' traits, due to how the available datasets were
constructed -- e.g., in gender-centric resources, labels are inferred
from the name of the texts' authors \citep{mislove2011understanding}.

The Rt-Gender corpus created by \cite{voigt-etal-2018-rtgender} stands
out among such resources. It was built to research how responses
towards a specific gender differ from responses directed to
another, in opposition to related corpora that collect
linguistic differences between genders.  This labelled dataset
potentially sets the ground for the next steps in style transfer.

\paragraph{Data} Works on \textit{gender} style transfer typically
follow the choice of data by \cite{reddy2016obfuscating}, who used
tweets posted in the US in 2013 and some reviews from the
Yelp\footnote{\url{https://www.yelp.com/dataset}} dataset, and
inferred gender information from the users' names.

For this style there also exists
\textsc{Pastel}\footnote{\url{https://github.com/dykang/PASTEL}}, a
corpus annotated with attributes of both \textit{unintended} and
\textit{intended} styles. That is the result of the crowdsourcing
effort conducted by \cite{kang-etal-2019-male}, in which $\approx$41K
parallel sentences were collected in a multimodal setting, and which
were annotated with the gender, age, country, political view,
education, ethnicity, and time of writing of their authors.

The need to collect attribute-specific rewrites further motivated
\cite{xu-etal-2019-alter} to create \textsc{Alter}. As a publicly
available tool\footnote{\url{https://github.com/xuqiongkai/ALTER}},
\textsc{Alter} was developed to overcome one major pitfall of
crowdsourcing when it comes to generating gold standards: human
annotators might fail to associate textual patterns to a gender label,
at least when dealing with short pieces of text. \textsc{Alter}
facilitates their rewriting tasks (specifically, to generate texts
which are not associated with a particular gender) by providing them
with immediate feedback.

\begin{table}
  \centering\small
  \caption{Examples of style transfer on a subset of \textit{persona} styles. \textit{Personality traits} sentences come from 
    \cite{shuster2019engaging}, \textit{gender}-related ones from \cite{sudhakar-etal-2019-transforming}, the \textit{age}-related 
    examples from \cite{preotiuc2016discovering} and \textit{background}-related examples from \cite{krishna-etal-2020-reformulating}. For each pair, the input is above.}
  \label{persona-examples}
\begin{tabularx}{\textwidth}{lX}
	\toprule[1pt]
	\multirow{4}{*}{\textbf{Gender and Age}} & \textbf{Male}: \textit{this is a spot that's making very solid food, with good quality product}  \\ 
	& \textbf{Female}: \textit{this is a cute spot that's making me very happy, with good quality product} \\ 
	\cmidrule(l){2-2}
	&\textbf{Young}: \textit{hilton worldwide starts its biggest global career event URL \#csr} \\ 
	& \textbf{Old}: \textit{hilton worldwide launches its largest global career event URL \#csr} \\
	\cmidrule(r){1-1}\cmidrule(l){2-2}
	\multirow{4}{*}{\textbf{Personality Traits}} & \textbf{Sweet}: \textit{That is a lovely sandwich} \\  
	& \textbf{Dramatic}: \textit{This sandwich looks so delicious! My goodness! } \\ 
	\cmidrule(l){2-2}
	& \textbf{Money-minded}: \textit{ I would totally pay \$100 for this plate} \\ 
	& \textbf{Optimistic}: \textit{It will taste positively wonderful} \\
	
		\cmidrule(r){1-1}\cmidrule(l){2-2}
	\multirow{4}{*}{\textbf{Background}} & \textbf{1990}: \textit{He was being
		terrorized into making a statement by the same means as the other so-called ``witnesses''} \\  
	& \textbf{1810}: \textit{Terror had been employed in the same manner with the other witnesses, to compel him to make
		a declaration.} \\ 
	\cmidrule(l){2-2}
	& \textbf{British}: \textit{As the BMA’s own study of alternative therapy showed, life is not as simple as that.} \\ 
	& \textbf{American}: \textit{As the F.D.A.’s own study of alternative therapy showed, life is not as simple as that.} \\	
	\bottomrule[1pt]
\end{tabularx}

\end{table}

\paragraph{Methods}
Though not focused on transfer, \cite{preotiuc2016discovering} were
the first to show that automatic paraphrases can exhibit the style of
writers of different ages and genders, by manipulating the lexical
choices made by a text generator.  A phrase-based translation model
learned that certain sequences of words are more typically used by
certain age/gender groups and, together with a language model of the
target demographics, it used such information to translate tweets from
one group to the other. Their translations turned out to perform
lexical substitution, a strategy that was more directly addressed by
others. \cite{reddy2016obfuscating}, for instance, performed
substitution in order to defeat a gender classifier. They did so with
the guidance of three metrics: one measured the association between
words and the target gender label, thus indicating the words to
replace to fool the classifier as well as possible substitutes;
another quantified the semantic and syntactic similarity between the
words to be changed and such substitutes; and the last measured the
suitability of the latter in context.

A pitfall of such heuristics, noticed by the authors themselves, is
that style and content-bearing words are equal candidates
for the edit.  Some neural methods bypassed the issue with a similar
3-step procedure.  That is the case of
\cite{sudhakar-etal-2019-transforming}, who proposed a variation of
the pipeline in \cite{li-etal-2018-delete}. There, (1) only
style-bearing words are deleted upon the decision of a
\textsc{Bert}-based transformer, where an attention head encodes the
stylistic importance of each token in a sentence.  Next, (2) candidate
substitutes are retrieved: sentences from a target-style corpus are
extracted to minimize the distance between the content words of the
input and theirs.  Lastly, (3) the final output is generated with a
decoder-only transformer based on \textsc{Gpt}, having learned a
representation of both the content source words and the retrieved
attribute words.  It should be noted that this method was not
 designed to transfer genre-related attributes specifically (it
achieves different results when dealing with other styles). Also,
\cite{madaan-etal-2020-politeness} addressed \textit{gender} as an
ancillary task. They used a similar methodology (further discussed in
Section~\ref{nontargeted-registers} under \textit{Politeness}) that
first identifies style at the word level, and then changes such words
in the output.

\cite{prabhumoye-etal-2018-style}, instead, separated content and
style at the level of the latent input representation, by employing
backtranslation as both a paraphrasing and an implicit disentangling
technique. Since machine translation systems are optimized for
adequacy and fluency, using them in a backtranslation framework can
produce paraphrases that are likely to satisfy at least two style
transfer desiderata (content preservation and naturalness). To change
the input attribute and comply with the third criterion, the authors
hinged on the assumption that machine translation reduces the
stylistic properties of the input sentence, and produces an output in
which they are less distinguishable.  With this rationale, a sentence
in the source language was translated into a pivot language; encoding
the latter in the backtranslation step then served to produce a
style-devoid representation, and the final decoding step conditioned
towards a specific gender attribute returned a stylized paraphrase.

Modelling content and style-related personal attributes separately is
in direct conflict with the finding by \cite{kang-etal-2019-male}, who
pointed out that features used for classifying styles are of both
types. As opposed to the studies mentioned above, this work
transferred multiple \textit{persona} styles in conjunction (e.g.,
\textit{education} and \textit{gender}), and did so with a
sequence-to-sequence model trained on a parallel dataset.  Similarly,
the proposal of \cite{liu2020revision} did not involve any
content-to-style separation. With the aim of making style transfer
controllable and interpretable, they devised a method based on a
variational auto-encoder that performs the task in different steps.
It revises the input texts in a continuous space using both gradient
information and style predictors, finding an output with the
target attribute in such a space.

\paragraph{Evaluation}
While \cite{reddy2016obfuscating} carried out a small preliminary
analysis, others assessed the quality of the outputs with (at least
some of) the three criteria, both with automatic and human-based
studies.  For instance, \cite{sudhakar-etal-2019-transforming}
evaluated the success of the transfer with a classifier, and
quantified fluency in terms of perplexity.  For meaning preservation,
other than \textsc{Bleu}
\citep{kang-etal-2019-male,sudhakar-etal-2019-transforming},
\textit{gender} transfer was evaluated automatically with metrics
based on n-gram overlaps (e.g., \textsc{Meteor}) and embedding-based
similarities between output and reference sentences (e.g., Embedding
Average similarity and Vector Extrema of
\cite{liu-etal-2016-evaluate}, as found in
\cite{kang-etal-2019-male}).

\cite{sudhakar-etal-2019-transforming} also explored \textsc{Gleu} as
a metric that better correlates with human judgments.  Initially a
measure for error correction, \textsc{Gleu} fits the task of style
transfer because it is capable of penalizing portions of texts changed
inappropriately while rewarding those successfully changed or
maintained.  As for human evaluation, the authors asked their raters
to judge the final output only with respect to fluency and meaning
preservation, considering the transfer of \textit{gender} a too
challenging dimension to rate.  Their judges also evaluated texts
devoid of style-related attributes.

\subsubsection{Personality traits}
The category of \textit{personality traits} contains variables describing
characteristics of people that are stable over time,
sometimes based on biological facts \citep{cattell1946personality}.
Studied at first in the field of psychology, personality traits have
also been approached in NLP
\citep[i.a.]{plank-hovy-2015-personality,rangel2015overview}, as they
seem to correlate with specific linguistic features -- e.g., depressed
writers are more prone to using first-person pronouns and words with
negative valence \citep{rude2004language}. This has motivated research
to both recognize the authors' traits from their texts
\citep{celli2014workshop} and to infuse them within newly generated
text \citep{mairesse-walker-2011-controlling}.

Computational works typically leverage well-established schemas,
like the (highly debated) Myers-Briggs Type Indicators
\citep{myers2010gifts} and the established Big Five traits
\citep{john2008paradigm}. These turn out particularly useful because
they qualify people in terms of a handful of dimensions, either binary
(introvert-extrovert, intuitive-sensing, thinking-feeling,
judging-perceiving) or not (openness to experience, conscientiousness,
extraversion, agreeableness and neuroticism).

Accordingly, a style transfer framework would change the attribute
value along such dimensions.  Some human-produced examples are the
switch from the sweet to dramatic type of personality and the transfer
money-minded to optimistic in Table~\ref{persona-examples} (note that
not all attributes addressed in style transfer are equally accepted in
psychology).  More precisely, each dimension represents a different
personality-related style, and this makes traits particularly
difficult to transfer: the same author can be defined by a certain
amount of all traits, while many other styles only have one dimension
(e.g., the dimension of polarity for sentiment), with the two extreme
attributes being mutually exclusive (i.e., a sentence is either
positively polarized or has a negative valence).

The ability to transfer \textit{personality traits} brings clear
advantages. For instance, the idea that different profiles associate
to different consumer behaviours
\citep{foxall1988personality,gohary2014personality} may be exploited
to automatically tailor products on the needs of buyers;
personification algorithms could also improve health care services,
such that chatbots communicate sensitive information in a more
human-like manner, with a defined personality, fitting that of the
patients; further, they can be leveraged in the creation of virtual
characters.

\paragraph{Data} So far, this task explored the collection of image
captions crowdsourced by \cite{shuster2019engaging}, who asked
annotators to produce a comment for a given image which would evoke a
given personality trait.  Their dataset
\textsc{Personality-Captions}\footnote{\url{http://parl.ai/projects/personality_captions}}
contains 241,858 instances and spans across 215 personality types
(e.g., sweet, arrogant, sentimental, argumentative, charming). Note
that these variables do not exactly correspond to personality traits
established in psychology.  As an alternative, one could exploit the
corpus made available by \cite{oraby-etal-2018-controlling},
synthesized with a statistical generator.  It spans 88k meaning
representations of utterances in the restaurant domain and matched
reference outputs which display the Big Five personality traits of
extraversion, agreeableness, disagreeableness, conscientiousness and
unconsciousness.\footnote{\url{https://nlds.soe.ucsc.edu/stylistic-variation-nlg}}

\paragraph{Methods} \cite{cheng-etal-2020-improving} provided evidence
that the disentanglement between the content of a text and the
authors' personality (where personalities are categorical variables)
can take place. Observing that such a disentanglement is in fact arduous
to obtain, they proposed a framework based on information theory.
Specifically, they quantified the style-content dependence via mutual
information, i.e., a metric indicating how dependent two random
variables are, in this case measuring the degree to which the learned
representations are entangled. Hence, they defined the objective of
minimizing the mutual information upper bound (to represent style and
content into two independent spaces) while maximizing their mutual
information with respect to the input (to make the two types of
embeddings maximally representative of the original text).

Without complying with any psychological models,
\cite{bujnowski-etal-2020-empirical} addressed a task that could
belong to this node in our hierarchy. Neutral sentences were
transferred into ``cute'' ones, i.e., excited, positive and
slangy. For that, they trained a multilingual transformer on two
parallel datasets, one containing paired mono-style paraphrases and
the other containing stylized rewritings, for it to simultaneously
learn to paraphrase and apply the transfer.

\paragraph{Evaluation} Other than typical measures for style (i.e.,
style classifiers' accuracy) and content (\textsc{Bleu}),
\cite{cheng-etal-2020-improving} considered generation quality, i.e.,
corpus-level \textsc{Bleu} between the generated sentence and the
testing data, as well as the geometric mean of these three for an
overall evaluation of their system.

\subsubsection{Background}
Our last \textit{unintended} style of \textit{persona} is the
background of writers.  Vocabulary choices, grammatical and spelling
mistakes, and eventual mixtures of dialect and standard language
expose how literate the language user is
\citep{bloomfield1927literate}; dialect itself, or vernacular
varieties, marked by traits like copula presence/absence, verb
(un)inflection, use of tense
\citep{green1998aspect,martin1998sentence} can give away the
geographical or ethnic provenance of the users
\citep{pennacchiotti2011machine}.  Further, because these grammatical
markers are prone to changing along with word meanings, language
carries evidence about the historical time at which it is uttered
\citep{aitchison1981language}.

In this research streamline are style transfer works leveraging
the idea that there is a ``style of the time''
\citep{hughes2012quantitative}: they performed diachronic linguistic
variations, thus taking timespans as a transfer dimension (e.g.,
\cite{krishna-etal-2020-reformulating} transferred among the
1810--1830, 1890--1910, 1990--2010 attributes).  Others applied changes
between English varieties, for instance switching from British to
American English \citep{lee-etal-2019-neural}, as well as varieties
linked to ethnicity, like English Tweets to African American
English Tweets and vice versa \citep{krishna-etal-2020-reformulating},
or did the transfer between education levels
\citep{kang-etal-2019-male}.

The following are example outputs of these tasks, from
\cite{krishna-etal-2020-reformulating}: ``\textit{He was being
  terrorized into making a statement by the same means as the other
  so-called ``witnesses''.}'' (1990) $\rightarrow$ ``\textit{Terror had
  been employed in the same manner with the other witnesses, to compel
  him to make a declaration.} ''  (1810); ``\textit{As the BMA’s own
  study of alternative therapy showed, life is not as simple as
  that.}''  (British) $\rightarrow$ ``\textit{As the F.D.A.’s own
  study of alternative therapy showed, life is not as simple as
  that.}'' (American).

Such variations could be applied in real-world scenarios in order to
adjust the level of literacy of texts, making them accessible for all
readers or better resonating with the culture of a specific
audience.  Future research could proceed into more diverse
background-related styles, such as those which are not shared by all
writers at a given time or in a specific culture, but which pertain to
the private life of subsets of them.  For instance, considering
hobbies as a regular activity that shapes how people talk, at least
for some types of content, one could rephrase the same message in
different ways to better fit the communication with, say, an
enthusiast of plants, or rather with an addressee who is into book
collecting.

\paragraph{Data} Sources that have been used for English varieties are
the New York Times and the British National Corpus for English
\citep{lee-etal-2019-neural}.  \cite{krishna-etal-2020-reformulating}
employed the corpus of \cite{blodgett-etal-2016-demographic}
containing African American Tweets, and included this dialectal
information in their own
dataset\footnote{\url{http://style.cs.umass.edu}}; as for the
diachronic variations that they considered, texts came from the Corpus
of Historical American English \citep{davies2012expanding}.  Also the
\textsc{Pastel} corpus compiled by \cite{kang-etal-2019-male} contains
ethnic information, which covers some fine-grained labels, like
Hispanic/Latino, Middle Eastern, Caucasian and Pacific Islander. Their
resource includes data about the education of the annotators involved
in the data creation process, from unschooled individuals
to PhD holders.

\paragraph{Methods}
\cite{logeswaran2018content} followed the line of thought that
addresses content preservation and attribute transfer with separate
losses. They employed an adversarial term to discourage style
preservation, and an auto-reconstruction and a backtranslation term to
produce content-compatible outputs. Noticing that the
auto-reconstruction and backtranslation losses supported the models in
copying much of the input, they overcame the issue by interpolating
the latent representations of the input and of the generated
sentences.

Other methods used for this style are not based on disentanglement
techniques \citep[e.g.,][]{kang-etal-2019-male}.  Among those is the
proposal of \cite{lee-etal-2019-neural}, who worked under the
assumption that the source attribute is a noisy version of the target one,
and in that sense, style transfer is a backtranslation task: their
models translated from a ``clean'' input text to their noisy
counterpart, and then denoised it towards the target.
\cite{krishna-etal-2020-reformulating} fine-tuned pretrained language
models on automatically generated paraphrases.  They created a
pseudo-parallel corpus of stylized-to-neutral pairs and trained
different paraphrasing models in an ``inverse'' way, that is, each of
them learns to recover a stylistic attribute by reconstructing the input
from the artificially-created and style-devoid paraphrases. Hence, at
testing time, different paraphrasers transferred different
attributes (given a target attribute, the model trained to reconstruct
it was applied).

\paragraph{Evaluation}
\cite{krishna-etal-2020-reformulating} proposed some variations on the
typical measures for evaluation, hinging on an extensive survey of
evaluation practices. As for content preservation, they moved away
from n-gram overlap measures like \textsc{Bleu} which both disfavors
diversity in the output and does not highlight style-relevant words
over the others. Instead, they automatically assessed content with the
subword embedding-based model by \cite{wieting-gimpel-2018-paranmt}.
With respect to fluency, they noticed that perplexity might
misrepresent the quality of texts because it can turn out low for
sentences simply containing common words. To bypass this problem, they
exploited the accuracy of a Ro\textsc{Bert}a classifier trained on a
corpus that contains sentences judged for their grammatical
acceptability. Moreover, they jointly optimized automatic metrics by
combining accuracy, fluency and similarity at the sentence level,
before averaging them at the corpus level.

\begin{table}
  \centering\small
  \caption{Style transfer methods distributed across \textit{unintended dynamic}
    styles of our hierarchy (Time: \textit{writing time}, Subj.: \textit{subjective bias}).}
  \label{tab:unintendeddynamic}
  \begin{tabular}{l p{0.2\linewidth}p{0.2\linewidth}p{0.2\linewidth}p{0.2\linewidth}}
    \toprule[1pt]	
    & \multicolumn{1}{c}{Parallel} & \multicolumn{3}{c}{Non-parallel} \\ 
    \cmidrule(r){2-2}\cmidrule(lr){3-3}\cmidrule(lr){4-4}\cmidrule(l){5-5}
    & & Exp. Disent. & Imp. Disent. & No Disent.\\ 
    \cmidrule(l){3-3} \cmidrule(l){4-4} \cmidrule(l){5-5}
    \mrrt{1}{Time} & Kang \citeyear{kang-etal-2019-male} & \texttt{-{}-} &\texttt{-{}-} &  \texttt{-{}-}\\ 
    \cmidrule(r){2-2}\cmidrule(lr){3-3}\cmidrule(lr){4-4}\cmidrule(l){5-5}
    \mrrt{1}{Subj.}& Pryzant \citeyear{pryzant2020automatically}& \texttt{-{}-} & \texttt{-{}-} & \texttt{-{}-}\\[5pt]
    \bottomrule[1pt]	
  \end{tabular}
\end{table}

\subsection{Dynamic states}

In the group of \textit{dynamic} styles, we arrange a few states in
which writers find themselves in particular contexts. Rather than
proxies for stable behaviours or past experiences, they are
short-lived qualities, which sometimes arise just in response to a
cue. Many facts influencing language slip into this category and
represent an opportunity for future exploration.  Some of them are:
the activity performed while communicating (e.g., moving vs.\
standing); motivational factors that contribute to how people say the
things they say (e.g., hunger, satisfaction); positive and negative
moods, as they respectively induce more abstract, high-level
expressions littered with adjectives, and a more analytic style,
focused on detailed information that abounds with concrete verbs
\citep{beukeboom2006mood}; the type of communication medium, known to
translate into how language is used -- for instance, virtual exchanges
are fragmentary, have specialized typography, and lack linearity
\citep{ferris2002writing}.

Another ignored but promising avenue is the transfer of
authenticity. Authenticity is a dynamic state transversing all the
styles we discussed so far, and at the same time defining a style on
its own.  In the broader sense, it is related to an idea of truth
\citep{authenticity-newman}, as it regards those qualities of texts
which allow to identify their author correctly: this is the type of
authenticity underlying the other \textit{unintended} leaves, i.e.,
the assumption that writers are spontaneous and do not mask nor alter
their personal styles.  Besides, a puzzling direction could be that of
``values'' or ``expressive authenticity''
\citep{authenticity-newman}. Writers may be more or less genuinely
committed to the content they convey. Authenticity in the sense of
sincerity would be the correspondence between people's internal states
and their external expressions, with a lack of authenticity resulting
in a lie. The binomial authentic-deceptive fits style transfer: all
content things being equal, what gives a lie away is its linguistic
style \citep{newman2003lying}. Therefore, an authenticity-aware style
transfer tool could help understand deceptive communication, or
directly unveil it.  Yet, the transfer between authenticity attributes
appears puzzling because successful liars are those who shape their
content in a style that seems convincing and trustworthy
\citep{friedman1990language}.

Below are the dynamic states that, to the best of our knowledge, are
the only ones present in the style transfer literature (they are
visualized in Table~\ref{tab:unintendeddynamic}, with some
corresponding examples in Table~\ref{tab:unintendedexamples}).

\subsubsection{Writing time} 
An instance of \textit{dynamic states}-related styles 
in the literature is the time at which writers produce an
utterance.  Information revolving around the writing time of texts was
collected by \cite{kang-etal-2019-male}, and is contained in their
\textsc{Pastel} corpus.  The authors considered daily time spans such
as Night and Afternoon, that represent the stylistic attributes to
transfer in text.  These attributes were tackled with the methods
discussed above, under \textit{persona} and \textit{background} 
(the success of their transfer was evaluated with the same techniques).

\subsubsection{Subjective bias} 
Talking of subjectivity in language evokes the idea that words do not
mirror an external reality, but reflect it as is seen by the speakers
\citep{wierzbicka1988semantics}. In this sense, language has the power
to expose personal bias.  NLP has risen to a collective endeavor to
mitigate the prejudices expressed by humans and reflected in the
computational representations of their texts
\citep{bolukbasi2016man,zhao-etal-2018-learning}. For its part, style
transfer has surged to the challenge of debiasing language by directly
operating on the texts themselves.

Although bias comes in many forms (e.g., stereotypes harmful
to specific people or groups of people), only one clear-cut definition
has been assumed for conditional text rewriting: bias as a form of
inappropriate subjectivity, emerging when personal assessment should
be obfuscated as much as possible. That is the case with
encyclopedias and textbooks whose authors are required to suppress
their own worldviews.  An author's personal framing, however, is not
always communicated openly. This is exemplified by the sentence
``\textit{John McCain exposed as an unprincipled politician}'',
reported in the only style transfer work on this topic
\citep{pryzant2020automatically}. Here, the bias would emerge from the
word ``\textit{exposed}'', a factive verb presupposing the truth of
its object.  The goal of style transfer is to move the text towards a
more neutral rendering, like one containing the verb
``\textit{described}''.

Bias (and the choice of terms that reinforce it) can operate beyond
the conscious level \citep{chopik2017age}.  Further, circumventing
one's skewed viewpoints seems to take an expert effort -- as suggested
by the analysis of \cite{pryzant2020automatically} on their own
corpus, senior Wikipedia revisors are more likely to
neutralize texts than less experienced peers.  Therefore, we collocate
the style \textit{subjective bias} under the \textit{unintended}
group, and specifically, as a division of \textit{dynamic states}
because prior judgments are open to reconsideration.

\paragraph{Data} \cite{pryzant2020automatically} released a
corpus\footnote{\url{https://github.com/rpryzantneutralizing-bias}} of
aligned sentences, where each pair consists of a biased version and
its neutralized equivalent.  The texts are Wikipedia revisions
justified by a neutral point of view tag, comprising 180k pre and post
revision pairs.
	
\paragraph{Methods} With the goal of generating a text that is
neutralized, but otherwise similar in meaning to an input,
\cite{pryzant2020automatically} introduced two algorithms. One, more
open to being interpreted, has two components: a neural sequence
tagger that estimates the probability that a word in a sentence is
subjectively biased, and a machine translation-based step dedicated to
editing while being informed by probabilities about
subjectivity. The alternative approach directly performs the edit,
with \textsc{Bert} as an encoder and with an attentional
\textsc{Lstm} as a decoder leveraging a copy and coverage mechanisms.

\paragraph{Evaluation} The models' accuracy was equated to the
proportion of texts that reproduced the changes of editors. In the
human-based evaluation, the success of models was measured with the
help of English-speaking crowdworkers who passed preliminary
tests proving their ability to identify subjective bias.

\begin{table}
	\centering\small
	\caption{Examples of \textit{dynamic states}, namely \textit{writing time} (from  \citet{kang-etal-2019-male}) and \textit{subjective bias} (taken from \citet{pryzant2020automatically}). Note that the former is transferred in combination with other styles (i.e., \textit{background}).}
	\label{tab:unintendedexamples}
	\begin{tabular*}{\textwidth}{ll}
		\toprule[1pt]
		\multirow{2}{*}{\textbf{Writing time}} & \textbf{Morning}: \textit{the flowers were in full bloom.}  \\ 
		& \textbf{Afternoon}: \textit{Tulips are one of the magnificent varieties of flowers.} \\ 
		\cmidrule(l){1-1}\cmidrule(l){2-2}
		\multirow{2}{*}{\textbf{Subjective Bias}} & \textbf{Biased}:  \textit{John McCain exposed as an unprincipled politician}\\  
		& \textbf{De-biased}:  \textit{John McCain described as an
			unprincipled politician} \\ 

		\bottomrule[1pt]
	\end{tabular*}
\end{table}

\section{Intended styles}
\label{sec:intended}
The second branch of the hierarchy
stems from the observation that some linguistic variations are
intentional. By \textit{intended} we refer to styles that people
modify contextually to the audience they address, their relationship,
their social status and the purpose of their communication. Due to a
complex interaction between individuals, society and contingent situations
\citep{brown1979speech}, it is not uncommon for speakers to
change their language as they change their role in everyday life,
alternating between non-occupational roles (stranger, friend),
professional positions (doctor, teacher) and kinship-related parts
(mother, sibling). Such variations occur as much in speech
conversations, as they do in texts \citep{biber2012register}. 

We split this group of styles into the \textit{targeted} and
\textit{non-targeted} subcategories.  The \textit{non-targeted} ones,
which are the non-evaluative (or non-aspect-based) styles, further
develop into the circumstantial and conventional nodes. While all
\textit{non-targeted} leaves can be associated with an idea of
linguistic variation, many of them are specifically closer to what
theoretical work calls ``registers'' and ``genres''.  Understanding
the characteristics of these two concepts would shed light on the
linguistic level at which the transfer of \textit{non-targeted}
features of text should operate; yet, there is no agreement on the
difference between genres and registers, and a precise indication of
what differentiates them from style is missing as well
\citep{biber1995dimensions}.  In our discussion, we follow
\cite{lee2001genres}: by genre we mean novels, poems, technical
manuals, and all such categories that group texts based on criteria
like intended audience or purpose of production; whereas registers are
linguistic varieties solicited by an interpersonal context, each of
which is functional to immediate use.  Therefore, we place the
culturally-recognized categories to which we can assign texts among
the \textit{conventional genres}, and we collocate linguistic patterns
that arise in specific situations among the \textit{circumstantial
  registers}.  Note that these two classes of styles are not mutually
exclusive: a formal register can be instantiated in an academic prose
as well as in a sonnet.

\begin{table}
  \caption{Literature on \textit{intended, targeted} styles (Sarc: \textit{sarcasm}; Political: \textit{political slant}) divided by
    method.}
  \label{tab:intendedtargeted}
  \centering\small
  \renewcommand{\arraystretch}{0.8}
    \begin{tabular}{r p{0.2\linewidth}p{0.2\linewidth}p{0.2\linewidth}p{0.2\linewidth}}
      \toprule[1pt]	
      & \multicolumn{1}{c}{Parallel} & \multicolumn{3}{c}{Non-parallel} \\ 
      \cmidrule(r){2-2}\cmidrule(lr){3-3}\cmidrule(lr){4-4}\cmidrule(l){5-5}
      && Exp.\ Disent. & Imp.\ Disent. & No Disent. \\ 
      \cmidrule(r){3-3}\cmidrule(l){4-4}\cmidrule(l){5-5} 
      \mrrt{5}{Emotions}
      & Chakrabarty \citeyear{chakrabarty-etal-2021-entrust}  
                                     & Helbig \citeyear{helbig-etal-2020-challenges}
      & Li \citeyear{li2020complementary} \par
        Nangi \citeyear{nangi-etal-2021-counterfactuals}
      &  
        Dryjanski \citeyear{dryjanski-et-al-2018}\par
        Lample \citeyear{lample2018multiple}\par
        Smith \citeyear{smith2019zero}\par
        Troiano\citeyear{Troiano2020}\par
        Riley \citeyear{riley-etal-2021-textsettr}
      \\
      \cmidrule(l){2-2}\cmidrule(l){3-3} \cmidrule(l){4-4}  \cmidrule(l){5-5} 
      \mrrt{10}{Sentiment}
      &
        Jin \citeyear{jin-etal-2019-imat}\par
        Cavalin \citeyear{cavalin-etal-2020-disjoint}
                                     &
                                       Guerini \citeyear{guerini2008valentino}\par
                                       Whitehead \citeyear{whitehead-cavedon-2010-generating}\par
                                       Li \citeyear{li-etal-2018-delete}\par
                                       Xu \citeyear{xu-etal-2018-unpaired}\par
                                       John \citeyear{john-etal-2019-disentangled}\par
                                       Sudhakar \citeyear{sudhakar-etal-2019-transforming}\par
                                       Wu \citeyear{wu-etal-2019-hierarchical-reinforced}\par
                                       Wu \citeyear{ijcai2019-maskinfill}\par
                                       Lee \citeyear{lee-2020-stable}\par
                                       Madaan \citeyear{madaan-etal-2020-politeness}\par
                                       Wen \citeyear{wen-etal-2020-decode}\par
                                       Malmi \citeyear{malmi-etal-2020-unsupervised}\par
                                       Reid \citeyear{reid-zhong-2021-lewis}\par
                                       Lee \citeyear{lee-etal-2021-enhancing}
      &
        Hu \citeyear{pmlr-v70-hu17e}\par
        Shen \citeyear{shen2017style}\par
        Fu \citeyear{fu2018}\par
        Liao \citeyear{liao-etal-2018-quase}\par
        Logeswaran \citeyear{logeswaran2018content}\par
        Prabhumoye \citeyear{prabhumoye-etal-2018-multilingual-back-translation}\par
        Prabhumoye \citeyear{prabhumoye-etal-2018-style}\par
        Singh \citeyear{singh2018sentiment}\par
        Tian \citeyear{tian2018structured}\par
        Yang \citeyear{yang2018unsupervised}\par
        Yang \citeyear{yang-etal-2019-specificity}\par
        Zhang \citeyear{zhang-etal-2018-learning}\par
        Zhao \citeyear{zhao-et-al-2018}\par
        Fang \citeyear{fang-etal-2019-implicit}\par
        Kruengkrai \citeyear{kruengkrai-2019-learning}\par
        Leeftink \citeyear{leeftink2019towards}\par
        Lai \citeyear{lai-etal-2019-multiple}\par
        Li \citeyear{li-etal-2019-domain}\par
        Tikhonov \citeyear{tikhonov-etal-2019-style}\par
        Yamshchikov \citeyear{yamshchikov-etal-2019-decomposing}\par
        Cheng \citeyear{cheng-etal-2020-improving}\par
        Li \citeyear{li2020complementary}\par
        Lin \citeyear{lin-etal-2020-learning}\par
        Nangi \citeyear{nangi-etal-2021-counterfactuals}
      & 
        Mueller \citeyear{pmlr-v70-mueller17a}\par
        Guu \citeyear{guu-etal-2018-generating}\par
        Dai \citeyear{dai-etal-2019-style}\par
        Gong \citeyear{gong-etal-2019-reinforcement}\par
        Lample \citeyear{lample2018multiple}\par
        Luo \citeyear{luo-etal-2019-towards}\par
        Luo \citeyear{Luo19DualRL}\par
        Pang \citeyear{pang-gimpel-2019-unsupervised}\par
        Wang \citeyear{wang_2019_latent_edit}\par
        Xu \citeyear{xu2019formality}\par
        Chawla \citeyear{chawla-yang-2020-semi}\par
        Gong \citeyear{gong2020rich}\par
        He \citeyear{he2020a}\par
        Li \citeyear{li-etal-2020-dgst}\par
        Liu \citeyear{liu2020revision}\par
        Mai \citeyear{mai-etal-2020-plug}\par
        Li \citeyear{li-etal-2021-text}\par
        Jafaritazehjani \citeyear{jafaritazehjani-etal-2020-style}\par
        Reif \citeyear{reif2021recipe}\par
        Riley \citeyear{riley-etal-2021-textsettr}
      \\
      \cmidrule(r){2-2}\cmidrule(r){3-3}\cmidrule(l){4-4}  \cmidrule(l){5-5} 
      \mrrt{2}{Sarc.}
      & Peled \citeyear{peled-reichart-2017-sarcasm}
                                     &\texttt{-{}-}
      & 
        Chakrabarty \citeyear{chakrabarty-etal-2020-r}\par
        Mishra \citeyear{mishra-etal-2019-modular}
                       &\texttt{-{}-}
      \\
      \cmidrule(r){2-2}\cmidrule(r){3-3}\cmidrule(l){4-4}  \cmidrule(l){5-5} 
      \mrrt{5}{Political}
      & Kang \citeyear{kang-etal-2019-male} 
                                     &
                                       Sudhakar \citeyear{sudhakar-etal-2019-transforming}\par
                                       Madaan \citeyear{madaan-etal-2020-politeness}\par
      &
        Chen \citeyear{chen2018learning}\par
        Prabhumoye \citeyear{prabhumoye-etal-2018-multilingual-back-translation} \par
        Prabhumoye \citeyear{prabhumoye-etal-2018-style}\par
        Tian \citeyear{tian2018structured}\par
        Nangi \citeyear{nangi-etal-2021-counterfactuals}
                       &\texttt{-{}-}
      \\[5pt]
      \bottomrule[1pt]	
      
    \end{tabular}
  \end{table}

\subsection{Targeted}
\label{targeted}

The presence of writers in language becomes particularly evident when
they assess a topic of discourse.  They applaud, disapprove and convey
values.  Communications of this type, which pervade social media, have
provided fertile ground for the growth and success of opinion mining
in NLP.  Opinion mining is concerned with the computational processing
of stances and emotions targeted towards entities, events, and their
properties \citep{hu2006opinion}. The same sort of information is the
bulk of study for the \textit{targeted} group in our hierarchy.  It is
``targeted'' because it reflects the relational nature of language,
often directed \textit{towards} an object
\citep{brentano2012psychology}: people state their stances or feelings
\textit{about} things or \textit{with respect to} properties.  Hence,
under this group are styles that pertain to the language of evaluations, 
like \textit{sarcasm} and \textit{emotions}.

The tasks of mining opinions and transferring them are kin in that
they use similar texts and observe similar phenomena.  Yet, they
differ in a crucial respect. Each of them looks for information at
different levels of granularity.  The former task not only recognizes
sentiment and opinions, but also extracts more structured information
such as the holder of the sentiment, the target and the aspects of the
target of an opinion \citep{liu2012survey}.  Instead, style transfer
only changes the subjective attitudes of writers.

Dealing with evaluations makes the transfer of \textit{targeted}
styles particularly troublesome.  To appreciate what is at stake here,
let us take an example that explicitly mentions an emotion,
``\textit{I'm happy for you}''. A style transfer task might generate a
paraphrase that expresses another state, for instance sadness, and
might do so by changing the emotion word into, e.g.,
``\textit{sad}''. Would such a modification change the stylistic attribute and
preserve the meaning of the input? This question urges attention: to
date, it is unclear whether this research line can aim at satisfying
the three transfer criteria, and therefore, whether it addresses style
transfer at all. Works in the field have not provided an answer, nor
have other studies in NLP offered key insights. As a matter of fact,
some of the styles at hand are cognitive concepts whose realization in
text is yet to be fully understood (are they content or style, or
both?). The problem arises not only with input texts containing
explicit markers of style (e.g., ``\textit{happy}'' for
emotions). Even when attitudes are expressed less directly in a
sentence (e.g., ``\textit{I managed to pass the exam}''), the issue of
shifting its stylistic attribute (and only its stylistic attribute)
remains.  Current studies solely suggest that the transfer is
effortless for some texts but not for others, and that it can occur
through various strategies -- not necessarily by swapping emotion
words \citep{helbig-etal-2020-challenges}.

An exhaustive overview of the relevant style transfer literature is
available in Table~\ref{tab:intendedtargeted}. Examples of the tasks
can be found in Table~\ref{tab:intendedexamples}.

\begin{table}
  \centering\small
  \caption{Examples of some \textit{intended (targeted)} styles, namely, \textit{emotion state}, \textit{sentiment} and \textit{sarcasm} coming from \cite{helbig-etal-2020-challenges},  \cite{li-etal-2018-delete}  and \cite{mishra-etal-2019-modular} respectively.}
  \label{tab:intendedexamples}
  \begin{tabular*}{\textwidth}{ll}
    \toprule[1pt]
    \multirow{2}{*}{\textbf{Emotion State}} & \textbf{Anger}: \textit{This soul-crushing drudgery plagues him}  \\ 
                                              & \textbf{Joy}: \textit{This fulfilling job motivates him} \\ 
    \cmidrule(l){1-1}\cmidrule(l){2-2}
    \multirow{2}{*}{\textbf{Sentiment}} & \textbf{Positive}:  \textit{great food but horrible staff and very very rude workers!}\\  
                                              & \textbf{Negative}:  \textit{great food, awesome staff, very personable and very efficient atmosphere!} \\ 
    \cmidrule(l){1-1}\cmidrule(l){2-2}
    \multirow{2}{*}{\textbf{Sarcasm}} & \textbf{Non-sarcastic}: \textit{Hate when the bus is late.} \\  
                                              & \textbf{Sarcastic}: \textit{Love when the bus is late.} \\  
    \bottomrule[1pt]
\end{tabular*}
\end{table}

\subsubsection{Emotion state}
Language carries a great deal of information about the writers'
emotions. These mental states have sparked research based on
classification
\cite[i.a.]{abdul-mageed-ungar-2017-emonet,Felbo2017,Schuff2017} and
generation
\cite[i.a.]{zhou-wang-2018-mojitalk,song-etal-2019-generating,huang-etal-2018-automatic},
but they have found little space in the study of transfer.  Indeed,
the multifaceted ways in which emotions are realized in language --
e.g., explicit mentions (``\textit{I am happy}''), implicit pointers
(``\textit{I was on cloud nine}''), descriptions of salient events
(``\textit{Cool, I passed the exam!}'') -- place this phenomenon at
the turn between \textit{what} is said and \textit{how} that is done
\citep{casel-etal-2021-emotion}.  As emphasized by the works on
emotion transfer, it is still debatable whether emotions can be
changed without distorting the semantic content of a text
\citep{helbig-etal-2020-challenges,Troiano2020}.

Assuming that emotions can be considered a style, their
transfer requires rewriting a source text such that the output conveys
the same message and a new emotional nuance.  Source and target
attribute labels can be borrowed from various traditions in
psychology. Past research in emotion analysis has used diverse
schemas, which describe emotions in multi-dimensional spaces
\citep{Buechel2017,Preotiuc2016} or in terms of some underlying
cognitive components \citep{Hofmann2020,Troiano2022,Stranisci2022}. On
the other hand, style transfer has only leveraged discrete
psychological models and has mapped between emotion names. Given a
source sentence like ``\textit{I was going to knock down a pedestrian
  with my car}'', that the writer associates to a fearful
circumstance, a joyful counterpart could be ``\textit{I wanted to
  overturn a pedestrian with my car}'' \citep{Troiano2020}. There are
also publications that do not follow any established emotion
schema. That is the case of \cite{lample2018multiple}, who performed
the transfer between two discrete writer's feelings, i.e., relaxed and
annoyed, and \cite{smith2019zero}, who preferred a richer set of
labels that mix different affective states and emotions. They put them
under the umbrella term of ``sentiment'', despite including more
fine-grained labels than polarity, such as the states of being
annoyed, ecstatic and frustrated.

\cite{chakrabarty-etal-2021-entrust} are an exception in this
panorama. Rather than focusing on the mental states per se, they
considered the \textit{appeal} to emotions, as an argumentative
strategy that makes texts persuasive to an audience.  These authors
leveraged the association between emotions and arguments, and rewrote
the latter to obtain more trustworthy variants (e.g., without
appealing to fear), thus paraphrasing sentences like ``\textit{At this
  dire moment, we all need to amplify our voices in defense of free
  speech.}'' as ``\textit{At this crucial moment, we all need to
  amplify our voices in support of free speech.}''.

It should be noted that discrete labels account for only part of
humans' emotion episodes. Other aspects are the strength of such
experiences, that is, their intensity \citep{sonnemans1994structure},
and the degree of arousal and dominance that they induce in the
concerned individuals \citep{mehrabian1996pleasure}. Style transfer
could be done in the future based on such models, for instance by
controlling not only what emotion is transferred but also to what
degree, similar to other generation studies that condition both the
emotion and the emotional strength of texts
\citep[i.a.]{ghosh-etal-2017-affect,goswamy-etal-2020-adapting}. This
might make the task of changing the emotion connotation more feasible
(e.g., the transfer might be possible between different
emotions but only for specific levels of intensity).

Since emotions pervade communication, there is
an unbounded number of applications where the related branch of style
transfer could be put to use -- from clinical to political
contexts. As an example, style transfer tools might support the
production of arguments by infusing a specific emotion in them, thus
enhancing their persuasive power; vice versa, they could be employed
to strip emotions away from existing arguments in order to isolate
their factual core. In the domain of education, they could give an
emotional slant to learning materials, to stimulate the learning
process \citep{zull2006key}.  Augmenting emotions or making them
explicit might also facilitate textual understanding for individuals
who struggle to interpret the expression of affective states, like
people on the autism spectrum, or suffering from alexithymia
\citep{poquerusse2018alexithymia}.  In commerce, they could be used to
rewrite trailers of books, movies or the presentation of any other
product, with a higher emotional impact. Lastly, any chatbot capable
of emotion transfer may adjust the affective connotation for the same
semantic gist depending on its users.

We recognize that placing \textit{emotion state} in the
\textit{intended} set of styles is a questionable choice. There are
some features of this mental fact that stir it towards the
\textit{unintended} side: people might not necessarily be aware that
emotions seep out of their written productions, neither do they
purposefully experience them (emotions are reactions to salient events
\citep{scherer2005emotions}). However, publications on emotion
transfer used data that humans consciously produced around
emotion-bearing events and impressions. Therefore, we include them in
the present category.

\paragraph{Data} There exists a comparably large set of emotion
corpora from various domains \citep{bostan-klinger-2018-analysis}, but
only a small subset has interested style transfer. Among them are
\textsc{Tec}, the corpus of Tweets from
\cite{mohammad_emotional_2012}, \textsc{Isear}, a collection of
descriptions of events that elicited emotional responses in their
experiencers \citep{scherer1994evidence}, and the
\textsc{EmpatheticDialogues}
dataset\footnote{\url{https://github.com/facebookresearch/EmpatheticDialogues}}
from \cite{rashkin-etal-2019-towards}, found in \cite{smith2019zero},
which encompasses a wide range of mental states.  A corpus that is not
dedicated to emotions but contains them as personality-related labels
is the \textsc{Personality-Caption} dataset
\citep{shuster2019engaging}, leveraged by \cite{li2020complementary}.

Concerning emotions and arguments,
\cite{chakrabarty-etal-2021-entrust} collected 301k textual instances
from the subreddit \textit{Change My View}, a forum for
persuasive discussions. They created a parallel corpus with the help
of a masked language model and a resource that labels nouns and
adjectives with their connotations, including the label
\textit{Emotion Association} \citep{allaway-mckeown-2021-unified}. The
authors matched the words in the arguments they gathered to the
entries in such an external dictionary. They masked those which are associated
with fear, trust, anticipation and joy, and constrained the
replacements proposed by the language model to have a different
emotional association than the original one.

A number of other emotion-related datasets could be adopted in
the future, which cover different textual domains and follow varied
psychological theories. Examples are the 10k English sentences of
\cite{Buechel2017} labelled with dimensional emotion information in the
Valence-Arousal-Dominance schema, the emotion-bearing dialogues of
\cite{Li2017}, and the literary texts made available by
\cite{Kim2017a} annotated both with discrete emotions and the
communication channels that express them (e.g., description of facial
expressions or body movements).
	
\paragraph{Methods} Being an under-explored task, emotion style
transfer was tackled by \cite{helbig-etal-2020-challenges} with a
pipeline transparent for investigation. Subsequent components (1)
identify textual portions to be changed, (2) find appropriate new
words to perform the lexical substitution, and (3) from the resulting
alternatives, pick one depending on its fluency, content preservation
and presence of a target attribute. Each step was instantiated with many
strategies, like (1) a rule-based identification of words vs.\ a
selection mechanism informed by the attention scores of an emotion
classifier, (2) retrieving new words from WordNet vs.\ leveraging the
similarity between input embeddings and those of possible substitutes,
(3) re-ranking the outputs with different weights for the three
transfer criteria.  The approach of \cite{dryjanski-et-al-2018}
used a neural network to perform phrase insertion, but it is
similar to that of \cite{helbig-etal-2020-challenges} in the idea
that specific portions of texts should be targeted for the change.

A filtering step based on re-ranking was also explored in
\cite{Troiano2020}, where style transfer is defined as a
backtranslation post-processing. The authors leveraged the idea
that neural machine translation systems maximize both the output
fluency and its faithfulness to the input (thus guaranteeing content
preservation and naturalness), and focused on their ability to
generate multiple and lexically diverse outputs as a way to promote
emotion variability.  Hence, with the help of an emotion classifier,
they re-ranked backtranslations with respect to their association with
the target emotion, and to perform the transfer, they selected the
text that best fulfilled such a requirement.  Similarly,
\cite{chakrabarty-etal-2021-entrust} generated multiple styled
rewritings, picking the one with the same meaning as the input -- in
their case, the one with the highest entailment relation to the
original text.  Their model was a fine-tuned \textsc{Bart} which
learned to generate texts on their parallel data (with the
artificially-created text being the input and the original argument
representing the target).  Generation was further controlled by
inserting a special separator token as a delimiter for the words that
the model needed to edit during fine-tuning.
	
Though not directly formulated in emotion-related terms, an effort of
emotion style transfer can be found in
\cite{nangi-etal-2021-counterfactuals}.  There, the produced
paraphrases display a different degree of excitement than the original
texts, mirroring the notion of arousal in the continuous models of
emotions. This paper aimed at gaining control over the strength of the
transfer by integrating counterfactual logic
in a generative model. With a series of losses to promote
disentanglement, their variational auto-encoder was trained to find
two separate embeddings for style and content. Counterfactuals came
into play in the form of a generation loss which guided the model to
find a new representation for the input attribute, specifically, a
representation that can push the prediction made by a style classifier
(given the style embeddings) towards the target attribute.

\paragraph{Evaluation} In a small-scale human evaluation,
\cite{helbig-etal-2020-challenges} defined a best-worst scaling task:
two annotators chose the best paraphrase for a given sentence, picking
among four alternatives generated from different pipeline configurations.
	
Consistent with the idea of making arguments more trustworthy,
\cite{chakrabarty-etal-2021-entrust} conducted a human evaluation in
which workers on Amazon Mechanical Turk rated arguments with respect to the
presence of fear, while simultaneously taking into consideration the
preservation of meaning (i.e., a trustworthy text would have been
penalized if it altered the input meaning).

\subsubsection{Sentiment}
Sentiment in NLP refers to the expression of a subjective and polarized
opinion \citep{liu2012sentiment}.  
A few works aimed at creating paraphrases that preserve the sentiment
but not the content of the input texts (e.g., ``\textit{It is sunny
  outside! Ugh, that means I must wear sunscreen.}''  $\rightarrow$
``\textit{It is rainy outside! Ugh, that means I must bring an
  umbrella.}'', as illustrated in \cite{feng-etal-2019-keep}).
Going in the opposite direction, style transfer rephrases an input text to alter
its polarity, which is either positive (``\textit{I was extremely
  excited in reading this book}''), negative (``\textit{The book was
  awful}''), neutral (``\textit{I've read the book}''), or is
characterized by some polarity gradation (``\textit{That's a quite nice
  book}'').

What a successful transfer of \textit{sentiment} should look like is
difficult to establish. The issue becomes clear by considering
examples of a transfer input and output, such as ``\textit{this
  restaurant has awesome pizza}'' and ``\textit{this restaurant has
  awful pizza}''.  On the one hand, these sentences are (intuitively)
stylistically the same -- which casts doubt on the status of sentiment
as a style.  On the other, they showcase that changing the polarity of
a text also affects its semantics. We stand by the view of
\cite{tikhonov2018wrong}, who denied that sentiment can be taken as a
linguistic dimension unrelated to content. They highlighted that if
sentiment is not independent of a text's semantics, but rather its
function, then the transfer attempt is contradictory (as content
changes, so does the ``sentiment style''). Consistent with this is an
observation of \cite{guu-etal-2018-generating}, who proposed a
generation system able to control for the attribute of a prototype
text with a series of edits.  With their model having to distort the
meaning of the prototype as little as possible, they noticed that an
edit like ``\textit{my son hated the delicious pizza}'' for the
prototype ``\textit{my son enjoyed the delicious pizza}'' would miss
the goal of content preservation.  To overcome this problem,
\cite{prabhumoye-etal-2018-style} relaxed the condition of keeping the
content untouched in favor of maintaining \textit{intent}, or the
purpose for which a text was produced (e.g., to move a critique).

Nevertheless, transferring \textit{sentiment} represents today a
hallmark for most of the state-of-the-art style transfer methods, due
to polarity being represented in many and relatively large datasets,
together with its possible industrial applications. A case in point
can be found in \cite{gatti2012creatively}, who created an application
that subverts the messages conveyed by posters by exaggerating their
sentiment, both positively and negatively. Moreover, sentiment is
relatively easy to recognize: given its polar nature, it has
distinctive linguistic markers, and it is often sufficient to perform
changes at this lexical level for the transfer to be considered
achieved \citep{fu-etal-2019-rethinking}. We hence include
\textit{sentiment} in our hierarchy, and we refer to it as a style for
convenience, to report on the massive amount of works that did so.

\paragraph{Data} A fair share of sentiment-polarized datasets consists
of mono-style resources. Commonly used are Yelp
reviews\footnote{\url{https://www.kaggle.com/yelp-dataset/yelp-dataset}},
Amazon
reviews\footnote{\url{https://cseweb.ucsd.edu//\~jmcauley/datasets.html\#amazon_reviews}}
and \textsc{Imdb}
reviews\footnote{\url{https://www.kaggle.com/lakshmi25npathi/imdb-dataset-of-50k-movie-reviews}}.
Arguing that superior performance is observed for any
sequence-to-sequence task with parallel data,
\cite{cavalin-etal-2020-disjoint} employed a semantic similarity
measure to derive parallel data from (non-parallel) Amazon and Yelp
reviews. Also, \cite{jin-etal-2019-imat} and
\cite{kruengkrai-2019-learning} derived a pseudo-parallel corpus from
mono-style data by aligning semantically similar sentences from the
sides of the source and target attributes.  For a subset of the Yelp
reviews, they collected human-generated styled
variations.\footnote{\url{https://github.com/zhijing-jin/IMaT}}

\paragraph{Methods} Many approaches that attempted to obtain a
sentiment neutralized latent representation of the content
\cite[e.g.,][]{pmlr-v70-hu17e} employed methods like adversarial
training
\citep{shen2017style,fu2018,fang-etal-2019-implicit,zhao-et-al-2018,lin-etal-2020-learning},
and fed this latent representation into a decoder to generate content
with the desired polarity. Reinforcement learning-based methods
have been adopted for sentiment transfer as well, to bypass the dependency on
differentiable learning objectives like loss terms
\citep{gong-etal-2019-reinforcement,luo-etal-2019-towards,Luo19DualRL}.
In the cycled reinforcement learning approach of
\cite{xu-etal-2018-unpaired}, a ``neutralization'' module removed
sentiment from the semantic content of a sentence, and an
``emotionalization'' module introduced the style with the desired
attribute in the newly generated text.  A policy gradient-based method
rewarded the neutralization step using the quality of the generated
text from the emotionalization
phase.\footnote{\cite{xu-etal-2018-unpaired} use the terms
  ``sentiment'' and ``emotion'' interchangeably (their
  emotionalization module transfers in fact sentiment). Psychology, on
  the other hand, separates emotions from other affective states
  \citep{scherer2005emotions}.}

Explicit disentanglement by identifying and changing style markers has
been claimed effective in sentiment style transfer
\citep{guerini2008valentino,whitehead-cavedon-2010-generating,li-etal-2018-delete,xu-etal-2018-unpaired,sudhakar-etal-2019-transforming,ijcai2019-maskinfill,leeftink2019towards,lee-2020-stable,madaan-etal-2020-politeness,malmi-etal-2020-unsupervised},
because such markers are less subtle compared to those of other styles
(e.g., \textit{personality traits}).  Strategies 
designed to this end use frequency statistics-based methods
\citep{li-etal-2018-delete, madaan-etal-2020-politeness}, sentiment
lexica \citep{wen-etal-2020-decode}, techniques based on the attention
scores of a style classifier
\citep{xu-etal-2018-unpaired,zhang-etal-2018-learning,sudhakar-etal-2019-transforming,yang-etal-2019-specificity,reid-zhong-2021-lewis}
or a combination of them \citep{ijcai2019-maskinfill}. The
sentiment-devoid content is then used as a template to generate text
with the target sentiment. 
\cite{wu-etal-2019-hierarchical-reinforced} achieved this with
the contribution of two agents: one that iteratively proposes where
 the re-wordings should occur in the text, and another that performs such
local changes. In \cite{reid-zhong-2021-lewis}, concurrent edits
across multiple spans were made possible by generating a template with
the Levenshtein edit operations (e.g., insert, replace, delete) which
guided the transformation of the input text towards the desired attribute. 
	
As stated by \cite{yamshchikov-etal-2019-decomposing}, the fact that
content and style are hard to separate at the lexical level
does not undermine the possibility that they can be separated in their
latent representations -- with the quality of such disentanglement
depending on the used architecture.  The machine translation framework of
\cite{prabhumoye-etal-2018-style}, already described in relation to
\textit{genre} style transfer (see Section~\ref{persona}), aimed at
producing a style-devoid representation in the encoding step of the
backtranslation. Compared to them, \cite{john-etal-2019-disentangled}
pushed the disentanglement even further, by dividing such
representation in two separate components, that is, a space of
sentiment and a space for the content of the sentence (where the
content is defined with bag-of-words, style-neutral features).  For a
given input, an auto-encoder represented the content (but not the
style), which was then fed to the decoder, concatenated with an
embedding of the desired output attribute. This is similar to
\cite{liao-etal-2018-quase}, who used two encoders to model content
and target attribute (a value of the rating of sentences/reviews
representing polarity).  Claiming that the conditioning structure is
essential for the performance of a style transfer model,
\cite{lai-etal-2019-multiple} refrained from treating the target
attribute simply as part of the initial vector fed to the decoder. Instead, they
concatenated the style vector with the output of a Gated
Recurrent Unit \citep{chung2015gated} cell at each time step.
Style information was implicitly obfuscated at
the token level by \cite{lee-etal-2021-enhancing}
under the assumption that the alternative option of
explicit removal of tokens would result in an information loss.
They opted for an adversarial strategy, which reversed the attention
scores of a style discriminator to obtain a style-devoid content
representation, and they applied conditional layer normalization on this
representation, to adapt it to the target attribute distribution.

In opposition to typical disentanglement-based studies,
\cite{yang2018unsupervised} noticed that classifiers that guide the
decoding step towards the desired attribute can be insufficient (their
error signal is sometimes too weak to train the generator), and that
their presence in adversarial setups as discriminators can lead to
unstable optimization. To solve this problem, the authors moved to
language models as a different type of discriminator which overcomes
the need for adversarial training: a language model trained on the
target sentiment data would not only assign low probabilities to
outputs that do not contain the desired sentiment, but it would also
allow outcome introspection (which word is responsible for such low
probability?). In a similar vein, \cite{li2020complementary} proposed 
to gradually incorporate the style-conditional supervision signals in the
successive training iterations, as long as the output quality does not
degenerate.  While these studies focused on the semantics of the input
and the generated sentences, \cite{gong2020rich} advocated the need
for including the representation of their syntactic information in the
transfer process.  They encoded a sentence by considering dependency
trees (to capture word relations) and structured semantic information
(i.e., semantic roles) with the help of a Graph Neural Network
\citep{marcheggiani-titov-2017-encoding}, providing evidence that they
can help a model identify the core information to be preserved.
	
Many limitations of disentanglement were pointed out in other
sentiment-based style transfer studies (e.g., using fix-sized vectors
for the latent representations might fail to retain the rich semantic
information characterizing long texts), with some of them 
casting doubt on the feasibility of the style-to-content separation
\citep[e.g.,][]{jafaritazehjani-etal-2020-style}.  As an alternative to
the manipulation of latent representations, \cite{dai-etal-2019-style}
added a style embedding as an input to their transformer encoder,
while \cite{li-etal-2020-dgst} directly proposed a novel architecture
composed of two generators and no discriminator. They performed style
transfer with a sentence noisification approach: after introducing
noise to an input text, they found a number of 
variations, and used them to learn the transfer by having the model
reconstruct the original input attribute. The novel method proposed by
\cite{li-etal-2021-text}, which did not resort to disentanglement,
used a generative adversarial network and a style classifier to
regularize the distribution of latent representations from an
auto-encoder. Instead, in the generative framework that
\cite{guu-etal-2018-generating} presented, a sequence of revisions was produced
for some prototype sentences. First, they extracted a prototype from a
corpus, next, they sampled an edit vector encoding the edit to be
performed: both were fed into the neural editor to produce 1k
sequences, and the sequence with the highest likelihood to contain
the target attribute was selected.

According to \cite{li-etal-2019-domain}, a further problem that
researchers should consider is that leveraging data from various
domains might result in poor transfer performances. A
model learned on movie reviews might not be appropriate to transfer 
polarity on restaurant
reviews. Hence, they presented a domain adaptive 
approach which modifies sentiment in a domain-aware manner.	
Others focused on how to leverage pretrained text-to-text models.  For
instance, \cite{mai-etal-2020-plug} formulated a ``plug and play''
approach that allows to employ pretrained auto-encoders,
and in which the transfer is learned within the latent space of the
auto-encoder itself (i.e., embedding-to-embedding). For few-shot style
transfer, \cite{riley-etal-2021-textsettr} leveraged the presumably
strong textual representations inherent to T5 \citep{2020t5}. Their
encoder-decoder model was trained to reconstruct a corrupted input.
Generation was
conditioned on a fixed-width style vector (similar to
\cite{lample2018multiple}) extracted from the preceding sentence,
assuming that style is a feature which spans over large
context windows. At inference time, the stylistic vector was
inferred from a set of style transfer exemplar pairs. Interestingly,
they demonstrated that a single model trained on generic web data can
transfer multiple styles, including dialect, emotiveness, formality, and
politeness.

\paragraph{Evaluation} As reported in the analysis of evaluation
practices in (sentiment) style transfer by \cite{mir-etal-2019-evaluating},
content preservation is typically evaluated in an automatic fashion
with metrics devised for machine translation, like \textsc{Bleu},
language models' perplexity over the generated texts serves as a
score for fluency, and sentiment classifiers quantify the transfer
strength (i.e.,  transfer accuracy would be the percentage of output
sentences that are classified as belonging to the target attribute).
To overcome the limitations of these metrics,
they suggested some alternative approaches.
In their view, transfer strength is quantified by the Earth
Mover's Distance: observing the cost of turning the style distribution of
the input into that of the output \citep{rubner1998metric}
would acknowledge the transfer even if the output did not
properly display the target attribute, but leaned toward it more than
the input. With respect to content preservation, the authors
experimented with two different settings, i.e., one in which the
style-related words coming from a style lexicon were removed and one
in which they were masked. Hence, they computed the Word Mover
Distance to quantify the distance between the input and output word
embeddings \citep{kusner2015word}.  Lastly, naturalness was assessed
via adversarial evaluation, with classifiers having to distinguish the
input texts written by humans from the output of the generation system.

\cite{mir-etal-2019-evaluating} also proposed some best practices with
respect to human evaluation, with the main idea that annotators should
be asked to perform pairwise comparisons: by rating the stylistic
difference between input and output, by comparing the two after
masking their style markers, and by choosing which of them is the most
natural.

\cite{yamshchikov-etal-2019-decomposing} leveraged human productions 
to propose some measures for the decomposition of textual information into content and
styles (they corroborated the idea that better decomposition leads to
better \textsc{Bleu} scores between output and human paraphrases). Yet
another strategy was put forward by
\cite{pang-gimpel-2019-unsupervised}.  They quantified content
preservation as the average of the cosine similarities over all
input/output sentence pairs, and observed perplexity using a language model
trained on concatenated source and target attribute datasets. Moreover, 
they introduced a strategy to adapt to the task at hand
which summarizes different metrics into a single score.

\subsubsection{Sarcasm}
Sarcasm represents a form of verbal irony \citep{Kreuz1989-KREHTB-3}.
\cite{alba2014evaluative} held that the usage of irony covers a
spectrum of evaluative purposes: to criticize (negative evaluation),
to praise (positive evaluation), or to express a neutral
stance. Sarcasm falls within the scope of negative evaluations because
it emerges as ``a sharp and often satirical or ironic utterance
designed to cut or give
pain''.\footnote{Merriam-Webster. (n.d.). Sarcasm. In
  Merriam-Webster.com dictionary. Retrieved October 15, 2021, from
  \url{https://www.merriam-webster.com/dictionary/sarcasm}.}  While
some studies hesitated in drawing an exact distinction between irony
and sarcasm \citep[i.a.]{UTSUMI20001777}, others did so and considered
it as a figure of speech with a specific target and a negative
connotation \citep[i.a.]{clift1999irony,alba2014evaluative}.

Being a figurative device, sarcasm is also characterized by a
contradiction between the literal and intended meaning of a statement.
It requires an understanding of the context in which an expression is
uttered, or a mutually shared assumption between the involved parties,
for the right interpretation to be grasped \citep{camp-sarcasm-2012}.
For example, the exclamation ``\textit{What a clever idea!}''
following a dull statement would be sarcastic, as the intended meaning
(i.e., the idea is unclever) conveys an unfavorable assessment, while
the utterance ``\textit{I now realize what a bad actor you are!}''
(after the actor got an award) would be ironic but devoid of any
sarcastic effect.  By insisting on the view of sarcasm in terms of
meaning inversion, \cite{camp-sarcasm-2012} actually identified
distinct subclasses of sarcasm -- depending on the illocutionary force
of the text, its evaluative attitude and its propositional content.

Most computational studies dedicated to such a phenomenon revolve around
classification. These works investigated the role of lexical
features, punctuation, emojis, sentence length, and sentiment, as
potential markers of sarcastic texts, and focused predominantly on 
social media communication
\citep[i.a.]{gonzalez-ibanez-etal-2011-identifying,
  barbieri-etal-2014-modelling, SULIS2016132,
  ling-sarcasm-irony-2016}.  
 There are also a few studies on sarcasm generation in style transfer.
 Even though they do not explicitly
formulate it as a transfer problem, they essentially use
an attribute-mapping principle, where a literal input is translated into a
sarcastic one or vice versa. \cite{peled-reichart-2017-sarcasm}
called this task ``sarcasm interpretation'', which consists in
interpreting and spelling out the actual intention of a sarcastic statement.

\paragraph{Data} A parallel sarcasm corpus, arguably the first of its
kind, was introduced by \cite{peled-reichart-2017-sarcasm}. These
authors crawled tweets with the hashtag ``\#sarcasm'' and used
crowdsourcing to generate non-sarcastic alternatives. The
resulting dataset includes 3k sarcastic tweets and five non-sarcastic
variants for each of them.

\paragraph{Methods} Driven by the idea that sarcastic statements have
strong polarized connotations, 
\cite{peled-reichart-2017-sarcasm} presented a machine
translation-based algorithm targeting textual sentiment to
``interpret'' sarcasm and turn a sarcastic expression into a literal
one.  \cite{mishra-etal-2019-modular} also leveraged the relation
between sarcasm and sentiment, and managed to introduce the
figurative-to-literal incongruity using an unsupervised approach with four
steps: the first neutralizes the input statement that expresses a negative
opinion, by removing the sentiment information with a classifier and a
self-attention based filtering -- e.g., ``\textit{Hate when the bus is
  late}'' $\rightarrow$ ``\textit{the bus is late}''; next, 
positive sentiment is injected into the neutralized sentence with a
sequence-to-sequence model trained on the neutralized and positive
sentence pairs -- e.g., ``\textit{the bus is late}'' $\rightarrow$
``\textit{love when the bus is late}''; the third step retrieves a
negative-situation phrase fitting the input from their own collection
of facts (e.g., \textit{canceled at short notice}, \textit{getting
  yelled at by people}) using an information retrieval system, with
the input acting as a query (e.g., ``\textit{waiting for bus}''); and
as a last step, the sarcastic statement is synthesized from the
positive keywords and negative situation phrases, with a reinforcement
reward.
	
\cite{chakrabarty-etal-2020-r} worked with similar assumptions. Their
system first reversed the valence of the input sentence by lexical
antonym replacement or negation removal -- e.g., ``\textit{zero
  visibility in fog makes driving difficult}'' $\rightarrow$
``\textit{zero visibility in fog makes driving easy}''.  Next, it
generated common sense knowledge using \textsc{Comet}
\citep{bosselut-etal-2019-comet}, a pretrained language model
fine-tuned on the ConceptNet knowledge graph \citep{speer-conceptnet},
by supplying keywords from the input and leveraging the
\textit{causes} relation - e.g., (\textit{zero}, \textit{visibility},
\textit{fog}, \textit{driving}, \textit{difficult}) $\rightarrow$
\textit{accident}. Lastly, this knowledge served to retrieve
candidate sentences, which were corrected for grammatical consistency
and ranked on a \textit{contradiction} score, similar to a natural
language inference problem.
	
\paragraph{Evaluation} Standard measures useful to quantify the
lexical closeness between a candidate and a reference (\textsc{Bleu},
\textsc{Rouge}, \textsc{Pinc} \citep{chen-dolan-2011-collecting}) were
reported for automatic evaluations \citep{peled-reichart-2017-sarcasm,
  chakrabarty-etal-2020-r}. In addition,
\cite{mishra-etal-2019-modular} presented a metric, the ``percentage
of length increment'', based on the assumption that sarcasm requires
more context than its literal counterpart.
	
As for the human evaluations, \cite{peled-reichart-2017-sarcasm}
collected ratings on the fluency and the adequacy of an interpretation,
\cite{mishra-etal-2019-modular} on the fluency and the relatedness to an input,
and \cite{chakrabarty-etal-2020-r} on the creativity, level of sarcasm,
humor and grammaticality. \cite{mishra-etal-2019-modular} also had
annotators label the sentiment of the transfer outputs.

\subsubsection{Political slant} 
Countless studies have been conducted on the relationship between
politics and language
\citep[i.a.]{orwell1946politics,spencer2018moral,shapiro1986language,habermas2006political}.
In the public sphere, verbal communication is strategic for political
manoeuvres.  It creates meanings around problems and events to favor
specific courses of action.  The idea that language puts things into a
compelling narrative for particular ideologies is one that
\cite{foucault1995order} developed further. He went as far as claiming
that it is language that constructs its users -- and not the users
constructing language, as the twentieth-century linguistics purported
(e.g., the Sapir–Whorf hypothesis in \cite{hoijer1954language}).
Indeed, every public debate inaugurates the use of some statements or
expressions: to accept one or the other is to embrace an ideology, to
present oneself as liberal or conservative, as an activist or a
separator, as a victim of the authority or a supporter
\citep{edelman1985political}.  These roles are the political
categories useful for style transfer.

NLP provides a parsimonious solution to address such a style (e.g., it
transfers broad attributes like ``democratic'' and
``republican''). However, it simplifies the complexity of political
language and the theories revolving around it.  The role of the
activist, of the authority, etc., not only guides people in opting for
certain linguistic variations but it imposes constraints upon
\textit{what} they say: a police chief, for instance, is called to
praise order over anarchy \citep{edelman1985political}.  This picture
suggests that content and political slant style are inextricably bound
together. Style transfer takes a different perspective and only taps
on the communicative attitudes of different political groups. A style
transfer result would look like the following: ``\textit{as a hoosier,
  i thank you, rep. visclosky.}''  (democratic) $\rightarrow$
``\textit{as a hoosier, i’m praying for you sir}'' (republican). That
is, moving from one attribute to the other does not necessarily imply
distorting an expressed political opinion, but generating one that
keeps the \textit{intent} of the original text (in this case, to thank
the senator) while changing the cues about the speaker's political
affiliation \citep{prabhumoye-etal-2018-style}. An exception to this
perspective is the work by \cite{chen2018learning}, who treated
political slant as a biased opinion to be altered (hence, we include
this style among those which are arguably closer to content, marked
with an asterisk in Figure~\ref{fig:hierarchy}).

Linguistics-oriented studies that investigated the rhetorical devices
of political communication
\citep{beard2000language,black2018analysing,rank1980,reisigl2008}
remain neglected in style transfer.  Yet, they provide fruitful
insights. Among others is the idea that debates, arguments, and
propaganda are filled with stylistic inventiveness to marshal support
and resonate with a large audience (e.g., political messages can be
disguised under some words that evoke objectivity -- like synonyms of
``essential'' or ``true'' \citep{edelman1985political}).  Future style
transfer studies could rewrite the language of promises as
ordinary language, devoid of sensationalisms and rhetoric intents, to
observe if the same message is conveyed, whether its persuasive
strength changes, and ultimately, to help people establish if certain
political claims are valid or are just embellished deceptions.

\paragraph{Data} Ideated to study responses to gender, the corpus of
\cite{voigt-etal-2018-rtgender} has also supported research in political
slant transfer.  Rt-Gender is a rich multi-genre dataset, with one subset
including Facebook posts from the members of the House and Senate in
the United States, and their top-level responses. The posts
include a label indicating if the Congressperson is affiliated with
the Republican or the Democratic party.  Posts and responses are
publicly available\footnote{\url{https://cs.cmu.edu/˜tsvetko/rtgender/}},
but all information that could identify the users was removed for
privacy.
	
The RtGender creators claimed that the dataset is controlled for
content by nature, because the members of the Congress discuss similar
topics. This represents an advantage for style transfer. According to
\cite{prabhumoye-etal-2018-style}, what reveals political slant are
both topic and sentiment, markedly different for the two affiliations,
like in the examples ``\textit{defund them all, especially when it
  comes to the illegal immigrants}'' and ``\textit{we need more strong
  voices like yours fighting for gun control}'' uttered by a
republican and a democratic, respectively. Researchers interested in
deepening such observation could make use of the dataset released by
\cite{mohammad2015sentiment}, as it includes electoral tweets
annotated for sentiment, emotion, purpose of the communication (e.g.,
to agree, disagree, support), and information related to some
rhetorical traits (e.g., whether it is sarcastic, humorous, or
exaggerated).
	
To address political opinions more specifically, \cite{chen2018learning}
collected 2196 pairs of news article headlines found on the platform
\textit{all-sides.com}, each of which is either left-oriented or
right-oriented, depending on the newspapers and portals where
they were published.

\paragraph{Methods} As for the stance flipping task addressed by
\cite{chen2018learning}, the authors started from the observation that
not all news headlines are biased enough for a model to learn the
task. Hence, they trained a generative model on the body of their
articles, whose sentences are not semantically paired. Hence, they
reproduced the cross-alignment setting proposed by
\cite{shen2017style} to transfer sentiment in the absence of parallel
data, training two encoders and two decoders (one for each
transfer direction).

No other method has been implemented exclusively for this task.  The
ones that have been applied are the backtranslation frameworks of
\cite{prabhumoye-etal-2018-style} and
\cite{prabhumoye-etal-2018-multilingual-back-translation} used for
\textit{sentiment} and \textit{gender} style transfer, which include a
separate decoder for each attribute (republican vs.\ democratic), and
the tag-and-generate pipeline proposed by
\cite{madaan-etal-2020-politeness} in the context of
\textit{politeness} (discussed in the next section).

\paragraph{Evaluation} \cite{prabhumoye-etal-2018-style} set up a
comparison task. Eleven annotators compared the models' outputs with
an input sentence. In line with the definition of the task, they had
to choose the paraphrase that maintained the intent of the source
sentence, while changing its slant. The annotators also had the
option to express no preference for any output. Their results showed
that most of the time people did not select any of the outputs,
suggesting that state-of-the-art systems still have a long way to go.
	
\cite{chen2018learning} framed the human evaluation task as one in
which annotators judged the degree to which two headlines have
opposite bias.
\cite{prabhumoye-etal-2018-multilingual-back-translation}, instead,
refrained from measuring the presence of the target attributes in their
human evaluation setting because judgments on political slants can be biased
by personal worldviews.

\begin{table}
  \caption{Literature on \textit{intended, non-targeted} styles
    corresponding to \textit{circumstantial registers} in our
    hierarchy (Polite: \textit{politeness}, Off.:
    \textit{offensiveness}, Lit: \textit{literality}), divided by method.}
  \label{tab:intendednontargetedcircumstancial}
\centering\small
  \begin{tabular}{r p{0.2\linewidth}p{0.2\linewidth}p{0.2\linewidth}p{0.2\linewidth}}
    \toprule[1pt]
    & \multicolumn{1}{c}{\textbf{Parallel}} & \multicolumn{3}{c}{\textbf{Non-parallel}} \\ 
    \cmidrule(r){2-2} \cmidrule(l){3-5} 
    && \textbf{Exp. Disent.} & \textbf{Imp. Disent.} & \textbf{No Disent.}\\ 
    \cmidrule(lr){3-3}\cmidrule(lr){4-4}\cmidrule(l){5-5}
    
    \mrrt{10}{Formality}
    & 
      Niu \citeyear{niu-etal-2018-multi-task-formality}\par
      Rao \citeyear{rao-tetreault-2018-dear}\par
      Etinger \citeyear{czeresnia-etinger-black-2019-formality}\par
      Ge \citeyear{ge-etal-2019-automatic}\par
      Jin \citeyear{jin-etal-2019-imat}\par
      Xu \citeyear{xu2019formality}\par
      Wag \citeyear{wang-etal-2019-harnessing}\par
      Chawla \citeyear{chawla-yang-2020-semi}\par
      Cheng \citeyear{cheng-etal-2020-contextual}\par
      Wang \citeyear{wang-etal-2020-formality}\par
      Zhang \citeyear{zhang-etal-2020-parallel}\par
      Briakou \citeyear{briakou-etal-2021-ola}\par
      Lai \citeyear{lai-etal-2021-thank}\par
      Yao \citeyear{yao-yu-2021-improving}
                                            &  \texttt{-{}-}
    &
      Li \citeyear{li-etal-2019-domain} \par
      Nangi \citeyear{nangi-etal-2021-counterfactuals}
    &  
      Gong \citeyear{gong-etal-2019-reinforcement}\par
      Luo \citeyear{Luo19DualRL}\par
      Shang \citeyear{shang-etal-2019-semi}\par
      He \citeyear{he2020a}\par
      Yang \citeyear{yang-klein-2021-fudge}\par
      Riley \citeyear{riley-etal-2021-textsettr}\par
      Reif \citeyear{reif2021recipe}
    \\ 
    \cmidrule(r){2-2}\cmidrule(lr){3-3}\cmidrule(lr){4-4}\cmidrule(l){5-5}
    \mrrt{2}{Polite}
    &\texttt{-{}-}
                                            &
                                              Madaan \citeyear{madaan-etal-2020-politeness}\par
                                              Reid \citeyear{reid-zhong-2021-lewis}
    &\texttt{-{}-}
                             &
                               Riley \citeyear{riley-etal-2021-textsettr}
    \\
    \cmidrule(r){2-2}\cmidrule(lr){3-3}\cmidrule(lr){4-4}\cmidrule(l){5-5}
    \mrrt{3}{Humor}
    &
      Weller \citeyear{weller-etal-2020-humor} 
                                            &
                                              Li \citeyear{li-etal-2018-delete}\par
                                              Sudhakar \citeyear{sudhakar-etal-2019-transforming}\par
    Madaan \citeyear{madaan-etal-2020-politeness}
    &
     Li \citeyear{li2020complementary} 
                             &
                               Zhu \citeyear{zhu2019neural}\par
                               Wang \citeyear{wang_2019_latent_edit}\par
                               Xu \citeyear{xu2019formality}
    \\
    \cmidrule(l){2-2} \cmidrule(l){3-3} \cmidrule(l){4-4}  \cmidrule(l){5-5} 
    \mrrt{2}{Off.}
    &
      Cheng \citeyear{cheng-etal-2020-contextual}
                                            &
                                              Su \citeyear{su-etal-2017-rephrasing}\par
                                              Tran \citeyear{tran-etal-2020-towards}
    &\texttt{-{}-}
                             & 
                              Nogueira \citeyear{nogueira-dos-santos-etal-2018-fighting}
    \\
    \cmidrule(r){2-2}\cmidrule(lr){3-3}\cmidrule(lr){4-4}\cmidrule(l){5-5}
    \mrrt{1}{Lit.}
    &
      Chakrabarty \citeyear{chakrabarty-etal-2020-generating}
                                            &\texttt{-{}-}
    &\texttt{-{}-}
                             &\texttt{-{}-}
    \\[5pt]
    \bottomrule[1pt]
  \end{tabular}
\end{table}

\subsection{Non-targeted: circumstantial r´egisters}
Registers are functional variations of a language \citep{halliday}.
Like the styles subsumed under the \textit{targeted} group,
registers have specific lexico-grammatical patterns -- e.g., the
distribution of pronouns and nouns differs between a casual
conversation and an official report \citep{biber2009register}. Unlike
the \textit{targeted} styles, they are not oriented towards an object,
but are general linguistic routines that mirror some behavioural
conventions. For example, in high-context cultures the discourse
becomes more courteous when addressing an older interlocutor or
someone perceived as higher in the social hierarchy. This is a hint of the
complexity of this family of styles: as noticed by \cite{hudson},
``one man’s dialect is another man's register''.

We show an overview of the \textit{intended, non-targeted} styles
regarding \textit{circumstantial registers} in
Table~\ref{tab:intendednontargetedcircumstancial}.  These types of
styles have also witnessed the definition of a new framework for style
transfer: according to \cite{cheng-etal-2020-contextual}, a reasonable
way of changing the characteristic attributes of a sentence is to take
into account the context in which the sentence occurs, and to produce
a stylized paraphrase that is coherent with it. The task of contextual
style transfer would reproduce more faithfully what happens in real
communications, where texts are never uttered out of context (e.g.,
sentences combine into paragraphs).

The readers might notice that some of these styles could also belong in the
\textit{targeted} category. As an example, humor can serve to express
an evaluative stance, similar to \textit{sarcasm}. However, such
styles are socially-motivated, and we consider them
\textit{registers} in that sense.

\begin{table}
  \centering\small
  \caption{Examples of style transfer on different
    \textit{circumstantial registers} -- \textit{formality,
      politeness, humor, figurative language} and
    \textit{offensiveness} -- taken from
    \cite{rao-tetreault-2018-dear, madaan-etal-2020-politeness,
      weller-etal-2020-humor, chakrabarty-etal-2020-generating,
      nogueira-dos-santos-etal-2018-fighting}, respectively.}
  \begin{tabular*}{\textwidth}{ll}
    \toprule[1pt]
    \multirow{2}{*}{\textbf{Formality}} & \textbf{Informal}: \textit{I’d say it is punk though.}  \\
                                        & \textbf{Formal}: \textit{However, I do believe it to be punk.} \\ 
    \cmidrule(r){1-1}\cmidrule(l){2-2}
    \multirow{2}{*}{\textbf{Politeness}} & \textbf{Impolite}: \textit{Send
                                           me the data.} \\  
                                        & \textbf{Polite}: \textit{Could you please send me the data.} \\ 
    \cmidrule(r){1-1}\cmidrule(l){2-2}
    \multirow{2}{*}{\textbf{Humor}} & \textbf{Non-humorous}:  \textit{Meet the wealthy donors pouring millions into the 2018 elections.}\\  
                                        & \textbf{Humorous}:  \textit{Meet the wealthy sadists pouring millions into the 2018
                                          elections} \\ 
    \cmidrule(r){1-1}\cmidrule(l){2-2}
    \multirow{2}{*}{\textbf{Figurative/Simile}} & \textbf{Literal}:  \textit{You just started staring off into
                                                  space and smiling dangerously}\\  
                                        & \textbf{Figurative}:  \textit{You just started staring off into space and smiling like a lunatic} \\
    \cmidrule(r){1-1}\cmidrule(l){2-2}
    \multirow{2}{*}{\textbf{Offensive}} & \textbf{Offensive}:  \textit{what a f**king circus this is.}\\  
                                        & \textbf{Non-offensive}:  \textit{what a big circus this is.} \\ 
    \bottomrule[1pt]
  \end{tabular*}
\end{table}

\subsubsection{Formality} The sentences ``\textit{His work was
  impressive and worthy of appreciation}'' and ``\textit{His work was
  damn good}'' show how texts can vary with respect to formality, an
important dimension of linguistic variation
\citep{heylighen1999formality} that characterizes the register of a
communication act. A formal text is explicit, accurate, and often
required to minimize misunderstandings, for instance in academic works
and legal documents.  On the other hand, an informal text has a
spontaneous and phatic nature. Being more relaxed, it can include
colloquial/slang terms, ellipses, contractions
\citep{heylighen1999formality, graesser2014coh, li2016new} and, on
social media, also emojis, acronyms, consecutive
punctuation (``\ldots'', ``!!!'').

The concept of (in)formality encompasses multiple features, like
seriousness--triviality, shared knowledge and familiarity
\citep{irvine1979formality, brown1979speech}, but style transfer
usually adopts the more straightforward dichotomy of formal vs.\
informal, often treated as endpoints of a continuum
\citep{graesser2014coh, heylighen1999formality}.

\paragraph{Data} Research on \textit{formality} transfer has been largely
supported by the Grammarly's Yahoo Answers Formality Corpus
(\textsc{Gyafc})\footnote{\url{https://github.com/raosudha89/GYAFC-corpus}}. Introduced
by \cite{rao-tetreault-2018-dear}, it contains around 110K
formal/informal sentence pairs, where the informal side was generated
via crowdsourcing. Next, the corpus curated by
\cite{briakou-etal-2021-ola},
\textsc{Xformal}\footnote{\url{https://github.com/Elbria/xformal-Fostyle transfer}},
extended formality data to multiple languages.  Like \textsc{Gyafc},
\textsc{Xformal} was built by extracting texts in the topic ``family
\& relationship'' from an existing corpus of Yahoo answers. Such
texts, which are in Brazilian Portuguese, Italian and French, were
characterized by an informal style. Crowdworkers on the platform 
Amazon Mechanical Turk\footnote{\url{https://www.mturk.com}}
provided multiple formal rewrites for each of them.

Depending on a single dataset might hinder the generalization
capability over unseen domains. Hence, by taking \textsc{Gyafc} as
ground truth, a few works based on data augmentation methods have
created and made available more style transfer instances.  The
formality classifier of \cite{xu2019formality} was trained on
\textsc{Gyafc} and made predictions on unlabelled texts; such
predictions were filtered for a threshold confidence score of 99.5\%.
\cite{czeresnia-etinger-black-2019-formality} augmented data with the
assumption that POS tags are representative of style-independent
semantics.  After training a classifier on \textsc{Gyafc}, they
applied it on a style-unlabelled corpus and created formal-informal
sentence pairs, by aligning sentences that become equal as soon as
their respective style markers are replaced with the corresponding POS
tags.

\cite{zhang-etal-2020-parallel} augmented approximately 4.9M sentence
pairs with three techniques: backtranslation, formality
discrimination, and multi-task transfer.  Backtranslation 
employed a sequence-to-sequence model trained on parallel data in the
formal to informal direction. It was then used to generate 1.6M
informal sentences, given formal ones coming from the ``entertainment
\& music'' and ``family \& relationships'' domains on Yahoo Answers
L6\footnote{\url{https://webscope.sandbox.yahoo.com}}. Also the
formality discrimination method exploited the observation that
machine-translated informal texts can be rendered more formal: a
number of informal English sentences from Yahoo Answers L6 were
translated to different pivot languages and then back, followed by a
discriminator with a predefined threshold that further filtered the
augmented data, giving a total of 1.5M pairs.  While these two
strategies used the newly generated texts to augment data, the
multi-task transfer method relied on sentence pairs 
annotated from previous tasks.  For that, style transfer was
formulated as a problem of Grammatical Error Correction under the
assumption that informal sentences are prone to containing grammatical
errors, character repetitions, spelling mistakes, unexpected
capitalization, and so on. Accordingly, to improve the transfer of
formality, they used the training data points for the Grammatical Error
Correction task as augmented texts, namely, the \textsc{Gec} data
\citep{mizumoto-etal-2011-mining,tajiri-etal-2012-tense} and the
\textsc{Nucle} corpus \citep{dahlmeier-etal-2013-building}.

Different from such resources, the Enron-Context corpus released by
\cite{cheng-etal-2020-contextual} contains paragraph-level data.
It includes emails randomly sampled from the
Enron dataset \citep{klimt-2004-enron}, in which 
sentences identified as informal by human annotators were rewritten in a
more formal manner.

\paragraph{Methods} The availability of a relatively large parallel
dataset has made \textit{formality} transfer a go-to task. 
\cite{rao-tetreault-2018-dear} spurred extensive research,
benchmarking the performance of phrase-based and neural machine
translation for this style. Following their work,
\cite{ge-etal-2019-automatic} performed style transfer on the
\textsc{Gyafc} corpus as a problem of grammatical error correction.

Others have moved the challenge of formality transfer into a
multi-lingual setting: \cite{niu-etal-2018-multi-task-formality} opted
for a multi-task learning approach to jointly perform monolingual
transfer and multilingual formality-sensitive machine translation;
\cite{briakou-etal-2021-ola} leveraged machine translation for
inter-language style transfer, learned both in a supervised and
unsupervised manner. The translation model of
\cite{yang-klein-2021-fudge} conditioned the output translation
towards formality with the help of future discriminators. These consisted in
some style predictors operating on an incomplete text sequence, which
inform as to whether the desired attribute will hold for the complete
text sequence, and can thus help adjust the generators' original
probabilities.

Many solutions were motivated by the need for massive amounts of
parallel data to prevent overfitting in machine translation
models. Among them are data augmentation attempts, like those by
\cite{czeresnia-etinger-black-2019-formality} and
\cite{zhang-etal-2020-parallel}.  The latter employed augmented texts
to pretrain models, but acknowledging that such texts are less than
perfect, the models were subsequently fine-tuned on the original
natural data.  \cite{xu2019formality} augmented data with a formality
classifier.  They trained a transformer model on a parallel corpus
with each instance prefixed with a token to indicate the direction of
transfer, such that a single model could go from formal to informal
and vice versa.

This was also achieved by \cite{wang-etal-2020-formality}, a work
belonging to the line of research that leverages pretrained language
models. A sequence-to-sequence model with a single encoder
captured the style-independent semantic representations with auxiliary
matching losses, and two decoders were dedicated to each target attribute,
jointly trained for bi-directional transfer.  In
\cite{chawla-yang-2020-semi}, a pretrained language model-based
discriminator helped to maximize the likelihood of the target attribute
being in the output, and a mutual information maximization loss
between input and output supported diversity in generation.
\cite{lai-etal-2021-thank} worked on the parallel texts from \textsc{Gyafc} to
fine-tune large pretrained language models, \textsc{Gpt}-2
\citep{Radford2019} and \textsc{Bart} \citep{lewis-etal-2020-bart} and
augmented them with rewarding strategies based on style discriminators
(targeting the transfer of the attributes) and \textsc{Bleu}
(targeting content preservation). They argued that pretrained models
contribute to better content preservation, even with limited training
data.

\cite{wang-etal-2019-harnessing} transformed informal sentences into
formal ones in a rule-based fashion, with some transfer rules 
incorporated in their language model.  The encoder was presented
with an input as a concatenation of the original informal sentence and
its formal revision to mitigate the consequent problem of noisy
parallel data. \cite{yao-yu-2021-improving} explored a similar
architecture. The encoder's input was created by concatenating the
original sentence and additional information, comprising a list of all
matched rules and the corresponding text alternatives, arranged as
tuples. Keeping all rules in the input allowed the model to identify
which ones to use dynamically.

Other approaches in \textit{formality} transfer that circumvented the
use of parallel corpora were reinforcement learning
\citep{xu2019formality} and probabilistic modelling
\citep{he2020a}. The work by \cite{cheng-etal-2020-contextual} stands
out in this panorama, in that it alters the formality of sentences
while simultaneously considering the topic coherence to the text
surrounding them. The context-aware model they proposed employs one
decoder that translates the joint features from two separate encoders
(which represent the main sentence and its contextual paragraph,
respectively).

\paragraph{Evaluation} Outside NLP, researchers have used measurements
based on diagnostic linguistic features to quantify the formality of
text. A popular measure is the F-score (formality score) which is
sensitive to the frequencies of different word classes in text,
ranging from articles and pronouns to adjectives and interjections
\citep{heylighen1999formality}. There also exists a composite score
that measures formality: Defined by \cite{graesser2014coh}, it is
based on five principal component dimensions of
Coh-Metrix\footnote{\url{http://www.cohmetrix.com/}}, and it takes
into account syntax, discourse, and goals of communication (e.g.,
syntactic simplicity, referential cohesion, word concreteness,
narrativity).

Style transfer studies have never opted for these measures.  
Indeed, while \cite{rao-tetreault-2018-dear} raised
the issue that the evaluation of style transfer (both human and
automatic) is in need for best practices, \textit{formality} transfer has
insisted on evaluating the transfer accuracy with a style classifier,
in line with other styles.

\subsubsection{Politeness}
Linguistic politeness reflects the evaluation of a social
context. Guided by a person's experience of social interactions
\citep{meier1995defining,holtgraves2013language} and socio-cultural
environment, politeness can uphold interpersonal relationships. Its
markers (e.g., ``\textit{please}'') affect how the speaker is
perceived: as a considerate individual or, on the contrary, as
discourteous \citep{meier1995defining}. Most studies in style transfer
focus on the broad attributes of ``polite'' and its opposite,
``impolite''.  However, according to some theories, the latter should
be explicitly distinguished from rudeness, which is always intentional
-- impoliteness can instead occur accidentally
\citep{segarra2007become, terkourafi2008toward}.

\textit{Politeness} transfer would change a formulation like
``\textit{You are wrong}'' into ``\textit{I think you might be
  mistaken}''.  To date, this style appears in a limited number of
publications, despite its link to formality as well as its potential
to assist automatic writing (e.g., to help non-native speakers produce
polite responses, as they might ignore some nuances in the target
language).

\paragraph{Data} The transfer task in
\cite{madaan-etal-2020-politeness} is restricted to action-derivatives
(e.g., ``\textit{Let's stay in touch}'') which are rewritten as polite
requests (e.g., ``\textit{Can you call me when you get back?}''). As
these constructs are frequent in official communication, the authors
built a politeness dataset starting from a collection of emails
exchanged within the Enron corporation, contained in the
Enron corpus \citep{klimt-2004-enron}.  With the application of some
filtering heuristics, 1.39 million sentences were gathered, annotated,
and filtered with a politeness score assigned by a classifier.  This
dataset is open
source\footnote{https://github.com/tag-and-generate/politeness-dataset}
and includes both the texts and the politeness scores.
		
Politeness labels are also present in the resource of
\cite{danescu2013computational}.  Included in the collection of styled
corpora from \cite{kang-hovy-2021-style}, it encompasses 10k requests
produced in the context of Wikipedia edits and other administrative
functions, as well as Stack Exchange, where requests are related to a
variety of topics. Their work focused on the politeness
markers of requests, characterized by strategies that
minimize imposition through indirect phrases (e.g., ``\textit{Could
  you please ...}'') or apologies (e.g., ``\textit{I'm sorry, but
  ...}'').

\paragraph{Method} The task was introduced by
\cite{madaan-etal-2020-politeness}. Observing the complex,
socio-cultural nature of politeness, these authors limited their study
to the use of formal language among North American English speakers.
They defined impoliteness as a lack of politeness markers, and adopted
a tag-and-generate approach. The linguistic realizations of the
potential marker positions were tagged in the source sentence and the
target attribute markers were then generated in such positions.

\cite{reid-zhong-2021-lewis}, who tested their method on the same dataset,
introduced an unsupervised explicit disentanglement procedure. First,
it transformed input texts into style-agnostic templates thanks
to the attention scores of a style classifier; then, it filled the tagged
positions in the templates using fine-tuned pretrained language
models. Unlike other infilling methods for style transfer
\citep{wang-etal-2019-harnessing, malmi-etal-2020-unsupervised},
theirs allowed concurrent edits over multiple textual spans.

\paragraph{Evaluation} For the automatic evaluation of transfer
accuracy, \cite{madaan-etal-2020-politeness} calculated the percentage
of generated sentences on which a classifier recognized the target attribute. 
For human evaluation, their annotators
judged the match with the target attribute on a 5-point scale.

\subsubsection{Humor}
Most theories on linguistic humor agree that this phenomenon arises from an
incongruity \citep[i.a.]{morreall1983taking, gruner1997game,
  rutter1997stand}.  Just like sarcasm, which assumes the existence of
two incompatible interpretations for the same text, humor
is given by the resolution of such interpretations
\citep{raskin-1979-semantic, attardo-raskin-1991-joke,
  ritchie-1999-incongruity-resolution}. In order to understand a joke,
the receiver needs to identify the punchline (i.e., an
incongruity) and then to resolve it by grasping its relationship with the main
context of utterance.  In communication, humor can serve as a tool to
relieve tension or lighten the mood, encourage solidarity, further
interactions within groups, and introduce new perspectives
\citep{meyer2000}. On the other hand, humor can
cause communication failures, if not perceived as intended.

A significant gap exists between computational studies of the style
\textit{humor}\footnote{\cite{amin-burghardt-2020-survey} presents a
  comprehensive overview of research in computational humor
  generation} and the theories underlying this concept, which remains
also overlooked in style transfer.  This style has an extremely
subjective nature and, unlike others, it is not characterized by a
defined pair of opposite attributes.  In fact, only a few researchers
considered the labels ``non-humorous'' and ``humorous''
\citep{weller-etal-2020-humor}, while the majority of them did the
transfer between the attributes ``humorous'', ``factual'' and
``romantic''
\citep{li-etal-2018-delete,sudhakar-etal-2019-transforming,wang_2019_latent_edit}.
This indicates a possible future line of research in which factuality
and romantic intimacy could stand as styles by themselves.

\paragraph{Data} \cite{weller-etal-2020-humor} used the Humicroedit
dataset \citep{hossain-2019-president}, a resource where
crowdworkers made single word edits to render
a regular news headline more humorous (e.g., ``\textit{Meet the wealthy
  \underline{donors} pouring millions into the 2018 elections}''
$\rightarrow$ ``\textit{Meet the wealthy \underline{sadists} pouring
  millions into the 2018 elections}''). Humicroedit contains around
15k edited headlines.
A similar corpus was presented by \cite{west-2019-satire}. It was
curated using an online game by asking participants to edit a humorous
headline and make it sound serious. Not evaluated to date, this
dataset could be useful for future research.  

Additional data can be found in the \textsc{Captions} corpus
\citep{caption-dataset}, which provides humorous captions describing
images.  Romantic and factual labels are also present as attributes
opposite to humorous. Instead, researchers who prefer to treat
``non-humorous'' as such opposite could make use of the Short Text
Corpus for Humor
Detection\footnote{\url{https://github.com/CrowdTruth/Short-Text-Corpus-For-Humor-Detection}}
and the Short Jokes
Dataset\footnote{\url{https://github.com/amoudgl/short-jokes-dataset}}
indicated by \cite{kang-hovy-2021-style}. These authors also provided
a small sample of texts (2k instances) which allow to consider
personal romanticism as a style on its own, with the two attributes
``romantic'' and ``non-romantic''.

\paragraph{Method} \cite{weller-etal-2020-humor} did an exploratory
investigation of the usability of the Humicroedit humor-based corpus
for style transfer purposes. A transformer-based sequence-to-sequence
model was trained for humor generation and a random POS tag 
replacement was taken as a baseline. 

As humor is not the main focus of the other works mentioned above, we refer
the reader to their respective discussions, under \textit{formality}
and \textit{sentiment}.
 
\paragraph{Evaluation} \cite{weller-etal-2020-humor} conducted a
human-based evaluation regarding the fluency and the level of humor of
texts, which were rated on a 5-point scale. The authors reported that
the manually edited sentences were considered more humorous than the
machine-generated ones, which in turn were better than random
replacements. This positively asserted the potential for the humor
generation task, highlighting at the same time the subjectivity of the
phenomenon in question.  A similar conclusion was drawn by
\cite{amin-burghardt-2020-survey}.  Focusing on the broader task of
humor generation, they analyzed possible evaluation approaches: human
ratings on a Likert scale for humorousness, human ratings on a
Likert scale for the likeness that a humorous text was written by a
human -- the soft Turing test as in \cite{yu-etal-2018-neural} -- and
``humorous frequency'' as the proportion of funny instances out of a
set of generated texts. All of them failed to present a criterion to
evaluate humor in text objectively.

\subsubsection{Offensiveness}
Under the expression ``offensive language'' we place facts related to
abusive language and harmful/hateful speech
\citep{nobata2016abusive,hateoffensive,schmidt-wiegand-2017-survey}.
Offensiveness is the negative extremity in the formality and
politeness spectrum, and it is usually resorted to with the intention
of attracting attention, offending\footnote{It should be noted that
  some studies
  \cite[e.g.,][]{waseem-hovy-2016-hateful,davidson2017automated} refrain from
  equating ``hate speech'' to language with offensive intentions,
  while others treat both as the same category to be detected
  \citep{Plazadelarco2021,Grimminger2021}.} or intimidating, and to
express anger, frustration and resentment
\citep{sue2007racial,popucsoi2018get}. Extensive research has stemmed
from this phenomenon, typically observed in the current social
media-communicating world, where any type of information can be
publicly discussed. While offensive behaviour detection
\citep[e.g.]{razavi-2010-offensive, davidson2017automated,
  founta-2019-abuse} has aimed at identifying and prohibiting offensive
material that exists online, style transfer studies like
\cite{su-etal-2017-rephrasing} and
\cite{nogueira-dos-santos-etal-2018-fighting} reformulated offensive
texts (e.g., ``\textit{That is f**king disgusting}'') in more gentle
terms (e.g., ``\textit{That is repulsive}''), or 
removed profanities \citep{tran-etal-2020-towards}.

Whether a text is derogatory or hurtful does not solely depend on the
presence of abusive words.  \cite{waseem-etal-2017-understanding}
brought up a typology of abusive language detection tasks which
clarifies that language can be belittling even without explicit slurs
or an explicit target person (or group of persons) to whom it is
directed.  Rhetorical questions and comparisons are only two examples
of how toxicity can emerge without swear words
\citep{van-aken-etal-2018-challenges}, but harm can find its way into
language with many more and more complex strategies -- e.g., jokes and
sarcasm \citep{wiegand-etal-2021-implicitly-abusive}.  While these
insights encourage researchers to make informed decisions as to the
most appropriate features to consider, depending on the type of
offensiveness in question, works in style transfer do not
necessarily consider all such factors.

In the future, studies related to this group of styles could address
the challenge of making texts not only less toxic but also more
inclusive of minorities.

\paragraph{Data} To overcome the lack of parallel data,
\cite{nogueira-dos-santos-etal-2018-fighting} opted to create a
non-parallel resource, and did so by employing the offensive language
and hate speech classifier from \cite{davidson2017automated}. The
final dataset contains approximately 2M and 7M sentences from Twitter
and Reddit, respectively, with the majority of instances being
non-offensive. Also \cite{cheng-etal-2020-contextual} created a parallel dataset of
offensive and non-offensive texts (the latter were 
crowdsourced by asking annotators to produce two non-offensive
alternatives for a given offensive input).
	
As for dictionary-based approaches, several open-access sources
are available. For instance, \cite{tran-etal-2020-towards} compiled a
vocabulary of offensive terms by crawling a list of more than 1k
English expressions made available by Luis von Ahn's research
group\footnote{\url{https://www.cs.cmu.edu/~biglou/resources/bad-words.txt}},
and an online platform that contains an ever-growing inventory of
profanities\footnote{\url{https://www.noswearing.com/}}.

\paragraph{Method} \cite{nogueira-dos-santos-etal-2018-fighting}
employed an encoder-decoder model with an attention mechanism. They
ensured output quality with a cycle consistency loss and the help of a
collaborative classifier providing signal about the effectiveness of
the transfer. Interestingly, they noted that their model was unable to
handle implicit offensive content (e.g., ordinarily inoffensive words
used offensively), indicating that offensiveness cannot always be
addressed at a lexical level by changing a few words.
	
Still, other researchers focused on the editing of offensive lexical
items.  For paraphrasing profane texts in Chinese,
\cite{su-etal-2017-rephrasing} manually devised a rule-based system,
equipped with an extensive set of profanity detection and paraphrasing
strategies (the rules were language-specific, hence the system is not
extendable to other languages). Similarly, \cite{tran-etal-2020-towards}
developed a transparent
modular pipeline around the idea that a text is offensive if
it contains profanity. The pipeline had different modules. First
comes the retrieval module: it extracts ten part-of-speech (POS) tag
sequences from a dataset of non-offensive texts, which are similar to
the POS sequence found in an offensive sentence.  Next is the
generation module, which creates non-offensive sentences by matching
the words from the input into possible positions in the generated POS
sequences, and then filling the unmatched positions with a pretrained
language model.  An edit step further corrects word order. The
selected output was the one with the best fluency, meaning preservation
and transfer -- which in this case corresponds to the absence of
profanities.

\paragraph{Evaluation} In addition to the regular metrics for content
preservation and fluency,
\cite{nogueira-dos-santos-etal-2018-fighting} reported the
classification accuracy using the classifier from
\cite{davidson2017automated}.

\subsubsection{Literality} Figurative language can be considered a
style because it embellishes things that could be said plainly (e.g.,
the statement ``\textit{He is a couch potato}'' creatively conveys
that a person is inactive).  It includes (but is not limited to)
metaphors, similes, idioms and oxymorons, each of which has 
distinctive features and requires different levels of cognitive
processing.  Expressions of this type have non-standard
meanings, which are somewhat derivative of their literal ones
\citep{paul-1970-figurative-language}.  This makes the distinction
between figurative and literal styles blurred.  Instead of
dichotomies, they represent different sites on a continuum
\citep{gibbs-2006-figurative}.

Computational studies on figurative language have favored metaphors
\citep{niculae-yaneva-2013-computational}, but the only form of
figurative expression that has entered the style transfer
literature is the simile, ``a figure of speech comparing two
essentially unlike things and often introduced by \textit{like} or
\textit{as}'' \citep{paul-1970-figurative-language}.  Similes are
figurative precisely because the items they compare are
essentially dissimilar from one another \citep{bredin-1998-simile},
unlike direct comparisons. Thus, ``\textit{She is like her mother}''
is not a simile, while ``\textit{Her smile is like sunshine}'' is.

\cite{chakrabarty-etal-2020-generating} were the first to frame \textit{simile}
generation as a style transfer task. Their goal was to replace the
literal expression (usually an adjective or an adverb) at the end of a
sentence with a figurative substitute (e.g., ``\textit{You just
  started staring off into space and smiling dangerously}''
$\rightarrow$ ``\textit{You just started staring off into space and
  smiling like a lunatic}'').

\paragraph{Data} A parallel dataset for similes with approximately 87k
sentences was created by \cite{chakrabarty-etal-2020-generating}. It
was built in an automatic manner, crawling self-labelled simile
expressions from Reddit via the comparative phrase \textit{like a}
(e.g., ``\textit{The boy was like an ox}''). The authors employed
\textsc{Comet} \citep{bosselut-etal-2019-comet}, a pretrained language
model fine-tuned on the ConceptNet \citep{speer-conceptnet} knowledge
graph, to replace the logical object of the comparison (here,
``\textit{an ox}'') with its shared property (here, ``\textit{being
  strong}'') in order to generate the parallel sentence (e.g.,
``\textit{The boy was strong}'').

\paragraph{Method} \cite{chakrabarty-etal-2020-generating} exploited a
simplified lexical structure followed by a simile, with clearly
defined roles for the lexical elements. In the example ``\textit{Her
  smile is like sunshine}'', the author intended to describe the
topic, \textit{her smile}, by comparing it to a logical object,
\textit{sunshine}, via a shared property, i.e., their brightness.
The parallel dataset they curated with distant supervision served to
fine-tune \textsc{Bart} \citep{lewis-etal-2020-bart}, a pretrained
language model that is a combination of bidirectional and
auto-regressive transformers.  They also conducted experiments with
baseline models based on conditional generation, metaphor masking and
retrieval using \textsc{Comet}
\citep{bosselut-etal-2019-comet}. Hence, they demonstrated that
incorporating structured common sense knowledge through \textsc{Comet}
is effective and can be employed in related creative text generation
tasks. The fine-tuned \textsc{Bart} model successfully generated
novel sentences and generalized over unseen properties.

\paragraph{Evaluation} For automatic evaluation,
\cite{chakrabarty-etal-2020-generating} reported \textsc{Bleu} after
removing the common prefix in the generated and reference
sentences. Moreover, they leveraged \textsc{Bert}Score
\citep{zhang-2020-bert-score}, a measure indicating the similarity
between candidate and reference sentences that uses contextual
embeddings, for the
contextual vectors of the logical object of the comparison
phrases. Human evaluation aimed at comparing the literal utterances
against six generated outputs, rated on a scale of 1-to-5 with respect
to creativity, overall quality, relevance of the comparison object in
portraying the shared property, and relevance of the suggested
comparison object in the given topic context.

\begin{table}
  \centering\small
  \caption{Literature on \textit{intended, non-targeted} styles corresponding to \textit{conventional genres}, divided by method.}
  \label{tab:intendednontargetedconventional}
    \begin{tabular}{r p{0.2\linewidth}p{0.2\linewidth}p{0.2\linewidth}p{0.2\linewidth}}
      \toprule[1pt]	
      & \multicolumn{1}{c}{Parallel} & \multicolumn{3}{c}{Non-parallel} \\ 
      \cmidrule(r){2-2} \cmidrule(l){3-5} 
      && Exp.\ Disent. & Imp.\ Disent. & No Disent.\\ 
      \cmidrule(r){3-3}\cmidrule(l){4-4}  \cmidrule(l){5-5} 
      \mrrt{5}{News} &  \texttt{-{}-}&\texttt{-{}-}
      &
        Romanov \citeyear{romanov-etal-2019-adversarial}\par
        Fu \citeyear{fu2018}
                       &
                         Gatti \citeyear{gatti2015slogans}\par
                         Gatti \citeyear{gatti2016heady}\par
                         Lee \citeyear{lee-etal-2019-neural}\par
                         Zhang \citeyear{zhang-etal-2018-shaped}\par
                         Chen \citeyear{chen2021controlled}
      \\
    \cmidrule(r){2-2}\cmidrule(lr){3-3}\cmidrule(lr){4-4}\cmidrule(l){5-5}
      \mrrt{1}{Tech.} & Cao \citeyear{cao-etal-2020-expertise} & \texttt{-{}-} & \texttt{-{}-} &  \texttt{-{}-}\\ 
    \cmidrule(r){2-2}\cmidrule(lr){3-3}\cmidrule(lr){4-4}\cmidrule(l){5-5}
      \mrrt{4}{Literature}
      &
        Xu \citeyear{xu-etal-2012-paraphrasing}\par
        Jhamtani \citeyear{jhamtani-etal-2017-shakespearizing}\par
        Carlson \citeyear{carlson2018evaluating}\par
        Bujnowski \citeyear{bujnowski-etal-2020-empirical}
                                     &
                                       Gero \citeyear{gero-etal-2019-low} 
      &
        Romanov \citeyear{romanov-etal-2019-adversarial}
                       &
                         Mueller \citeyear{pmlr-v70-mueller17a}\par
                         Pang \citeyear{pang-gimpel-2019-unsupervised}\par
                         Shang \citeyear{shang-etal-2019-semi}\par
                         Krishna \citeyear{krishna-etal-2020-reformulating}\par
                         He \citeyear{he2020a}
    \\ 
    \cmidrule(r){2-2}\cmidrule(lr){3-3}\cmidrule(lr){4-4}\cmidrule(l){5-5}
      \mrrt{2}{Lyrics} & \texttt{-{}-}& \texttt{-{}-}&\texttt{-{}-} &
                              Lee \citeyear{lee-etal-2019-neural}\par
                              Krishna \citeyear{krishna-etal-2020-reformulating}
      \\[5pt]	 
      \bottomrule[1pt]		
    \end{tabular}
\end{table}

\subsection{Non-targeted: conventional genres}
\label{nontargeted-registers}

Established textual varieties, like poems, newspaper articles and
academic productions flow into the \textit{conventional} category
(see an overview in
Table~\ref{tab:intendednontargetedconventional}). This family of
styles includes institutionalized types of communication, which are
encoded within one (or many) culture(s) \citep{biber1995dimensions}.
Hence, they follow some systematic norms, and for this reason they are 
different from \textit{circumstantial} styles, in which
linguistic choices are due to social and contingent situations. 

Different genres (henceforth, styles) are recognizable
by some markers that can be more or less explicit (e.g., \textit{the
  objective of this paper is...} vs.\ \textit{once upon a time...})
\citep{coutinho2009describe}. Scientific articles, for instance, put
constraints on one's vocabulary choices and syntactic structures, as opposed to literary
genres, which allow for freer linguistic constructions (e.g., including evaluative
adjectives, metaphors, etc.)
\citep{biber1995dimensions}.
Their transfer includes objectives
like the versification of a prose, the satirization of a novel, or the
simplification of technical manuals. Tasks with such kinds of
styles are appealing for end users -- turning poems into paraphrases
has the potential to support education and transforming existing news
headlines to produce catchier ones can be useful for
advertisement. They also bear a potential value from a 
theoretical perspective: style transfer can foster academic attempts to
describe what genre is, because manipulating markers offers different
conditions of investigation, and this might help explain how readers
decide about the membership of a text into a certain category.

\subsubsection{Forums/newspapers}
While the transfer of \textit{newspaper}-based attributes has taken a
number of forms, early attempts involved the concept of ``blending''.
Blending consists in rephrasing and incorporating a piece of text with
a secondary (arbitrary) idea, to produce an utterance that evokes not
only the original meaning but also the newly juxtaposed one.  For
instance, a given expression (a slogan like ``\textit{Make love not
  war}'', or a cliché, a song, a movie title) can be blended with the
daily news (e.g., the headline ``\textit{Women propose sex strike for
  peace}''), such that the result will contain a reference to both
(e.g., ``\textit{Make peace not war}'' \citep{gatti2015slogans}).
These initial works did not explicitly formulate the task as style
transfer, but as one where the stylistic attributes used to
communicate the news of the day are rendered more similar to a
well-known expression.

Without tapping on notions related to creativity,
\cite{lee-etal-2019-neural} addressed the problem of transferring the
stylistic features of forums to news (e.g., ``\textit{i guess you need to refer to
  bnet website then}'' $\rightarrow$ ``\textit{I guess you need to
  refer to the bnet website then}''), which in their view amounts to a
task of formality transfer and \cite{fu2018} ventured the goal of
scientific paper to newspaper title transfer (``\textit{an efficient
  and integrated algorithm for video enhancement in challenging
  lighting conditions}'' $\rightarrow$ ``\textit{an efficient and
  integrated algorithm, for video enhancement in challenging power
  worldwide}'').  The transfer was also made between the stylistic
attributes of different newspapers.  \cite{zhang-etal-2018-shaped}
showed that publishers can be taken proxies for style (e.g., the New
York Times\footnote{\url{https://www.nytimes.com}} has a different stylistic cipher
from the Associated Press\footnote{\url{https://apnews.com/}}) as they
tend to use different wording patterns.

\begin{table}
  \centering\small
  \caption{Examples of style transfer outputs on different
    \textit{conventional genres} of text -- \textit{forums \&
      newspapers, literature, technical language} and \textit{song
      lyrics} -- taken from \cite{chen2021controlled,
      xu-etal-2012-paraphrasing,
      cao-etal-2020-expertise,
      lee-etal-2019-neural}.}
  \label{genres-examples}
  \begin{tabular*}{\textwidth}{ll}
    \toprule[1pt]
    \multirow{4}{*}{\textbf{Newspapers}} 
    & \textbf{Economic Frame}: \textit{``It’s time for Congress to take action,'' says a spokesman for the bill’s}\\ 
    & \textit{sponsors, who want a flexible spending limit.}  \\
    & \textbf{Legality Frame}: \textit{``Illegal aliens’ is a growing problem in the country,'' says a spokesman}\\
    & \textit{for the measure’s sponsors} \\ 
    \cmidrule(r){1-1}\cmidrule(l){2-2}
    \multirow{2}{*}{\textbf{Literature}} & \textbf{Early Modern English}: \textit{I will bite thee by the ear for that jest} \\  
    & \textbf{Contemporary English}: \textit{I'll bite you by the ear for that joke} \\ 
    \cmidrule(r){1-1}\cmidrule(l){2-2}
    \multirow{2}{*}{\textbf{Technical Language}} & \textbf{Expert}:  \textit{Many cause dyspnea, pleuritic chest pain, or both.}\\ 
    & \textbf{Layman}: \textit{The most common symptoms, ..., are shortness of breath and chest pain.}\\ 
    \cmidrule(r){1-1}\cmidrule(l){2-2}
    \multirow{2}{*}{\textbf{Song Lyrics}} & \textbf{Hip-hop}:  \textit{Yo, where the hell you been?}\\  
    & \textbf{Pop}: \textit{Yo, where the hell are you?} \\ 
    \bottomrule[1pt]
  \end{tabular*}
\end{table}

Taking a different approach, a line of research addressed the problem
of ``reframing'' news. This type of conditioned paraphrasing consists
in changing the perspective from which a topic is conveyed
\citep{chen2021controlled}, for the audience to focus on some
of its aspects and prefer a particular interpretation. There, the
stylistic attributes of \textit{newspapers} are the frames that are
evoked by a piece of text (e.g., economics-, legality-related
frames). These can prompt two texts to have the same
denotation/reference but different connotations, which is the case for
``\textit{undocumented workers}'' and ``\textit{illegal aliens}''.
This task is similar to the argument rewriting discussed with respect
to \textit{emotional state}, it is close to \textit{sentiment} (as it
connects to rewriting with a more positive or negative presentation of
the topic) and it touches upon the notion of contextual style transfer
(discussed under \textit{formality}) because it needs to ensure that
an output sentence is coherent with the surrounding context. Some
examples are in Table~\ref{genres-examples}.

\paragraph{Data} A useful newspaper dataset for style transfer was
created by \cite{de-mattei-etal-2020-invisible}, even though their
work regarded style-aware generation rather than transfer. They
collected news that are lexically similar from two newspapers, a
subset of which are topic-aligned. \cite{gatti2016heady} used the news
of the day, extracted from the RSS feed of the New York Times and BBC
News, and \cite{lee-etal-2019-neural} resorted to articles from the
New York Times and comments from Reddit.

Another dataset dedicated to news articles is the Gigaword
corpus\footnote{\url{https://catalog.ldc.upenn.edu/LDC2011T07}}
\citep{parker2011english}.  This resource was acquired over several
years by the Linguistic Data Consortium, and it spans seven
international sources of English newswire (i.e., Agence France-Presse,
Associated Press Worldstream, Central News Agency of Taiwan, Los
Angeles Times/Washington Post Newswire Service, New York Times, Xinhua
News Agency, and Washington Post/Bloomberg Newswire Service).
\cite{fu2018} focused instead on news titles. They built a
dataset\footnote{\url{https://github.com/fuzhenxin/textstyletransferdata}}
of 108,503 titles belonging to the science and technology categories
and which come from the UC Irvine Machine Learning Repository
\citep{uci_repo}. As an attribute opposite to ``news'',
their corpus contains scientific-oriented language, specifically paper
titles crawled from academic websites.
	
The reframing study of \cite{chen2021controlled} made use of the
corpus published by \cite{card-etal-2015-media}. Encompassing more
than 35k news articles about death penalty, gun control, immigration,
same-sex marriage and tobacco, the corpus is annotated with 15 framing
dimensions (e.g., economics, morality, politics) developed by
\cite{boydstun2014tracking}.

\paragraph{Methods} \cite{gatti2015slogans} performed lexical
substitution by extracting keywords from the news and inserting them
in well-known expressions coming from slogans, movie titles, song
titles and clichés: after pairing the two data based on a similarity
measure, they used a dependency metrics to find the probability for the
words in the slogan of being replaced with the same part-of-speech
keywords from the news.
	
More recent neural attempts aimed at transferring news titles to
scientific paper titles. This was done by
\cite{romanov-etal-2019-adversarial}, who fit in the 
picture of disentanglement based on adversarial methods. They had an encoder produce a
continuous style vector and a meaning vector for a given input.
Compared to other adversarial approaches, these authors employed two
complementary forces. One was a discriminator that penalized the
encoder if the meaning embeddings still carried information about
style; the other was a motivator, and it pushed the encoder to
produce representations that facilitate the correct attribute
classification -- encouraging, rather than
penalizing, was proven to make the separation between the two types of
embeddings bolder.
	
Moving on to news reframing, \cite{chen2021controlled} characterized
the problem in the following terms: given three consecutive sentences
and a target frame, the middle sentence can be masked, and a new one
generated to fill in such blank, which contains the target frame
and links the preceding and follow up sentences coherently. The
authors trained one generation model for each frame, and experimented
with three strategies.  Namely, fine-tuning a sequence-to-sequence
model on a specific frame, including knowledge about named entities to
promote topic coherence, and adding examples in the training data
(the sentence to be generated has a different frame compared to
the surrounding ones).

\paragraph{Evaluation} \cite{de-mattei-etal-2020-invisible} put
forward the idea that \textit{news} styles are more difficult to judge
than others (e.g., sentiment), and that humans are not as reliable
judges of said styles as machines.  They proposed a framework
for the automatic evaluation of style-aware generation that seems
handy for style transfer as well. Their automatic classifier had to
distinguish the newspaper style of lexically aligned headlines: such
an alignment pushed the classifier to make decisions based on
stylistic information rather than content-related one.
	
With respect to human evaluation, \cite{gatti2015slogans} asked people
if an output headline was grammatically correct and if it could work
as a headline for a given article, while \cite{chen2021controlled}
conducted an extensive study in which they presented crowdworkers
with multiple reframings for an input text, which had to be
evaluated for their contextual coherence, topical
congruence, and presence of a given frame.

\subsubsection{Technical language} The curse of knowledge, an
expression introduced by \cite{camerer1989curse}, is a cognitive bias
that arises in communication, for instance between professionals in a
certain field and less expert people.  It can be observed when a
well-informed agent assumes understanding from less informed ones,
thus hampering a successful exchange of ideas.  Style transfer methods
can be applied to such situations to simplify language and mitigate
the lack of shared knowledge between the two parties.

The task of automatic rewriting to make texts more easily readable
(while securing their relevant information) has sparked wide attention in
NLP
\citep{wubben-etal-2012-sentence,zhang-lapata-2017-sentence,zhao-etal-2018-integrating},
but only one work follows the paradigm of style transfer.
With a focus on scientific (or technical) texts,
\cite{cao-etal-2020-expertise} performed expertise style transfer
suggesting reformulations of sentences like ``\textit{Many cause
  dyspnea, pleuritic chest pain, or both.}'' as ``\textit{The most
  common symptoms, regardless of the type of fluid in the pleural
  space or its cause, are shortness of breath and chest
  pain.}''. Their goal was to demonstrate how paraphrasing medical
jargon can promote better understanding. Hence, for this task, 
the stylistic attribute of a text is given by the level of domain
knowledge that the text involves.

\paragraph{Data} An obvious prerequisite for style transfer
in a specialized genre is the availability of domain-specific
data. \cite{cao-etal-2020-expertise} introduced an expert-annotated
parallel
corpus\footnote{\url{https://srhthu.github.io/expertise-style-transfer/}}
in the medical domain.  It was derived from human-written medical
references tailored for consumers vs.\ healthcare professionals who, in
their view, are set apart by two major knowledge gaps: one related to
technical terminology (``\textit{dyspnea}'' $\rightarrow$
``\textit{shortness of breath}'') and one related to the understanding
of empirical evidence (e.g., ``\textit{About 1/1,000}'' $\rightarrow$
``\textit{quite small}'').

\paragraph{Methods} The major contribution of
\cite{cao-etal-2020-expertise} was the dataset itself, that they
evaluated with five state-of-the-art models from prior style transfer
\citep{pmlr-v70-hu17e, li-etal-2018-delete, dai-etal-2019-style} and
text simplification studies \citep{shardlow-nawaz-2019-neural,
  surya-etal-2019-unsupervised}.
		
\paragraph{Evaluation} The adopted evaluation methods in
\cite{cao-etal-2020-expertise} were transfer accuracy based on a
classifier's performance, fluency based on the perplexity of a
fine-tuned \textsc{Bert} model, and content preservation computed in
terms of \textsc{Bleu}.  In their human evaluation study, 
laypeople rated content
preservation in the model-generated output on a 1-to-5 scale, given
both the input and human-produced gold references.  The metrics
\textsc{Sari} \citep{xu-etal-2016-optimizing} was also used to
evaluate language simplicity, as it compares the n-grams in the
generated output with the input and human references, taking into
account the words that were added, deleted and retained by the model.
The authors concluded that for transfers regarding this style, there
exists a substantial difference between the quality of
machine-produced and human-produced texts.

\subsubsection{Literature}

Literature-centered styles have sparked many formulations of style
transfer. Most of them tackle the problem of making an old text sound
more modern, but ultimately, this type of task shifts the attributes
of several styles simultaneously. Even those works that present
themselves as mapping text between diachronically different language
varieties, in fact, transfer between textual structures (e.g., from
sonnets to plain sentences), including differences at various levels
of granularity: in the register, in the vocabulary choices, in the
senses of words, and in the syntactical constructions
\citep{jhamtani-etal-2017-shakespearizing}. This also occurs in some
studies that focus on author imitation -- i.e., rewriting sentences as
if that was done by a well-known author, to mimic their stylistic
touch \citep{he2020a}\footnote{A challenge of this family of styles is
  given by the name of the characters present in a story, which
  differs from author to author --an interesting study in this
  direction was made by \cite{stamatatos-2017-authorship}.}.

In this light, \textit{literature} in style transfer seems related to
a notion of idiostyle (i.e., a space of linguistic
idiosyncrasies specific to writers), which makes it kin to the
\textit{background} node of \textit{persona} in our hierarchy.
Nevertheless, we dedicate a separate discussion to it as an
\textit{intended} style because the writers' artistic speech might reflect
the (unintentionally expressed) style of the time but does not
coincide with it -- within certain time spans, it is actually the
idiostyle of established writers that creates a linguo-typological
variant of literary texts \citep{sydorenko2018notion}. Moreover,
such idiostyles need to be (intentionally) adapted to the genre of the
writers' literary productions, as these are
intended to have an audience.

There exist many examples of this stream of
research. \cite{shang-etal-2019-semi} paraphrased old
Chinese poems, \cite{bujnowski-etal-2020-empirical} and
\cite{carlson2018evaluating} switched between the prose attributes of various
versions of the Bible (``\textit{Then Samuel gave him an account of
  everything, keeping nothing back}'' $\rightarrow$ ``\textit{And
  Samuel told all things, and did not hold back}'');
\cite{xu-etal-2012-paraphrasing},
\cite{jhamtani-etal-2017-shakespearizing}, and \cite{he2020a} accounted for
the features of Shakespearean plays, transferring Early Modern to
contemporary English (``\textit{I will bite thee by the ear for that
  jest}'' $\rightarrow$ ``\textit{I'll bite you by the ear for that
  joke}'') or vice versa (``\textit{Send thy man away}'' $\rightarrow$
``\textit{Send your man away}''). A similar goal was addressed by
\cite{pang-gimpel-2019-unsupervised} but with Dickens' literature,
while \cite{krishna-etal-2020-reformulating} performed style transfer
with different styles and attributes, transforming tweets into Shakespearean-like
texts, Shakespearean texts into Joyce-sounding writings\footnote{Note that we
  did not mention transfer works that shift style from one author
  to the other by including multiple authors
  \citep[e.g.,][]{syed2020adapting,singh-etal-2021-drag}. As opposed
  to the Shakespeare-Joyce example given above, which paraphrases
  texts conditioned on a diachronical dimensions and with respect to
  their poem or poetry nature, these works take style as persistent
  characteristics of specific individuals. Hence, they cannot be
  generalized and subsumed under any specific style category in our
  hierarchy.}, Joyce-authored texts into Bible-styled ones, and Bible
verses into poems. These works hence exemplify that there are transfer
works in which the shift does not occur along one conceptual dimension
(e.g., presence vs.\ absence of Shakespeare's style), but rather go
from a style to another (e.g., from Shakespeare to Joyce). Therefore,
to view style as a non-categorical variable seems a good option for
this task. As delineated in \cite{romanov-etal-2019-adversarial}, this
would not only account for the reality of language in which the
attributes of different genres\footnote{By ``genre'' we mean what
  \cite{romanov-etal-2019-adversarial} call ```register''.} overlap,
but if applied to the literature of specific authors, it would allow
to understand how each author relates to the others in a continuous
stylistic space.

\cite{gero-etal-2019-low} offered yet another perspective, which
radically re-thinks the relation of style to content. They delineated
a well-defined notion of style in literature, starting from an early
quantitative study by \cite{mendenhall1887characteristic}, which
revealed that writers present some systematic features in their
vocabulary, word length, word frequencies and compositions. To
\cite{gero-etal-2019-low}, this means that words that are most
frequently used (i.e., non-content words) are actually those most
indicative of one's literary style.  They thus showed that non-content words
allow a classifier to determine style, and they leveraged those to transfer
between gothic novels, philosophy books, and pulp science fiction,
hereafter sci-fi.

\paragraph{Data} \cite{carlson2018evaluating} contributed to fixing
the lack of parallel data for style transfer. They collected a
high-quality parallel corpus without the involvement of any automatic
alignment effort. Their resource contains 34 versions of the Bible
produced by professionals and which are naturally aligned, given the
structure of such texts, i.e., in chapters and verses. Each version
corresponds to an English stylistic value (e.g., archaic, simple,
American). They made the dataset available for the texts that were
already public.

\cite{pang-gimpel-2019-unsupervised} limited themselves to two
variants of English, with the old one taken from Dickens' works in
Project Gutenberg\footnote{\url{https://www.gutenberg.org}} and the
modern version from the Toronto Books Corpus. Focusing on Chinese,
\cite{shang-etal-2019-semi} constructed a parallel corpus containing
old and modern versions of poems. \cite{xu-etal-2012-paraphrasing}
made a sentence-aligned corpus of Shakespearean plays and their modern
translations freely available.  \cite{krishna-etal-2020-reformulating}
built a non-parallel English corpus containing 15M sentences, which
contain 11 styles, including the Bible, Shakespeare, James Joyce.
Lastly, the philosophy texts, sci-fi and gothic novels of
\cite{gero-etal-2019-low} also come from mono-style sources. They were
extracted from Project Gutenberg and the Pulp Magazine
Archive\footnote{\url{https://archive.org/details/scifi-corpus}},
respectively.

\paragraph{Methods} The first attempt at dealing with literature
styles explored statistical machine translation
\citep{xu-etal-2012-paraphrasing}; on top of that,
\cite{carlson2018evaluating} went for sequence-to-sequence translation
models, trained for each target attribute.  A sequence-to-sequence network
was also leveraged by \cite{jhamtani-etal-2017-shakespearizing}. They
added both a pointer that facilitates the copy of input words, and a
dictionary of shakespearean-to-modern word pairs which allows to
retrofit pretrained word embeddings, thus accounting for novel words
or words that have changed in meaning.

On the unsupervised side, \cite{pang-gimpel-2019-unsupervised}
experimented with models that include losses corresponding to the
three criteria, and that could be used both for model tuning
and selection. Among such losses, many of which had been already
explored \citep[i.a.,]{shen2017style}, they tried to favor content
preservation with a reconstruction loss, a cyclic consistency
loss (similar to the former, but with the transfer happening twice,
i.e., from source to target and back), and a paraphrase loss
obtained with sentence-paraphrase pairs coming from a parallel
dataset.

Author mimicking was addressed with the probabilistic approach of
\cite{he2020a}; similarly aiming at minimizing the manually defined
objectives (e.g., content-to-style separation), the semi-supervised
method of \cite{shang-etal-2019-semi} employed an encoder-decoder that
learns to represent a style within a specific latent space, and a
projection function that maps the latent representations of one attribute
onto the other.  The two steps leveraged
non-parallel and parallel data respectively.  Instead,
\cite{krishna-etal-2020-reformulating} adopted their inverse
paraphrasing approach already introduced with the \textit{background}
styles.

Style and content were handled separately by
\cite{gero-etal-2019-low}.  In line with their POS-based
characterization of style, they defined some low-level linguistic
features (e.g., frequency of pronouns, prepositions) as the style of a
text, and they performed style transfer by inputting an
encoder-decoder with only the content words, which allowed the
generation to maintain them while adjusting the features of the target
attribute.  By contrast, \cite{pmlr-v70-mueller17a} refrained from
defining editing features or rules.  Claiming that revisions of
combinatorial structures are unlikely to be found by simple search
procedures, they addressed the Shakespearization of language as a
problem of finding improved rewrites of a text.
 
\paragraph{Evaluation} To measure the quality of paraphrases,
\cite{carlson2018evaluating},
\cite{jhamtani-etal-2017-shakespearizing} and
\cite{xu-etal-2012-paraphrasing} accompanied \textsc{Bleu}, a measure
that fundamentally favors textual similarity at the word level, with
\textsc{Pinc}, which instead rewards the diversity of the output from
the source text thanks to the number of n-grams in the candidate
output that do not appear in the source. To measure the
transfer strength criterion,
\cite{xu-etal-2012-paraphrasing} used
a language model to compute the posterior probability that a sentence
was generated from a model of the target language.

\cite{pang-gimpel-2019-unsupervised} introduced a way to measure the
success of transfer by aggregating the metrics: an adjusted geometric
mean between the accuracy, content preservation and perplexity, which
penalizes perplexity scores that are too low, often achieved with
short phrases but not meaningful sentences. For human evaluation,
their annotators decided which of two generated sentences they
preferred with respect to the three transfer criteria. The sentences
were taken from different model variants, to observe the correlation
between human judgments and each system.

\subsubsection{Song lyrics}
``\textit{Yo, where the hell you been?}''  $\rightarrow$ ``\textit{Yo,
  where the hell are you?}''  is an example of transfer from
\cite{lee-etal-2019-neural}, who shifted the genre of lyrics between
Hip Hop and Pop songs.  A similar attempt was made by
\cite{krishna-etal-2020-reformulating}. Their work did not directly
alter lyrics attributes (i.e., the music category to which lyrics
would belong), but it mapped such texts to a completely different
style. As a result, for instance, they made lyrics gain the style of
tweets produced by African American English writers (e.g., given the
input ``\textit{It's a good thing you don't have bus fare}'', an output would
be ``\textit{It's a goof thing u aint gettin no ticket}'').

\paragraph{Data} This task leveraged non-parallel lyrics resources from
MetroLyrics\footnote{\url{http://www.kaggle.com/gyani95/380000-lyrics-from-metrolyrics}}
in which more than 500k songs are associated to specific music genres.

\paragraph{Methods} \cite{lee-etal-2019-neural} treated the problem as
a denoising one, with the same model used to transfer the
\textit{background} of \textit{persona} described in
Section~\ref{methods-implicit}. The non-parallel source data were
noised with a model trained on clean-noisy sentence pairs extracted
from a language learner forum; the newly synthesized texts were then
re-ranked according to their proximity to the target attribute and to the
meaning of the source inputs; lastly, a denoising model was trained to
find the probability of a clean text (i.e., the target), given the
noisy one (i.e., the source).

\paragraph{Evaluation} Unlike other studies,
\cite{lee-etal-2019-neural} defined the transfer strength criterion as
the ratio between the probability of the output belonging to the
target domain and the probability of observing it in the source
domain.

\section{Discussion and conclusion}
\label{sec:conclusions}
Style transfer seems to have a bright future ahead owing to its myriad
of applications, from online communication (e.g., as an assistant) to
studies within NLP (e.g., for data augmentation), and its potential to
reveal facts about language.  ``Operating at all linguistic levels
(e.g., lexicology, syntax, text linguistics, and intonation) [...]
style may be regarded as a choice of linguistic means; as deviation
from a norm; as recurrence of linguistic forms; and as comparison.''
\citep{mukherjee}.  Language is creative, it is situated, and has to
do with our communicative competence: its users can give new meanings
to old words \citep{black1968labyrinth}, produce utterances within a
particular time and place \citep{bamman-etal-2014-distributed}, and
determine if they are appropriate in specific contexts
\citep{hymes-relativity}. Hence, the variety of realizations in which
the same message can be shaped stems from many distinct factors. On
the one hand are variations related to personal differences between
speakers (e.g., a person's class, gender, social environment) and on
the other are those occurring within the speech acts of a single
speaker \citep{labov1966social}.  We unified these insights into a
hierarchy of styles, as a way to relate them to one
another.

Our discussion started from the frameworks typically used to learn the
task. We summarized the method-oriented survey of \citet{hu2020text},
and showed that many publications consider transfer as a problem of
translation between attributes, others assume that style lurks in
certain portions of texts and transform it with localized textual
changes, or leverage special training functions to reach the three
output desiderata. Tables~\ref{tab:unintended},
\ref{tab:unintendeddynamic}, \ref{tab:intendedtargeted},
\ref{tab:intendednontargetedcircumstancial}, and
\ref{tab:intendednontargetedconventional} give an overview of the
studies we detailed by style and method, and they further include some
recent pre-prints that we did not explicitly mention in the main
text. Are current methods sufficient to tackle the complexity of a
style of interest?  The tables show that not all methods have been
evaluated for all styles. The reader is left with the decision of
whether this is a signal for promising research gaps, or instead
points at an important caveat of style transfer. Namely, some
approaches might be acceptable to alter, e.g., sentiment, like
retrieval-based frameworks, but they might miss the mark for styles in
which paraphrases can be expected to be bolder, non-limited to lexical
changes \citep{yamshchikov-etal-2019-decomposing}. In this sense, our
style-oriented survey was also meant to encourage new technical
development.

More importantly, we pushed style transfer to question the
styles it addresses, while acknowledging that many others (and more
varied attributes than binary ones) could be explored.  Our analysis
revealed that some are under-explored and inherently difficult to
transfer. An example is \textit{humor}, a multifaceted phenomenon with
tremendous variation depending on the culture and the social settings
in which it is deployed. Further, many styles are intertwined.  For
instance, we put \textit{background} with other stable traits as an
inter-speaker difference (i.e., under \textit{persona}), but this
choice does not account for speakers shifting their general speech
patterns over time (similar to a \textit{dynamic state}), as a result
of moving to a different dialect region or interacting with different
social groups. On a higher level in the hierarchy, style
contaminations are possible between \textit{intended} styles, and
between them and \textit{unintended} subsets, e.g., one can write a
poem while being romantic, and a certain cultural background can
emerge while being more or less polite.  This is also reflected in the
varied ways in which the publications themselves formulate the
transfer problem.  A case in point is \textit{literature}, which fits
multiple positions in the hierarchy, as it is addressed by some as a
diachronic variation \citep{romanov-etal-2019-adversarial} and by
others as author mimicking \citep{he2020a}.

The interconnection between the \textit{unintended} and
\textit{intended} branches of the hierarchy exemplifies that styles
are a multidimensional concept and cannot always be told apart from
one another. Informative in this regard are a number of studies that
did not revolve around transfer, such as those by
\cite{riloff-etal-2013-sarcasm}, \cite{mohammad-etal-2016-metaphor}
and \cite{felt-riloff-2020-recognizing} concerned with the link
between affective states (e.g., \textit{emotion state}) and figurative
language (i.e., \textit{literality}).  At the same time, only some
combinations of stylistic attributes might be acceptable. As pointed
out in an investigation of style inter-dependence
\citep{kang-hovy-2021-style}, the presence of impoliteness and
positive sentiment in the same text might be paradoxical.

A more serious theoretical understanding of style could inform future
computational research. For one thing, it could cast doubt on the
possibility of addressing style transfer with any feature of text that
can be shifted along some dimensions and that appears to tie in with
some extra-propositional content of texts -- a trend that currently
dominates the field. If anything, evaluation approaches can be refined
for said styles. The outputs of state-of-the-art systems reveal indeed
that the available evaluation metrics are inadequate, but the problem
might be upstream. Namely, the three criteria quantified by such
metrics arguably generalize across styles. Is a successful system for
the transfer of sentiment supposed to maintain meaning as much as a
politeness-conditioned system?  Precisely because different styles
have different linguistic realizations, expecting that the
systems addressing them (often, the very same system)
perform similarly seems somewhat unreasonable.  Transfer, meaning, and grammaticality may be
variously reached for each style, making it more urgent to ask
``\textit{to what extent can a method changing the polarity of a text
  retain its semantics?}'' than measuring if it did.  In other words,
an investigation of transfer with respect to individual styles can
redefine the task at hand and reconsider the attainable goals.

Readers might have noticed that we indistinctly called ``style'' both
linguistic variations (e.g., formality) and aspects that underlie them
(gender correlates with, but is not, style). We also disregarded 
if the selected articles actually deal with a feature of
language that corresponds to \textit{how} things are said: all the
styles that the body of research presents as such were included in
our hierarchy. In fact, this field lacks a stable definition of style
-- unsurprisingly, since no consensus exists on it.

Neither did we take the challenge to define ``style'' ourselves.  We
gave a loose characterization of it, adapting one that is established
among linguists \citep{bell}. That is, style correlates to external
factors, of which gender and personality are an instance. Still, the
example outputs we provided convey the following: to assume that a
text can be paraphrased with any attribute corresponds to taking style and
content as independent variables. In style transfer, the binomial
is thought of in terms groups of ``semantic equivalence'' 
subsuming textual instances that differ with respect to their stylistic
attribute. However, this view has an evident consequence for the field:
 if shaping a meaning into specific attributes seems
unfeasible (e.g., the transfer of sentiment comes at the expense of
losing content, contradicting the independence assumption), then such
attributes cannot define a goal for style transfer.  Content is
information predictive of a future (e.g., what word comes next?),
while style is additional information prior to generation and tapping on
some personal states of the writers. It is grounded in reality, in the
human experience (e.g., gender, ethnicity), and ultimately, in the
reasons that push speakers to communicate and that current machines
(struggling to transfer) do not have.

\section*{Acknowledgements}
This work was supported by Deutsche Forschungsgemeinschaft (project
CEAT, KL 2869/1-2) and the Leibniz WissenschaftsCampus T\"ubingen
``Cognitive Interfaces''.

\subsection*{Competing interests}
We do not declare any competing interests.

\bibliographystyle{nlelike}
\bibliography{lit}

\begin{thebibliography}{}

\bibitem[Abdul-Mageed and Ungar, 2017]{abdul-mageed-ungar-2017-emonet}
{ Abdul-Mageed, M.} {and} { Ungar, L.} 2017.
\newblock {E}mo{N}et: Fine-grained emotion detection with gated recurrent
  neural networks.
\newblock In {\em Proceedings of the 55th Annual Meeting of the Association for
  Computational Linguistics (Volume 1: Long Papers)}, pp. 718--728, Vancouver,
  Canada. Association for Computational Linguistics.

\bibitem[Aitchison, 1981]{aitchison1981language}
{ Aitchison, J.} 1981.
\newblock {\em Language change: Progress or decay?}
\newblock London: Fontana.

\bibitem[Alba-Juez and Attardo, 2014]{alba2014evaluative}
{ Alba-Juez, L.} {and} { Attardo, S.} 2014.
\newblock The evaluative palette of verbal irony.
\newblock {\em Evaluation in context}, 242:93--116.

\bibitem[Allaway and McKeown, 2021]{allaway-mckeown-2021-unified}
{ Allaway, E.} {and} { McKeown, K.} 2021.
\newblock A unified feature representation for lexical connotations.
\newblock In {\em Proceedings of the 16th Conference of the European Chapter of
  the Association for Computational Linguistics: Main Volume}, pp. 2145--2163,
  Online. Association for Computational Linguistics.

\bibitem[Amin and Burghardt, 2020]{amin-burghardt-2020-survey}
{ Amin, M.} {and} { Burghardt, M.} 2020.
\newblock A survey on approaches to computational humor generation.
\newblock In {\em Proceedings of the The 4th Joint SIGHUM Workshop on
  Computational Linguistics for Cultural Heritage, Social Sciences, Humanities
  and Literature}, pp. 29--41, Online. International Committee on Computational
  Linguistics.

\bibitem[Attardo and Raskin, 1991]{attardo-raskin-1991-joke}
{ Attardo, S.} {and} { Raskin, V.} 1991.
\newblock Script theory revis(it)ed: joke similarity and joke representation
  model.
\newblock {\em Humor: International Journal of Humor Research},
  4(3-4):293--348.

\bibitem[Balasubramanian et~al., 2021]{balasubramanian-etal-2021-polarized}
{ Balasubramanian, V.}, { Kobyzev, I.}, { Bahuleyan, H.}, { Shapiro, I.}, {and}
  { Vechtomova, O.} 2021.
\newblock Polarized-{VAE}: Proximity based disentangled representation learning
  for text generation.
\newblock In {\em Proceedings of the 16th Conference of the European Chapter of
  the Association for Computational Linguistics: Main Volume}, pp. 416--423,
  Online. Association for Computational Linguistics.

\bibitem[Bamman et~al., 2014]{bamman-etal-2014-distributed}
{ Bamman, D.}, { Dyer, C.}, {and} { Smith, N.~A.} 2014.
\newblock Distributed representations of geographically situated language.
\newblock In {\em Proceedings of the 52nd Annual Meeting of the Association for
  Computational Linguistics (Volume 2: Short Papers)}, pp. 828--834, Baltimore,
  Maryland. Association for Computational Linguistics.

\bibitem[Banerjee and Lavie, 2005]{banerjee2005meteor}
{ Banerjee, S.} {and} { Lavie, A.} 2005.
\newblock Meteor: An automatic metric for {MT} evaluation with improved
  correlation with human judgments.
\newblock In {\em Proceedings of the acl workshop on intrinsic and extrinsic
  evaluation measures for machine translation and/or summarization}, pp.
  65--72.

\bibitem[Bao et~al., 2019]{bao-etal-2019-generating}
{ Bao, Y.}, { Zhou, H.}, { Huang, S.}, { Li, L.}, { Mou, L.}, { Vechtomova,
  O.}, { Dai, X.-y.}, {and} { Chen, J.} 2019.
\newblock Generating sentences from disentangled syntactic and semantic spaces.
\newblock In {\em Proceedings of the 57th Annual Meeting of the Association for
  Computational Linguistics}, pp. 6008--6019, Florence, Italy. Association for
  Computational Linguistics.

\bibitem[Barbieri et~al., 2014]{barbieri-etal-2014-modelling}
{ Barbieri, F.}, { Saggion, H.}, {and} { Ronzano, F.} 2014.
\newblock Modelling sarcasm in {T}witter, a novel approach.
\newblock In {\em Proceedings of the 5th Workshop on Computational Approaches
  to Subjectivity, Sentiment and Social Media Analysis}, pp. 50--58, Baltimore,
  Maryland. Association for Computational Linguistics.

\bibitem[Beard, 2000]{beard2000language}
{ Beard, A.} 2000.
\newblock {\em The Language of Politics}.
\newblock Intertext (London). Routledge.

\bibitem[Beckmann and Wood, 2017]{beckmann2017dynamic}
{ Beckmann, N.} {and} { Wood, R.~E.} 2017.
\newblock Dynamic personality science. integrating between-person stability and
  within-person change.
\newblock {\em Frontiers in psychology}, 8:1486.

\bibitem[Bell, 1984]{bell}
{ Bell, A.} 1984.
\newblock Language style as audience design.
\newblock {\em Language in Society}, 13(2):145--204.

\bibitem[Beukeboom and Semin, 2006]{beukeboom2006mood}
{ Beukeboom, C.~J.} {and} { Semin, G.~R.} 2006.
\newblock How mood turns on language.
\newblock {\em Journal of experimental social psychology}, 42(5):553--566.

\bibitem[Biber, 1995]{biber1995dimensions}
{ Biber, D.} 1995.
\newblock {\em Dimensions of register variation: A cross-linguistic
  comparison}.
\newblock Cambridge University Press.

\bibitem[Biber, 2012]{biber2012register}
{ Biber, D.} 2012.
\newblock Register as a predictor of linguistic variation.
\newblock {\em Corpus linguistics and linguistic theory}, 8(1):9--37.

\bibitem[Biber and Conrad, 2009]{biber2009register}
{ Biber, D.} {and} { Conrad, S.} 2009.
\newblock {\em Register, genre, and style}.
\newblock Cambridge University Press.

\bibitem[Black, 1968]{black1968labyrinth}
{ Black, M.} 1968.
\newblock {\em The labyrinth of language}.
\newblock New York: Mentor.

\bibitem[Blodgett et~al., 2016]{blodgett-etal-2016-demographic}
{ Blodgett, S.~L.}, { Green, L.}, {and} { O{'}Connor, B.} 2016.
\newblock Demographic dialectal variation in social media: A case study of
  {A}frican-{A}merican {E}nglish.
\newblock In {\em Proceedings of the 2016 Conference on Empirical Methods in
  Natural Language Processing}, pp. 1119--1130, Austin, Texas. Association for
  Computational Linguistics.

\bibitem[Bloomfield, 1927]{bloomfield1927literate}
{ Bloomfield, L.} 1927.
\newblock Literate and illiterate speech.
\newblock {\em American speech}, 2(10):432--439.

\bibitem[Bo et~al., 2021]{bo2020authorship}
{ Bo, H.}, { Ding, S. H.~H.}, { Fung, B. C.~M.}, {and} { Iqbal, F.} 2021.
\newblock {ER}-{AE}: Differentially private text generation for authorship
  anonymization.
\newblock In {\em Proceedings of the 2021 Conference of the North American
  Chapter of the Association for Computational Linguistics: Human Language
  Technologies}, pp. 3997--4007, Online. Association for Computational
  Linguistics.

\bibitem[Bolukbasi et~al., 2016]{bolukbasi2016man}
{ Bolukbasi, T.}, { Chang, K.-W.}, { Zou, J.~Y.}, { Saligrama, V.}, {and} {
  Kalai, A.~T.} 2016.
\newblock Man is to computer programmer as woman is to homemaker? debiasing
  word embeddings.
\newblock {\em Advances in neural information processing systems},
  29:4349--4357.

\bibitem[Bosselut et~al., 2019]{bosselut-etal-2019-comet}
{ Bosselut, A.}, { Rashkin, H.}, { Sap, M.}, { Malaviya, C.}, { Celikyilmaz,
  A.}, {and} { Choi, Y.} 2019.
\newblock {COMET}: Commonsense transformers for automatic knowledge graph
  construction.
\newblock In {\em Proceedings of the 57th Annual Meeting of the Association for
  Computational Linguistics}, pp. 4762--4779, Florence, Italy. Association for
  Computational Linguistics.

\bibitem[Bostan and Klinger, 2018]{bostan-klinger-2018-analysis}
{ Bostan, L.-A.-M.} {and} { Klinger, R.} 2018.
\newblock An analysis of annotated corpora for emotion classification in text.
\newblock In {\em Proceedings of the 27th International Conference on
  Computational Linguistics}, pp. 2104--2119, Santa Fe, New Mexico, USA.
  Association for Computational Linguistics.

\bibitem[Boydstun et~al., 2014]{boydstun2014tracking}
{ Boydstun, A.~E.}, { Card, D.}, { Gross, J.}, { Resnick, P.}, {and} { Smith,
  N.~A.} 2014.
\newblock Tracking the development of media frames within and across policy
  issues.
\newblock {\em American Political Science Association 2014 Annual Meeting
  Paper}.

\bibitem[Bredin, 1998]{bredin-1998-simile}
{ Bredin, H.} 1998.
\newblock Comparisons and similes.
\newblock {\em Lingua}, 105(1):67--78.

\bibitem[Brennan et~al., 2012]{brennan2012adversarial}
{ Brennan, M.}, { Afroz, S.}, {and} { Greenstadt, R.} 2012.
\newblock Adversarial stylometry: Circumventing authorship recognition to
  preserve privacy and anonymity.
\newblock {\em ACM Transactions on Information and System Security (TISSEC)},
  15(3):1--22.

\bibitem[Brentano, 1874]{brentano2012psychology}
{ Brentano, F.} 1874.
\newblock {\em Psychology from an Empirical Standpoint}.
\newblock London: Routledge and Kegan Paul.

\bibitem[Briakou et~al., 2021a]{briakou-etal-2021-evaluating}
{ Briakou, E.}, { Agrawal, S.}, { Tetreault, J.}, {and} { Carpuat, M.} 2021a.
\newblock Evaluating the evaluation metrics for style transfer: A case study in
  multilingual formality transfer.
\newblock In {\em Proceedings of the 2021 Conference on Empirical Methods in
  Natural Language Processing}, pp. 1321--1336, Online and Punta Cana,
  Dominican Republic. Association for Computational Linguistics.

\bibitem[Briakou et~al., 2021b]{briakou-etal-2021-review}
{ Briakou, E.}, { Agrawal, S.}, { Zhang, K.}, { Tetreault, J.}, {and} {
  Carpuat, M.} 2021b.
\newblock A review of human evaluation for style transfer.
\newblock In {\em Proceedings of the 1st Workshop on Natural Language
  Generation, Evaluation, and Metrics (GEM 2021)}, pp. 58--67, Online.
  Association for Computational Linguistics.

\bibitem[Briakou et~al., 2021c]{briakou-etal-2021-ola}
{ Briakou, E.}, { Lu, D.}, { Zhang, K.}, {and} { Tetreault, J.} 2021c.
\newblock Ol{\'a}, bonjour, salve! {XFORMAL}: A benchmark for multilingual
  formality style transfer.
\newblock In {\em Proceedings of the 2021 Conference of the North American
  Chapter of the Association for Computational Linguistics: Human Language
  Technologies}, pp. 3199--3216, Online. Association for Computational
  Linguistics.

\bibitem[Briot et~al., 2020]{briot2020deep}
{ Briot, J.-P.}, { Hadjeres, G.}, {and} { Pachet, F.} 2020.
\newblock {\em Deep learning techniques for music generation}.
\newblock Springer.

\bibitem[Brown and Fraser, 1979]{brown1979speech}
{ Brown, P.} {and} { Fraser, C.} 1979.
\newblock Speech as a marker of situation.
\newblock In {\em Social markers in speech}, pp. 33--62. Cambridge University
  Press.

\bibitem[Bucholtz, 2006]{bucholtz2006word}
{ Bucholtz, M.} 2006.
\newblock Word up: Social meanings of slang in california youth culture.
\newblock {\em A cultural approach to interpersonal communication: Essential
  readings}, 243:267.

\bibitem[Buechel and Hahn, 2017]{Buechel2017}
{ Buechel, S.} {and} { Hahn, U.} 2017.
\newblock {E}mo{B}ank: Studying the impact of annotation perspective and
  representation format on dimensional emotion analysis.
\newblock In {\em Proceedings of the 15th Conference of the {E}uropean Chapter
  of the Association for Computational Linguistics: Volume 2, Short Papers},
  pp. 578--585, Valencia, Spain. Association for Computational Linguistics.

\bibitem[Bujnowski et~al., 2020]{bujnowski-etal-2020-empirical}
{ Bujnowski, P.}, { Ryzhova, K.}, { Choi, H.}, { Witkowska, K.}, { Piersa, J.},
  { Krumholc, T.}, {and} { Beksa, K.} 2020.
\newblock An empirical study on multi-task learning for text style transfer and
  paraphrase generation.
\newblock In {\em Proceedings of the 28th International Conference on
  Computational Linguistics: Industry Track}, pp. 50--63, Online. International
  Committee on Computational Linguistics.

\bibitem[Camerer et~al., 1989]{camerer1989curse}
{ Camerer, C.}, { Loewenstein, G.}, {and} { Weber, M.} 1989.
\newblock The curse of knowledge in economic settings: An experimental
  analysis.
\newblock {\em Journal of Political Economy}, 97(5):1232--1254.

\bibitem[Camp, 2012]{camp-sarcasm-2012}
{ Camp, E.} 2012.
\newblock Sarcasm, pretense, and the semantics/pragmatics distinction*.
\newblock {\em Noûs}, 46(4):587--634.

\bibitem[Cao et~al., 2020]{cao-etal-2020-expertise}
{ Cao, Y.}, { Shui, R.}, { Pan, L.}, { Kan, M.-Y.}, { Liu, Z.}, {and} { Chua,
  T.-S.} 2020.
\newblock Expertise style transfer: A new task towards better communication
  between experts and laymen.
\newblock In {\em Proceedings of the 58th Annual Meeting of the Association for
  Computational Linguistics}, pp. 1061--1071, Online. Association for
  Computational Linguistics.

\bibitem[Card et~al., 2015]{card-etal-2015-media}
{ Card, D.}, { Boydstun, A.~E.}, { Gross, J.~H.}, { Resnik, P.}, {and} { Smith,
  N.~A.} 2015.
\newblock The media frames corpus: Annotations of frames across issues.
\newblock In {\em Proceedings of the 53rd Annual Meeting of the Association for
  Computational Linguistics and the 7th International Joint Conference on
  Natural Language Processing (Volume 2: Short Papers)}, pp. 438--444, Beijing,
  China. Association for Computational Linguistics.

\bibitem[Carli, 1990]{carli1990gender}
{ Carli, L.~L.} 1990.
\newblock Gender, language, and influence.
\newblock {\em Journal of personality and social psychology}, 59(5):941.

\bibitem[Carlson et~al., 2018]{carlson2018evaluating}
{ Carlson, K.}, { Riddell, A.}, {and} { Rockmore, D.} 2018.
\newblock Evaluating prose style transfer with the bible.
\newblock {\em Royal Society open science}, 5:171920.

\bibitem[Casel et~al., 2021]{casel-etal-2021-emotion}
{ Casel, F.}, { Heindl, A.}, {and} { Klinger, R.} 2021.
\newblock Emotion recognition under consideration of the emotion component
  process model.
\newblock In {\em Proceedings of the 17th Conference on Natural Language
  Processing (KONVENS 2021)}, pp. 49--61, D{\"u}sseldorf, Germany. KONVENS 2021
  Organizers.

\bibitem[Cattell, 1946]{cattell1946personality}
{ Cattell, R.~B.} 1946.
\newblock Personality structure and measurement. i. the operational
  determination of trait unities.
\newblock {\em British Journal of Psychology}, 36(2):88.

\bibitem[Cavalin et~al., 2020]{cavalin-etal-2020-disjoint}
{ Cavalin, P.}, { Vasconcelos, M.}, { Grave, M.}, { Pinhanez, C.}, {and} {
  Alves~Ribeiro, V.~H.} 2020.
\newblock From disjoint sets to parallel data to train {S}eq2{S}eq models for
  sentiment transfer.
\newblock In {\em Findings of the Association for Computational Linguistics:
  EMNLP 2020}, pp. 689--698, Online. Association for Computational Linguistics.

\bibitem[Celli et~al., 2014]{celli2014workshop}
{ Celli, F.}, { Lepri, B.}, { Biel, J.-I.}, { Gatica-Perez, D.}, { Riccardi,
  G.}, {and} { Pianesi, F.} 2014.
\newblock The workshop on computational personality recognition 2014.
\newblock In {\em Proceedings of the 22nd ACM international conference on
  Multimedia}, pp. 1245--1246.

\bibitem[Chakrabarty et~al., 2020a]{chakrabarty-etal-2020-r}
{ Chakrabarty, T.}, { Ghosh, D.}, { Muresan, S.}, {and} { Peng, N.} 2020a.
\newblock {R}{\^{}}3: Reverse, retrieve, and rank for sarcasm generation with
  commonsense knowledge.
\newblock In {\em Proceedings of the 58th Annual Meeting of the Association for
  Computational Linguistics}, pp. 7976--7986, Online. Association for
  Computational Linguistics.

\bibitem[Chakrabarty et~al., 2021]{chakrabarty-etal-2021-entrust}
{ Chakrabarty, T.}, { Hidey, C.}, {and} { Muresan, S.} 2021.
\newblock {ENTRUST}: Argument reframing with language models and entailment.
\newblock In {\em Proceedings of the 2021 Conference of the North American
  Chapter of the Association for Computational Linguistics: Human Language
  Technologies}, pp. 4958--4971, Online. Association for Computational
  Linguistics.

\bibitem[Chakrabarty et~al., 2020b]{chakrabarty-etal-2020-generating}
{ Chakrabarty, T.}, { Muresan, S.}, {and} { Peng, N.} 2020b.
\newblock Generating similes effortlessly like a pro: A style transfer approach
  for simile generation.
\newblock In {\em Proceedings of the 2020 Conference on Empirical Methods in
  Natural Language Processing (EMNLP)}, pp. 6455--6469, Online. Association for
  Computational Linguistics.

\bibitem[Charteris-Black, 2018]{black2018analysing}
{ Charteris-Black, J.} 2018.
\newblock {\em Analysing Political Speeches}.
\newblock Macmillan International Higher Education.

\bibitem[Chawla and Yang, 2020]{chawla-yang-2020-semi}
{ Chawla, K.} {and} { Yang, D.} 2020.
\newblock Semi-supervised formality style transfer using language model
  discriminator and mutual information maximization.
\newblock In {\em Findings of the Association for Computational Linguistics:
  EMNLP 2020}, pp. 2340--2354, Online. Association for Computational
  Linguistics.

\bibitem[Chen and Dolan, 2011]{chen-dolan-2011-collecting}
{ Chen, D.} {and} { Dolan, W.} 2011.
\newblock Collecting highly parallel data for paraphrase evaluation.
\newblock In {\em Proceedings of the 49th Annual Meeting of the Association for
  Computational Linguistics: Human Language Technologies}, pp. 190--200,
  Portland, Oregon, USA. Association for Computational Linguistics.

\bibitem[Chen et~al., 2021]{chen2021controlled}
{ Chen, W.-F.}, { Al~Khatib, K.}, { Stein, B.}, {and} { Wachsmuth, H.} 2021.
\newblock Controlled neural sentence-level reframing of news articles.
\newblock In {\em Findings of the Association for Computational Linguistics:
  EMNLP 2021}, pp. 2683--2693, Punta Cana, Dominican Republic. Association for
  Computational Linguistics.

\bibitem[Chen et~al., 2018]{chen2018learning}
{ Chen, W.-F.}, { Wachsmuth, H.}, { Al~Khatib, K.}, {and} { Stein, B.} 2018.
\newblock Learning to flip the bias of news headlines.
\newblock In {\em Proceedings of the 11th International conference on natural
  language generation}, pp. 79--88.

\bibitem[Cheng et~al., 2020a]{cheng-etal-2020-improving}
{ Cheng, P.}, { Min, M.~R.}, { Shen, D.}, { Malon, C.}, { Zhang, Y.}, { Li,
  Y.}, {and} { Carin, L.} 2020a.
\newblock Improving disentangled text representation learning with
  information-theoretic guidance.
\newblock In {\em Proceedings of the 58th Annual Meeting of the Association for
  Computational Linguistics}, pp. 7530--7541, Online. Association for
  Computational Linguistics.

\bibitem[Cheng et~al., 2020b]{cheng-etal-2020-contextual}
{ Cheng, Y.}, { Gan, Z.}, { Zhang, Y.}, { Elachqar, O.}, { Li, D.}, {and} {
  Liu, J.} 2020b.
\newblock Contextual text style transfer.
\newblock In {\em Findings of the Association for Computational Linguistics:
  EMNLP 2020}, pp. 2915--2924, Online. Association for Computational
  Linguistics.

\bibitem[Chopik and Giasson, 2017]{chopik2017age}
{ Chopik, W.~J.} {and} { Giasson, H.~L.} 2017.
\newblock Age differences in explicit and implicit age attitudes across the
  life span.
\newblock {\em The Gerontologist}, 57(suppl\_2):S169--S177.

\bibitem[Chung et~al., 2015]{chung2015gated}
{ Chung, J.}, { Gulcehre, C.}, { Cho, K.}, {and} { Bengio, Y.} 2015.
\newblock Gated feedback recurrent neural networks.
\newblock In {\em International conference on machine learning}, pp.
  2067--2075. PMLR.

\bibitem[Clift, 1999]{clift1999irony}
{ Clift, R.} 1999.
\newblock Irony in conversation.
\newblock {\em Language in Society}, 28(4):523--553.

\bibitem[Coutinho and Miranda, 2009]{coutinho2009describe}
{ Coutinho, M.~A.} {and} { Miranda, F.} 2009.
\newblock To describe genres: Problems and strategies.
\newblock {\em Genre in a changing world}, pp. 35--55.

\bibitem[Czeresnia~Etinger and Black,
  2019]{czeresnia-etinger-black-2019-formality}
{ Czeresnia~Etinger, I.} {and} { Black, A.~W.} 2019.
\newblock Formality style transfer for noisy, user-generated conversations:
  Extracting labeled, parallel data from unlabeled corpora.
\newblock In {\em Proceedings of the 5th Workshop on Noisy User-generated Text
  (W-NUT 2019)}, pp. 11--16, Hong Kong, China. Association for Computational
  Linguistics.

\bibitem[Dahlmeier et~al., 2013]{dahlmeier-etal-2013-building}
{ Dahlmeier, D.}, { Ng, H.~T.}, {and} { Wu, S.~M.} 2013.
\newblock Building a large annotated corpus of learner {E}nglish: The {NUS}
  corpus of learner {E}nglish.
\newblock In {\em Proceedings of the Eighth Workshop on Innovative Use of {NLP}
  for Building Educational Applications}, pp. 22--31, Atlanta, Georgia.
  Association for Computational Linguistics.

\bibitem[Dai et~al., 2019]{dai-etal-2019-style}
{ Dai, N.}, { Liang, J.}, { Qiu, X.}, {and} { Huang, X.} 2019.
\newblock Style transformer: Unpaired text style transfer without disentangled
  latent representation.
\newblock In {\em Proceedings of the 57th Annual Meeting of the Association for
  Computational Linguistics}, pp. 5997--6007, Florence, Italy. Association for
  Computational Linguistics.

\bibitem[Danescu-Niculescu-Mizil et~al., 2013]{danescu2013computational}
{ Danescu-Niculescu-Mizil, C.}, { Sudhof, M.}, { Jurafsky, D.}, { Leskovec,
  J.}, {and} { Potts, C.} 2013.
\newblock A computational approach to politeness with application to social
  factors.
\newblock In {\em Proceedings of the 51st Annual Meeting of the Association for
  Computational Linguistics}, pp. 250--25.

\bibitem[Davidson et~al., 2017a]{hateoffensive}
{ Davidson, T.}, { Warmsley, D.}, { Macy, M.}, {and} { Weber, I.} 2017a.
\newblock Automated hate speech detection and the problem of offensive
  language.
\newblock In {\em Proceedings of the 11th International AAAI Conference on Web
  and Social Media}, ICWSM '17, pp. 512--515.

\bibitem[Davidson et~al., 2017b]{davidson2017automated}
{ Davidson, T.}, { Warmsley, D.}, { Macy, M.}, {and} { Weber, I.} 2017b.
\newblock Automated hate speech detection and the problem of offensive
  language.
\newblock In {\em Proceedings of the 11th International AAAI Conference on Web
  and Social Media}, ICWSM '17, pp. 512--515.

\bibitem[Davies, 2012]{davies2012expanding}
{ Davies, M.} 2012.
\newblock Expanding horizons in historical linguistics with the 400-million
  word corpus of historical american english.
\newblock {\em Corpora}, 7(2):121--157.

\bibitem[De~Mattei et~al., 2020]{de-mattei-etal-2020-invisible}
{ De~Mattei, L.}, { Cafagna, M.}, { Dell{'}Orletta, F.}, {and} { Nissim, M.}
  2020.
\newblock Invisible to people but not to machines: Evaluation of style-aware
  {H}eadline{G}eneration in absence of reliable human judgment.
\newblock In {\em Proceedings of the 12th Language Resources and Evaluation
  Conference}, pp. 6709--6717, Marseille, France. European Language Resources
  Association.

\bibitem[De~Saussure, 1959]{saussure}
{ De~Saussure, F.} 1959.
\newblock {\em Course in General Linguistics}.
\newblock Philosophical Library, New York.

\bibitem[Devlin et~al., 2019]{devlin-etal-2019-bert}
{ Devlin, J.}, { Chang, M.-W.}, { Lee, K.}, {and} { Toutanova, K.} 2019.
\newblock {BERT}: Pre-training of deep bidirectional transformers for language
  understanding.
\newblock In {\em Proceedings of the 2019 Conference of the North {A}merican
  Chapter of the Association for Computational Linguistics: Human Language
  Technologies, Volume 1 (Long and Short Papers)}, pp. 4171--4186, Minneapolis,
  Minnesota. Association for Computational Linguistics.

\bibitem[dos Santos et~al., 2018]{nogueira-dos-santos-etal-2018-fighting}
{ dos Santos, C.}, { Melnyk, I.}, {and} { Padhi, I.} 2018.
\newblock Fighting offensive language on social media with unsupervised text
  style transfer.
\newblock In {\em Proceedings of the 56th Annual Meeting of the Association for
  Computational Linguistics (Volume 2: Short Papers)}, pp. 189--194, Melbourne,
  Australia. Association for Computational Linguistics.

\bibitem[Dryja\'{n}ski et~al., 2018]{dryjanski-et-al-2018}
{ Dryja\'{n}ski, T.}, { Bujnowski, P.}, { Choi, H.}, { Podlaska, K.}, {
  Michalski, K.}, { Beksa, K.}, {and} { Kubik, P.} 2018.
\newblock Affective natural language generation by phrase insertion.
\newblock In {\em 2018 IEEE International Conference on Big Data (Big Data)},
  pp. 4876--4882.

\bibitem[Dua and Graff, 2017]{uci_repo}
{ Dua, D.} {and} { Graff, C.} 2017.
\newblock {UCI} machine learning repository.

\bibitem[Eckert, 1997]{eckert1997}
{ Eckert, P.} 1997.
\newblock Age as a sociolinguistic variable.
\newblock {\em The handbook of sociolinguistics}, pp. 151--167.

\bibitem[Eckert and McConnell-Ginet, 1999]{eckert_ginet_1999}
{ Eckert, P.} {and} { McConnell-Ginet, S.} 1999.
\newblock New generalizations and explanations in language and gender research.
\newblock {\em Language in Society}, 28(2):185–201.

\bibitem[Eckert and McConnell-Ginet, 2003]{eckert2003language}
{ Eckert, P.} {and} { McConnell-Ginet, S.} 2003.
\newblock {\em Language and gender}.
\newblock Cambridge University Press.

\bibitem[Edelman, 1985]{edelman1985political}
{ Edelman, M.} 1985.
\newblock Political language and political reality.
\newblock {\em PS}, 18(1):10--19.

\bibitem[Eisenstein et~al., 2010]{eisenstein-etal-2010-latent}
{ Eisenstein, J.}, { O{'}Connor, B.}, { Smith, N.~A.}, {and} { Xing, E.~P.}
  2010.
\newblock A latent variable model for geographic lexical variation.
\newblock In {\em Proceedings of the 2010 Conference on Empirical Methods in
  Natural Language Processing}, pp. 1277--1287, Cambridge, MA. Association for
  Computational Linguistics.

\bibitem[Emmery et~al., 2018]{emmery-etal-2018-style}
{ Emmery, C.}, { Manjavacas~Arevalo, E.}, {and} { Chrupa{\l}a, G.} 2018.
\newblock Style obfuscation by invariance.
\newblock In {\em Proceedings of the 27th International Conference on
  Computational Linguistics}, pp. 984--996, Santa Fe, New Mexico, USA.
  Association for Computational Linguistics.

\bibitem[Fang et~al., 2019]{fang-etal-2019-implicit}
{ Fang, L.}, { Li, C.}, { Gao, J.}, { Dong, W.}, {and} { Chen, C.} 2019.
\newblock Implicit deep latent variable models for text generation.
\newblock In {\em Proceedings of the 2019 Conference on Empirical Methods in
  Natural Language Processing and the 9th International Joint Conference on
  Natural Language Processing (EMNLP-IJCNLP)}, pp. 3946--3956, Hong Kong,
  China. Association for Computational Linguistics.

\bibitem[Felbo et~al., 2017]{Felbo2017}
{ Felbo, B.}, { Mislove, A.}, { S{\o}gaard, A.}, { Rahwan, I.}, {and} {
  Lehmann, S.} 2017.
\newblock Using millions of emoji occurrences to learn any-domain
  representations for detecting sentiment, emotion and sarcasm.
\newblock In {\em Proceedings of the 2017 Conference on Empirical Methods in
  Natural Language Processing}, pp. 1615--1625, Copenhagen, Denmark.
  Association for Computational Linguistics.

\bibitem[Felt and Riloff, 2020]{felt-riloff-2020-recognizing}
{ Felt, C.} {and} { Riloff, E.} 2020.
\newblock Recognizing euphemisms and dysphemisms using sentiment analysis.
\newblock In {\em Proceedings of the Second Workshop on Figurative Language
  Processing}, pp. 136--145, Online. Association for Computational Linguistics.

\bibitem[Feng et~al., 2019]{feng-etal-2019-keep}
{ Feng, S.~Y.}, { Li, A.~W.}, {and} { Hoey, J.} 2019.
\newblock Keep calm and switch on! preserving sentiment and fluency in semantic
  text exchange.
\newblock In {\em Proceedings of the 2019 Conference on Empirical Methods in
  Natural Language Processing and the 9th International Joint Conference on
  Natural Language Processing (EMNLP-IJCNLP)}, pp. 2701--2711, Hong Kong,
  China. Association for Computational Linguistics.

\bibitem[Ferris, 2002]{ferris2002writing}
{ Ferris, S.~P.} 2002.
\newblock Writing electronically: The effects of computers on traditional
  writing.
\newblock {\em Journal of electronic publishing}, 8(1).

\bibitem[Fink et~al., 2012]{fink2012inferring}
{ Fink, C.}, { Kopecky, J.}, {and} { Morawski, M.} 2012.
\newblock Inferring gender from the content of tweets: A region specific
  example.
\newblock {\em Proceedings of the International AAAI Conference on Web and
  Social Media}, 6(1).

\bibitem[Foucault, 1966]{foucault1995order}
{ Foucault, M.} 1966.
\newblock {\em Les Mots et les choses}.
\newblock Editions Gallimard.

\bibitem[Founta et~al., 2019]{founta-2019-abuse}
{ Founta, A.~M.}, { Chatzakou, D.}, { Kourtellis, N.}, { Blackburn, J.}, {
  Vakali, A.}, {and} { Leontiadis, I.} 2019.
\newblock A unified deep learning architecture for abuse detection.
\newblock In {\em Proceedings of the 10th ACM Conference on Web Science},
  WebSci '19,  105–114, New York, NY, USA. Association for Computing
  Machinery.

\bibitem[Foxall and Goldsmith, 1988]{foxall1988personality}
{ Foxall, G.~R.} {and} { Goldsmith, R.~E.} 1988.
\newblock Personality and consumer research: Another look.
\newblock {\em Journal of the Market Research Society}, 30(2):111--125.

\bibitem[Friedman and Tucker, 1990]{friedman1990language}
{ Friedman, H.~S.} {and} { Tucker, J.~S.} 1990.
\newblock Language and deception.
\newblock {\em Handbook of language and social psychology}, pp. 257--270.

\bibitem[Fu et~al., 2019]{fu-etal-2019-rethinking}
{ Fu, Y.}, { Zhou, H.}, { Chen, J.}, {and} { Li, L.} 2019.
\newblock Rethinking text attribute transfer: A lexical analysis.
\newblock In {\em Proceedings of the 12th International Conference on Natural
  Language Generation}, pp. 24--33, Tokyo, Japan. Association for Computational
  Linguistics.

\bibitem[Fu et~al., 2018]{fu2018}
{ Fu, Z.}, { Tan, X.}, { Peng, N.}, { Zhao, D.}, {and} { Yan, R.} 2018.
\newblock Style transfer in text: Exploration and evaluation.
\newblock {\em Proceedings of the AAAI Conference on Artificial Intelligence},
  32(1).

\bibitem[Gan et~al., 2017]{caption-dataset}
{ Gan, C.}, { Gan, Z.}, { He, X.}, { Gao, J.}, {and} { Deng, L.} 2017.
\newblock Stylenet: Generating attractive visual captions with styles.
\newblock In {\em 2017 IEEE Conference on Computer Vision and Pattern
  Recognition (CVPR)}, pp. 955--964.

\bibitem[Gao et~al., 2019]{gao-etal-2019-structuring}
{ Gao, X.}, { Zhang, Y.}, { Lee, S.}, { Galley, M.}, { Brockett, C.}, { Gao,
  J.}, {and} { Dolan, B.} 2019.
\newblock Structuring latent spaces for stylized response generation.
\newblock In {\em Proceedings of the 2019 Conference on Empirical Methods in
  Natural Language Processing and the 9th International Joint Conference on
  Natural Language Processing (EMNLP-IJCNLP)}, pp. 1814--1823, Hong Kong,
  China. Association for Computational Linguistics.

\bibitem[Gatt and Krahmer, 2018]{gatt2018survey}
{ Gatt, A.} {and} { Krahmer, E.} 2018.
\newblock Survey of the state of the art in natural language generation: Core
  tasks, applications and evaluation.
\newblock {\em Journal of Artificial Intelligence Research}, 61:65--170.

\bibitem[Gatti et~al., 2012]{gatti2012creatively}
{ Gatti, L.}, { Guerini, M.}, { Callaway, C.~B.}, { Stock, O.}, {and} {
  Strapparava, C.} 2012.
\newblock Creatively subverting messages in posters.
\newblock In {\em ICCC}, pp. 175--179.

\bibitem[Gatti et~al., 2015]{gatti2015slogans}
{ Gatti, L.}, { {\"O}zbal, G.}, { Guerini, M.}, { Stock, O.}, {and} {
  Strapparava, C.} 2015.
\newblock Slogans are not forever: Adapting linguistic expressions to the news.
\newblock In {\em Twenty-Fourth International Joint Conference on Artificial
  Intelligence}.

\bibitem[Gatti et~al., 2016]{gatti2016heady}
{ Gatti, L.}, { Ozbal, G.}, { Guerini, M.}, { Stock, O.}, {and} { Strapparava,
  C.} 2016.
\newblock Heady-lines: A creative generator of newspaper headlines.
\newblock In {\em Companion Publication of the 21st International Conference on
  Intelligent User Interfaces}, pp. 79--83.

\bibitem[Gatys et~al., 2016]{gatys2016image}
{ Gatys, L.~A.}, { Ecker, A.~S.}, {and} { Bethge, M.} 2016.
\newblock Image style transfer using convolutional neural networks.
\newblock In {\em Proceedings of the IEEE Conference on Computer Vision and
  Pattern Recognition}, pp. 2414--2423.

\bibitem[Ge et~al., 2019]{ge-etal-2019-automatic}
{ Ge, T.}, { Zhang, X.}, { Wei, F.}, {and} { Zhou, M.} 2019.
\newblock Automatic grammatical error correction for sequence-to-sequence text
  generation: An empirical study.
\newblock In {\em Proceedings of the 57th Annual Meeting of the Association for
  Computational Linguistics}, pp. 6059--6064, Florence, Italy. Association for
  Computational Linguistics.

\bibitem[Genette, 1997]{genette1997palimpsests}
{ Genette, G.} 1997.
\newblock {\em Palimpsests: Literature in the second degree}, volume~8.
\newblock U of Nebraska Press.

\bibitem[Gero et~al., 2019]{gero-etal-2019-low}
{ Gero, K.}, { Kedzie, C.}, { Reeve, J.}, {and} { Chilton, L.} 2019.
\newblock Low level linguistic controls for style transfer and content
  preservation.
\newblock In {\em Proceedings of the 12th International Conference on Natural
  Language Generation}, pp. 208--218, Tokyo, Japan. Association for
  Computational Linguistics.

\bibitem[Ghosh et~al., 2017]{ghosh-etal-2017-affect}
{ Ghosh, S.}, { Chollet, M.}, { Laksana, E.}, { Morency, L.-P.}, {and} {
  Scherer, S.} 2017.
\newblock Affect-{LM}: A neural language model for customizable affective text
  generation.
\newblock In {\em Proceedings of the 55th Annual Meeting of the Association for
  Computational Linguistics (Volume 1: Long Papers)}, pp. 634--642, Vancouver,
  Canada. Association for Computational Linguistics.

\bibitem[Gibbs~Jr. and Colston, 2006]{gibbs-2006-figurative}
{ Gibbs~Jr., R.~W.} {and} { Colston, H.~L.} 2006.
\newblock Figurative language.
\newblock In {\em Handbook of psycholinguistics}, pp. 835--861. Elsevier.

\bibitem[Giles and Johnson, 1987]{giles1987ethnolinguistic}
{ Giles, H.} {and} { Johnson, P.} 1987.
\newblock Ethnolinguistic identity theory: A social psychological approach to
  language maintenance.
\newblock {\em International Journal of the Sociology of Language}, 68:69--99.

\bibitem[Gohary and Hanzaee, 2014]{gohary2014personality}
{ Gohary, A.} {and} { Hanzaee, K.~H.} 2014.
\newblock Personality traits as predictors of shopping motivations and
  behaviors: a canonical correlation analysis.
\newblock {\em Arab Economic and Business Journal}, 9(2):166--174.

\bibitem[Gong et~al., 2019]{gong-etal-2019-reinforcement}
{ Gong, H.}, { Bhat, S.}, { Wu, L.}, { Xiong, J.}, {and} { Hwu, W.-m.} 2019.
\newblock Reinforcement learning based text style transfer without parallel
  training corpus.
\newblock In {\em Proceedings of the 2019 Conference of the North {A}merican
  Chapter of the Association for Computational Linguistics: Human Language
  Technologies, Volume 1 (Long and Short Papers)}, pp. 3168--3180, Minneapolis,
  Minnesota. Association for Computational Linguistics.

\bibitem[Gong et~al., 2020]{gong2020rich}
{ Gong, H.}, { Song, L.}, {and} { Bhat, S.} 2020.
\newblock Rich syntactic and semantic information helps unsupervised text style
  transfer.
\newblock In {\em Proceedings of the 13th International Conference on Natural
  Language Generation}, pp. 113--119.

\bibitem[Gonz{\'a}lez-Ib{\'a}{\~n}ez et~al.,
  2011]{gonzalez-ibanez-etal-2011-identifying}
{ Gonz{\'a}lez-Ib{\'a}{\~n}ez, R.}, { Muresan, S.}, {and} { Wacholder, N.}
  2011.
\newblock Identifying sarcasm in {T}witter: A closer look.
\newblock In {\em Proceedings of the 49th Annual Meeting of the Association for
  Computational Linguistics: Human Language Technologies}, pp. 581--586,
  Portland, Oregon, USA. Association for Computational Linguistics.

\bibitem[Goswamy et~al., 2020]{goswamy-etal-2020-adapting}
{ Goswamy, T.}, { Singh, I.}, { Barkati, A.}, {and} { Modi, A.} 2020.
\newblock Adapting a language model for controlled affective text generation.
\newblock In {\em Proceedings of the 28th International Conference on
  Computational Linguistics}, pp. 2787--2801, Barcelona, Spain (Online).
  International Committee on Computational Linguistics.

\bibitem[Graesser et~al., 2014]{graesser2014coh}
{ Graesser, A.~C.}, { McNamara, D.~S.}, { Cai, Z.}, { Conley, M.}, { Li, H.},
  {and} { Pennebaker, J.} 2014.
\newblock Coh-metrix measures text characteristics at multiple levels of
  language and discourse.
\newblock {\em The Elementary School Journal}, 115(2):210--229.

\bibitem[Green et~al., 1998]{green1998aspect}
{ Green, L.} {and} { others} 1998.
\newblock Aspect and predicate phrases in african-american vernacular english.
\newblock {\em African-American English: structure, history, and use}, pp.
  37--68.

\bibitem[Grimminger and Klinger, 2021]{Grimminger2021}
{ Grimminger, L.} {and} { Klinger, R.} 2021.
\newblock Hate towards the political opponent: A {T}witter corpus study of the
  2020 {US} elections on the basis of offensive speech and stance detection.
\newblock In {\em Proceedings of the Eleventh Workshop on Computational
  Approaches to Subjectivity, Sentiment and Social Media Analysis}, pp.
  171--180, Online. Association for Computational Linguistics.

\bibitem[Gruner, 1997]{gruner1997game}
{ Gruner, C.~R.} 1997.
\newblock {\em The Game of Humor: A Comprehensive Theory of Why We Laugh}.
\newblock Transaction Publishers.

\bibitem[Guerini et~al., 2008]{guerini2008valentino}
{ Guerini, M.}, { Strapparava, C.}, {and} { Stock, O.} 2008.
\newblock {V}alentino: A tool for valence shifting of natural language texts.
\newblock In {\em Proceedings of the Sixth International Conference on Language
  Resources and Evaluation ({LREC}'08)}, Marrakech, Morocco. European Language
  Resources Association (ELRA).

\bibitem[Guu et~al., 2018]{guu-etal-2018-generating}
{ Guu, K.}, { Hashimoto, T.~B.}, { Oren, Y.}, {and} { Liang, P.} 2018.
\newblock Generating sentences by editing prototypes.
\newblock {\em Transactions of the Association for Computational Linguistics},
  6:437--450.

\bibitem[Habermas, 2006]{habermas2006political}
{ Habermas, J.} 2006.
\newblock Political communication in media society: Does democracy still enjoy
  an epistemic dimension? the impact of normative theory on empirical
  research1.
\newblock {\em Communication Theory}, 16(4):411--426.

\bibitem[Halliday, 1989]{halliday}
{ Halliday, M. A.~K.} 1989.
\newblock {\em Spoken and Written Language}.
\newblock Language education. Oxford University Press.

\bibitem[He et~al., 2020]{he2020a}
{ He, J.}, { Wang, X.}, { Neubig, G.}, {and} { Berg-Kirkpatrick, T.} 2020.
\newblock A probabilistic formulation of unsupervised text style transfer.
\newblock In {\em Proceedings of International Conference on Learning
  Representations}.

\bibitem[Helbig et~al., 2020]{helbig-etal-2020-challenges}
{ Helbig, D.}, { Troiano, E.}, {and} { Klinger, R.} 2020.
\newblock Challenges in emotion style transfer: An exploration with a lexical
  substitution pipeline.
\newblock In {\em Proceedings of the Eighth International Workshop on Natural
  Language Processing for Social Media}, pp. 41--50, Online. Association for
  Computational Linguistics.

\bibitem[Heylighen and Dewaele, 1999]{heylighen1999formality}
{ Heylighen, F.} {and} { Dewaele, J.-M.} 1999.
\newblock Formality of language: Definition, measurement and behavioral
  determinants.
\newblock {\em Interner Bericht, Center “Leo Apostel”, Vrije Universiteit
  Br{\"u}ssel}, 4.

\bibitem[Hofmann et~al., 2020]{Hofmann2020}
{ Hofmann, J.}, { Troiano, E.}, { Sassenberg, K.}, {and} { Klinger, R.} 2020.
\newblock Appraisal theories for emotion classification in text.
\newblock In {\em Proceedings of the 28th International Conference on
  Computational Linguistics}, pp. 125--138, Barcelona, Spain (Online).
  International Committee on Computational Linguistics.

\bibitem[Hoijer, 1954]{hoijer1954language}
{ Hoijer, H.~E.} 1954.
\newblock {\em Language in culture; conference on the interrelations of
  language and other aspects of culture.}
\newblock University of Chicago Press.

\bibitem[Holtgraves, 2001]{holtgraves2013language}
{ Holtgraves, T.~M.} 2001.
\newblock {\em Language as social action: Social psychology and language use}.
\newblock Psychology Press.

\bibitem[Hossain et~al., 2019]{hossain-2019-president}
{ Hossain, N.}, { Krumm, J.}, {and} { Gamon, M.} 2019.
\newblock {``}{P}resident vows to cut {\textless}taxes{\textgreater} hair{''}:
  Dataset and analysis of creative text editing for humorous headlines.
\newblock In {\em Proceedings of the 2019 Conference of the North {A}merican
  Chapter of the Association for Computational Linguistics: Human Language
  Technologies, Volume 1 (Long and Short Papers)}, pp. 133--142, Minneapolis,
  Minnesota. Association for Computational Linguistics.

\bibitem[Hu and Liu, 2006]{hu2006opinion}
{ Hu, M.} {and} { Liu, B.} 2006.
\newblock Opinion extraction and summarization on the web.
\newblock In {\em Proceedings of the 21st National Conference on Artificial
  Intelligence - Volume 2}, AAAI'06,  1621–1624. AAAI Press.

\bibitem[Hu et~al., 2022]{hu2020text}
{ Hu, Z.}, { Lee, R. K.-W.}, { Aggarwal, C.~C.}, {and} { Zhang, A.} 2022.
\newblock Text style transfer: A review and experimental evaluation.
\newblock {\em SIGKDD Explor. Newsl.}, 24(1):14–45.

\bibitem[Hu et~al., 2017]{pmlr-v70-hu17e}
{ Hu, Z.}, { Yang, Z.}, { Liang, X.}, { Salakhutdinov, R.}, {and} { Xing,
  E.~P.} 2017.
\newblock Toward controlled generation of text.
\newblock {\em Proceedings of Machine Learning Research}, 70:1587--1596.

\bibitem[Huang et~al., 2018]{huang-etal-2018-automatic}
{ Huang, C.}, { Za{\"\i}ane, O.}, { Trabelsi, A.}, {and} { Dziri, N.} 2018.
\newblock Automatic dialogue generation with expressed emotions.
\newblock In {\em Proceedings of the 2018 Conference of the North {A}merican
  Chapter of the Association for Computational Linguistics: Human Language
  Technologies, Volume 2 (Short Papers)}, pp. 49--54, New Orleans, Louisiana.
  Association for Computational Linguistics.

\bibitem[Hudson, 1993]{hudson}
{ Hudson, R.~A.} 1993.
\newblock {\em Sociolinguistics}.
\newblock Cambridge University Press.

\bibitem[Hughes et~al., 2012]{hughes2012quantitative}
{ Hughes, J.~M.}, { Foti, N.~J.}, { Krakauer, D.~C.}, {and} { Rockmore, D.~N.}
  2012.
\newblock Quantitative patterns of stylistic influence in the evolution of
  literature.
\newblock {\em Proceedings of the National Academy of Sciences},
  109(20):7682--7686.

\bibitem[Hymes, 1966]{hymes-relativity}
{ Hymes, D.} 1966.
\newblock Two types of linguistic relativity.
\newblock In {\em Sociolinguistics}, pp. 114--158. De Hague: Mouton.

\bibitem[Irvine, 1979]{irvine1979formality}
{ Irvine, J.~T.} 1979.
\newblock Formality and informality in communicative events.
\newblock {\em American Anthropologist}, 81(4):773--790.

\bibitem[Jafaritazehjani et~al., 2020]{jafaritazehjani-etal-2020-style}
{ Jafaritazehjani, S.}, { Lecorv{\'e}, G.}, { Lolive, D.}, {and} { Kelleher,
  J.} 2020.
\newblock Style versus content: A distinction without a (learnable) difference?
\newblock In {\em Proceedings of the 28th International Conference on
  Computational Linguistics}, pp. 2169--2180, Barcelona, Spain (Online).
  International Committee on Computational Linguistics.

\bibitem[Jhamtani et~al., 2017]{jhamtani-etal-2017-shakespearizing}
{ Jhamtani, H.}, { Gangal, V.}, { Hovy, E.}, {and} { Nyberg, E.} 2017.
\newblock Shakespearizing modern language using copy-enriched sequence to
  sequence models.
\newblock In {\em Proceedings of the Workshop on Stylistic Variation}, pp.
  10--19, Copenhagen, Denmark. Association for Computational Linguistics.

\bibitem[Jin et~al., 2022]{jinreview}
{ Jin, D.}, { Jin, Z.}, { Hu, Z.}, { Vechtomova, O.}, {and} { Mihalcea, R.}
  2022.
\newblock Deep learning for text style transfer: A survey.
\newblock {\em Computational Linguistics}, 48(1):155--205.

\bibitem[Jin et~al., 2019]{jin-etal-2019-imat}
{ Jin, Z.}, { Jin, D.}, { Mueller, J.}, { Matthews, N.}, {and} { Santus, E.}
  2019.
\newblock {IM}a{T}: Unsupervised text attribute transfer via iterative matching
  and translation.
\newblock In {\em Proceedings of the 2019 Conference on Empirical Methods in
  Natural Language Processing and the 9th International Joint Conference on
  Natural Language Processing (EMNLP-IJCNLP)}, pp. 3097--3109, Hong Kong,
  China. Association for Computational Linguistics.

\bibitem[John et~al., 2008]{john2008paradigm}
{ John, O.~P.}, { Naumann, L.~P.}, {and} { Soto, C.~J.} 2008.
\newblock Paradigm shift to the integrative big five trait taxonomy: History,
  measurement, and conceptual issues.
\newblock {\em Handbook of Personality Theory and Research}, pp. 114--158.

\bibitem[John et~al., 2019]{john-etal-2019-disentangled}
{ John, V.}, { Mou, L.}, { Bahuleyan, H.}, {and} { Vechtomova, O.} 2019.
\newblock Disentangled representation learning for non-parallel text style
  transfer.
\newblock In {\em Proceedings of the 57th Annual Meeting of the Association for
  Computational Linguistics}, pp. 424--434, Florence, Italy. Association for
  Computational Linguistics.

\bibitem[Kacmarcik and Gamon, 2006]{kacmarcik-gamon-2006-obfuscating}
{ Kacmarcik, G.} {and} { Gamon, M.} 2006.
\newblock Obfuscating document stylometry to preserve author anonymity.
\newblock In {\em Proceedings of the {COLING}/{ACL} 2006 Main Conference Poster
  Sessions}, pp. 444--451, Sydney, Australia. Association for Computational
  Linguistics.

\bibitem[Kang et~al., 2019]{kang-etal-2019-male}
{ Kang, D.}, { Gangal, V.}, {and} { Hovy, E.} 2019.
\newblock (male, bachelor) and (female, {P}h.{D}) have different connotations:
  Parallelly annotated stylistic language dataset with multiple personas.
\newblock In {\em Proceedings of the 2019 Conference on Empirical Methods in
  Natural Language Processing and the 9th International Joint Conference on
  Natural Language Processing (EMNLP-IJCNLP)}, pp. 1696--1706, Hong Kong,
  China. Association for Computational Linguistics.

\bibitem[Kang and Hovy, 2021]{kang-hovy-2021-style}
{ Kang, D.} {and} { Hovy, E.} 2021.
\newblock Style is {NOT} a single variable: Case studies for cross-stylistic
  language understanding.
\newblock In {\em Proceedings of the 59th Annual Meeting of the Association for
  Computational Linguistics and the 11th International Joint Conference on
  Natural Language Processing (Volume 1: Long Papers)}, pp. 2376--2387, Online.
  Association for Computational Linguistics.

\bibitem[Kim et~al., 2017]{Kim2017a}
{ Kim, E.}, { Pad\'{o}, S.}, {and} { Klinger, R.} 2017.
\newblock Investigating the relationship between literary genres and emotional
  plot development.
\newblock In {\em Proceedings of the Joint SIGHUM Workshop on Computational
  Linguistics for Cultural Heritage, Social Sciences, Humanities and
  Literature}, pp. 17--26, Vancouver, Canada. Association for Computational
  Linguistics.

\bibitem[Klimt and Yang, 2004]{klimt-2004-enron}
{ Klimt, B.} {and} { Yang, Y.} 2004.
\newblock The enron corpus: A new dataset for email classification research.
\newblock In {\em European Conference on Machine Learning}, pp. 217--226.
  Springer.

\bibitem[Kreuz and Glucksberg, 1989]{Kreuz1989-KREHTB-3}
{ Kreuz, R.~J.} {and} { Glucksberg, S.} 1989.
\newblock How to be sarcastic: The echoic reminder theory of verbal irony.
\newblock {\em Journal of Experimental Psychology: General}, 118(4):374--386.

\bibitem[Krishna et~al., 2020]{krishna-etal-2020-reformulating}
{ Krishna, K.}, { Wieting, J.}, {and} { Iyyer, M.} 2020.
\newblock Reformulating unsupervised style transfer as paraphrase generation.
\newblock In {\em Proceedings of the 2020 Conference on Empirical Methods in
  Natural Language Processing (EMNLP)}, pp. 737--762, Online. Association for
  Computational Linguistics.

\bibitem[Kruengkrai, 2019]{kruengkrai-2019-learning}
{ Kruengkrai, C.} 2019.
\newblock Learning to flip the sentiment of reviews from non-parallel corpora.
\newblock In {\em Proceedings of the 2019 Conference on Empirical Methods in
  Natural Language Processing and the 9th International Joint Conference on
  Natural Language Processing (EMNLP-IJCNLP)}, pp. 6311--6316, Hong Kong,
  China. Association for Computational Linguistics.

\bibitem[Kusner et~al., 2015]{kusner2015word}
{ Kusner, M.}, { Sun, Y.}, { Kolkin, N.}, {and} { Weinberger, K.} 2015.
\newblock From word embeddings to document distances.
\newblock In {\em International conference on machine learning}, pp. 957--966.
  PMLR.

\bibitem[Labov, 1966]{labov1966social}
{ Labov, W.} 1966.
\newblock {\em The social stratification of English in New York city}.
\newblock Washington, D.C.: Center for Applied Linguistics.

\bibitem[Lai et~al., 2019]{lai-etal-2019-multiple}
{ Lai, C.-T.}, { Hong, Y.-T.}, { Chen, H.-Y.}, { Lu, C.-J.}, {and} { Lin,
  S.-D.} 2019.
\newblock Multiple text style transfer by using word-level conditional
  generative adversarial network with two-phase training.
\newblock In {\em Proceedings of the 2019 Conference on Empirical Methods in
  Natural Language Processing and the 9th International Joint Conference on
  Natural Language Processing (EMNLP-IJCNLP)}, pp. 3579--3584, Hong Kong,
  China. Association for Computational Linguistics.

\bibitem[Lai et~al., 2021]{lai-etal-2021-thank}
{ Lai, H.}, { Toral, A.}, {and} { Nissim, M.} 2021.
\newblock Thank you {BART}! rewarding pre-trained models improves formality
  style transfer.
\newblock In {\em Proceedings of the 59th Annual Meeting of the Association for
  Computational Linguistics and the 11th International Joint Conference on
  Natural Language Processing (Volume 2: Short Papers)}, pp. 484--494, Online.
  Association for Computational Linguistics.

\bibitem[Lample et~al., 2019]{lample2018multiple}
{ Lample, G.}, { Subramanian, S.}, { Smith, E.~M.}, { Denoyer, L.}, { Ranzato,
  M.}, {and} { Boureau, Y.} 2019.
\newblock Multiple-attribute text rewriting.
\newblock In {\em 7th International Conference on Learning Representations,
  {ICLR} 2019, New Orleans, LA, USA, May 6-9, 2019}. OpenReview.net.

\bibitem[Lave et~al., 1991]{lave1991situated}
{ Lave, J.}, { Wenger, E.}, {and} { others} 1991.
\newblock {\em Situated learning: Legitimate peripheral participation}.
\newblock Cambridge university press.

\bibitem[Lee et~al., 2021]{lee-etal-2021-enhancing}
{ Lee, D.}, { Tian, Z.}, { Xue, L.}, {and} { Zhang, N.~L.} 2021.
\newblock Enhancing content preservation in text style transfer using reverse
  attention and conditional layer normalization.
\newblock In {\em Proceedings of the 59th Annual Meeting of the Association for
  Computational Linguistics and the 11th International Joint Conference on
  Natural Language Processing (Volume 1: Long Papers)}, pp. 93--102, Online.
  Association for Computational Linguistics.

\bibitem[Lee, 2001]{lee2001genres}
{ Lee, D.~Y.} 2001.
\newblock Genres, registers, text types, domains and styles: Clarifying the
  concepts and navigating a path through the bnc jungle.
\newblock {\em Language Learning \& Technology}, 5(3):37--72.

\bibitem[Lee, 2020]{lee-2020-stable}
{ Lee, J.} 2020.
\newblock Stable style transformer: Delete and generate approach with
  encoder-decoder for text style transfer.
\newblock In {\em Proceedings of the 13th International Conference on Natural
  Language Generation}, pp. 195--204, Dublin, Ireland. Association for
  Computational Linguistics.

\bibitem[Lee et~al., 2019]{lee-etal-2019-neural}
{ Lee, J.}, { Xie, Z.}, { Wang, C.}, { Drach, M.}, { Jurafsky, D.}, {and} { Ng,
  A.} 2019.
\newblock Neural text style transfer via denoising and reranking.
\newblock In {\em Proceedings of the Workshop on Methods for Optimizing and
  Evaluating Neural Language Generation}, pp. 74--81, Minneapolis, Minnesota.
  Association for Computational Linguistics.

\bibitem[Leeftink and Spanakis, 2019]{leeftink2019towards}
{ Leeftink, W.} {and} { Spanakis, G.} 2019.
\newblock Towards controlled transformation of sentiment in sentences.
\newblock In {\em International Conference on Agents and Artificial
  Intelligence}.

\bibitem[Lewis et~al., 2020]{lewis-etal-2020-bart}
{ Lewis, M.}, { Liu, Y.}, { Goyal, N.}, { Ghazvininejad, M.}, { Mohamed, A.}, {
  Levy, O.}, { Stoyanov, V.}, {and} { Zettlemoyer, L.} 2020.
\newblock {BART}: Denoising sequence-to-sequence pre-training for natural
  language generation, translation, and comprehension.
\newblock In {\em Proceedings of the 58th Annual Meeting of the Association for
  Computational Linguistics}, pp. 7871--7880, Online. Association for
  Computational Linguistics.

\bibitem[Li et~al., 2019]{li-etal-2019-domain}
{ Li, D.}, { Zhang, Y.}, { Gan, Z.}, { Cheng, Y.}, { Brockett, C.}, { Dolan,
  B.}, {and} { Sun, M.-T.} 2019.
\newblock Domain adaptive text style transfer.
\newblock In {\em Proceedings of the 2019 Conference on Empirical Methods in
  Natural Language Processing and the 9th International Joint Conference on
  Natural Language Processing (EMNLP-IJCNLP)}, pp. 3304--3313, Hong Kong,
  China. Association for Computational Linguistics.

\bibitem[Li et~al., 2016]{li2016new}
{ Li, H.}, { Graesser, A.~C.}, { Conley, M.}, { Cai, Z.}, { Pavlik, P.~I.},
  {and} { Pennebaker, J.~W.} 2016.
\newblock A new measure of text formality: An analysis of discourse of mao
  zedong.
\newblock {\em Discourse Processes}, 53(3):205--232.

\bibitem[Li et~al., 2018]{li-etal-2018-delete}
{ Li, J.}, { Jia, R.}, { He, H.}, {and} { Liang, P.} 2018.
\newblock Delete, retrieve, generate: a simple approach to sentiment and style
  transfer.
\newblock In {\em Proceedings of the 2018 Conference of the North {A}merican
  Chapter of the Association for Computational Linguistics: Human Language
  Technologies, Volume 1 (Long Papers)}, pp. 1865--1874, New Orleans,
  Louisiana. Association for Computational Linguistics.

\bibitem[Li et~al., 2020a]{li-etal-2020-dgst}
{ Li, X.}, { Chen, G.}, { Lin, C.}, {and} { Li, R.} 2020a.
\newblock {DGST}: a dual-generator network for text style transfer.
\newblock In {\em Proceedings of the 2020 Conference on Empirical Methods in
  Natural Language Processing (EMNLP)}, pp. 7131--7136, Online. Association for
  Computational Linguistics.

\bibitem[Li et~al., 2021]{li-etal-2021-text}
{ Li, X.}, { Sun, S.}, {and} { Wang, Y.} 2021.
\newblock Text style transfer: Leveraging a style classifier on entangled
  latent representations.
\newblock In {\em Proceedings of the 6th Workshop on Representation Learning
  for NLP (RepL4NLP-2021)}, pp. 72--82, Online. Association for Computational
  Linguistics.

\bibitem[Li et~al., 2020b]{li2020complementary}
{ Li, Y.}, { Li, C.}, { Zhang, Y.}, { Li, X.}, { Zheng, G.}, { Carin, L.},
  {and} { Gao, J.} 2020b.
\newblock Complementary auxiliary classifiers for label-conditional text
  generation.
\newblock {\em Proceedings of the AAAI Conference on Artificial Intelligence},
  34(05):8303--8310.

\bibitem[Li et~al., 2017]{Li2017}
{ Li, Y.}, { Su, H.}, { Shen, X.}, { Li, W.}, { Cao, Z.}, {and} { Niu, S.}
  2017.
\newblock {D}aily{D}ialog: A manually labelled multi-turn dialogue dataset.
\newblock In {\em Proceedings of the Eighth International Joint Conference on
  Natural Language Processing (Volume 1: Long Papers)}, pp. 986--995, Taipei,
  Taiwan. Asian Federation of Natural Language Processing.

\bibitem[Liao et~al., 2018]{liao-etal-2018-quase}
{ Liao, Y.}, { Bing, L.}, { Li, P.}, { Shi, S.}, { Lam, W.}, {and} { Zhang, T.}
  2018.
\newblock {Q}ua{SE}: Sequence editing under quantifiable guidance.
\newblock In {\em Proceedings of the 2018 Conference on Empirical Methods in
  Natural Language Processing}, pp. 3855--3864, Brussels, Belgium. Association
  for Computational Linguistics.

\bibitem[Lin, 2004]{lin2004rouge}
{ Lin, C.-Y.} 2004.
\newblock Rouge: A package for automatic evaluation of summaries.
\newblock In {\em Text summarization branches out}, pp. 74--81.

\bibitem[Lin et~al., 2020]{lin-etal-2020-learning}
{ Lin, K.}, { Liu, M.-Y.}, { Sun, M.-T.}, {and} { Kautz, J.} 2020.
\newblock Learning to generate multiple style transfer outputs for an input
  sentence.
\newblock In {\em Proceedings of the Fourth Workshop on Neural Generation and
  Translation}, pp. 10--23, Online. Association for Computational Linguistics.

\bibitem[Ling and Klinger, 2016]{ling-sarcasm-irony-2016}
{ Ling, J.} {and} { Klinger, R.} 2016.
\newblock An empirical, quantitative analysis of the differences between
  sarcasm and irony.
\newblock In { Sack, H.}, { Rizzo, G.}, { Steinmetz, N.}, { Mladeni{\'{c}},
  D.}, { Auer, S.}, {and} { Lange, C.}, editors, {\em The Semantic Web}, pp.
  203--216, Cham. Springer International Publishing.

\bibitem[Liu, 2012]{liu2012sentiment}
{ Liu, B.} 2012.
\newblock {\em Sentiment Analysis and Opinion Mining.}
\newblock Morgan \& Claypool.

\bibitem[Liu and Zhang, 2012]{liu2012survey}
{ Liu, B.} {and} { Zhang, L.} 2012.
\newblock A survey of opinion mining and sentiment analysis.
\newblock In {\em Mining text data}, pp. 415--463. Springer.

\bibitem[Liu et~al., 2016]{liu-etal-2016-evaluate}
{ Liu, C.-W.}, { Lowe, R.}, { Serban, I.}, { Noseworthy, M.}, { Charlin, L.},
  {and} { Pineau, J.} 2016.
\newblock How {NOT} to evaluate your dialogue system: An empirical study of
  unsupervised evaluation metrics for dialogue response generation.
\newblock In {\em Proceedings of the 2016 Conference on Empirical Methods in
  Natural Language Processing}, pp. 2122--2132, Austin, Texas. Association for
  Computational Linguistics.

\bibitem[Liu et~al., 2020a]{Liu_Fu_Zhang_Pal_Lv_2020}
{ Liu, D.}, { Fu, J.}, { Zhang, Y.}, { Pal, C.}, {and} { Lv, J.} 2020a.
\newblock Revision in continuous space: Unsupervised text style transfer
  without adversarial learning.
\newblock {\em Proceedings of the AAAI Conference on Artificial Intelligence},
  34(05):8376--8383.

\bibitem[Liu et~al., 2020b]{liu2020revision}
{ Liu, D.}, { Fu, J.}, { Zhang, Y.}, { Pal, C.}, {and} { Lv, J.} 2020b.
\newblock Revision in continuous space: Unsupervised text style transfer
  without adversarial learning.
\newblock {\em Proceedings of the AAAI Conference on Artificial Intelligence},
  34(05):8376--8383.

\bibitem[Logeswaran et~al., 2018]{logeswaran2018content}
{ Logeswaran, L.}, { Lee, H.}, {and} { Bengio, S.} 2018.
\newblock Content preserving text generation with attribute controls.
\newblock In { Bengio, S.}, { Wallach, H.}, { Larochelle, H.}, { Grauman, K.},
  { Cesa-Bianchi, N.}, {and} { Garnett, R.}, editors, {\em Advances in Neural
  Information Processing Systems}, volume~31. Curran Associates, Inc.

\bibitem[Luo et~al., 2019a]{luo-etal-2019-towards}
{ Luo, F.}, { Li, P.}, { Yang, P.}, { Zhou, J.}, { Tan, Y.}, { Chang, B.}, {
  Sui, Z.}, {and} { Sun, X.} 2019a.
\newblock Towards fine-grained text sentiment transfer.
\newblock In {\em Proceedings of the 57th Annual Meeting of the Association for
  Computational Linguistics}, pp. 2013--2022, Florence, Italy. Association for
  Computational Linguistics.

\bibitem[Luo et~al., 2019b]{Luo19DualRL}
{ Luo, F.}, { Li, P.}, { Zhou, J.}, { Yang, P.}, { Chang, B.}, { Sui, Z.},
  {and} { Sun, X.} 2019b.
\newblock A dual reinforcement learning framework for unsupervised text style
  transfer.
\newblock In {\em Proceedings of the 28th International Joint Conference on
  Artificial Intelligence, {IJCAI} 2019}.

\bibitem[Lyu et~al., 2021]{lyu-etal-2021-styleptb}
{ Lyu, Y.}, { Liang, P.~P.}, { Pham, H.}, { Hovy, E.}, { P{\'o}czos, B.}, {
  Salakhutdinov, R.}, {and} { Morency, L.-P.} 2021.
\newblock {S}tyle{PTB}: A compositional benchmark for fine-grained controllable
  text style transfer.
\newblock In {\em Proceedings of the 2021 Conference of the North American
  Chapter of the Association for Computational Linguistics: Human Language
  Technologies}, pp. 2116--2138, Online. Association for Computational
  Linguistics.

\bibitem[Madaan et~al., 2020]{madaan-etal-2020-politeness}
{ Madaan, A.}, { Setlur, A.}, { Parekh, T.}, { Poczos, B.}, { Neubig, G.}, {
  Yang, Y.}, { Salakhutdinov, R.}, { Black, A.~W.}, {and} { Prabhumoye, S.}
  2020.
\newblock Politeness transfer: A tag and generate approach.
\newblock In {\em Proceedings of the 58th Annual Meeting of the Association for
  Computational Linguistics}, pp. 1869--1881, Online. Association for
  Computational Linguistics.

\bibitem[Mai et~al., 2020]{mai-etal-2020-plug}
{ Mai, F.}, { Pappas, N.}, { Montero, I.}, { Smith, N.~A.}, {and} { Henderson,
  J.} 2020.
\newblock Plug and play autoencoders for conditional text generation.
\newblock In {\em Proceedings of the 2020 Conference on Empirical Methods in
  Natural Language Processing (EMNLP)}, pp. 6076--6092, Online. Association for
  Computational Linguistics.

\bibitem[Mairesse and Walker, 2011]{mairesse-walker-2011-controlling}
{ Mairesse, F.} {and} { Walker, M.~A.} 2011.
\newblock Controlling user perceptions of linguistic style: Trainable
  generation of personality traits.
\newblock {\em Computational Linguistics}, 37(3):455--488.

\bibitem[Malmi et~al., 2020]{malmi-etal-2020-unsupervised}
{ Malmi, E.}, { Severyn, A.}, {and} { Rothe, S.} 2020.
\newblock Unsupervised text style transfer with padded masked language models.
\newblock In {\em Proceedings of the 2020 Conference on Empirical Methods in
  Natural Language Processing (EMNLP)}, pp. 8671--8680, Online. Association for
  Computational Linguistics.

\bibitem[Marcheggiani and Titov, 2017]{marcheggiani-titov-2017-encoding}
{ Marcheggiani, D.} {and} { Titov, I.} 2017.
\newblock Encoding sentences with graph convolutional networks for semantic
  role labeling.
\newblock In {\em Proceedings of the 2017 Conference on Empirical Methods in
  Natural Language Processing}, pp. 1506--1515, Copenhagen, Denmark.
  Association for Computational Linguistics.

\bibitem[Martin and Wolfram, 1998]{martin1998sentence}
{ Martin, S.} {and} { Wolfram, W.} 1998.
\newblock The sentence in african-american vernacular english.
\newblock {\em African American English: structure, history, and use}, pp.
  11--36.

\bibitem[Mehrabian, 1996]{mehrabian1996pleasure}
{ Mehrabian, A.} 1996.
\newblock Pleasure-arousal-dominance: A general framework for describing and
  measuring individual differences in temperament.
\newblock {\em Current Psychology}, 14(4):261--292.

\bibitem[Meier, 1995]{meier1995defining}
{ Meier, A.} 1995.
\newblock Defining politeness: Universality in appropriateness.
\newblock {\em Language Sciences}, 17(4):345--356.

\bibitem[Mendenhall, 1887]{mendenhall1887characteristic}
{ Mendenhall, T.~C.} 1887.
\newblock The characteristic curves of composition.
\newblock {\em Science}, 9(214):237--249.

\bibitem[Mendoza-Denton and Iwai, 1993]{mendoza1993}
{ Mendoza-Denton, N.} {and} { Iwai, M.} 1993.
\newblock {They speak more caucasian}: generational differences in the speech
  of japanese-americans.
\newblock {\em Proceedings of the First Annual Symposium About Language and
  Society}, pp. 58--67.

\bibitem[Meyer, 2006]{meyer2000}
{ Meyer, J.~C.} 2006.
\newblock Humor as a double-edged sword: Four functions of humor in
  communication.
\newblock {\em Communication Theory}, 10(3):310--331.

\bibitem[Mir et~al., 2019]{mir-etal-2019-evaluating}
{ Mir, R.}, { Felbo, B.}, { Obradovich, N.}, {and} { Rahwan, I.} 2019.
\newblock Evaluating style transfer for text.
\newblock In {\em Proceedings of the 2019 Conference of the North {A}merican
  Chapter of the Association for Computational Linguistics: Human Language
  Technologies, Volume 1 (Long and Short Papers)}, pp. 495--504, Minneapolis,
  Minnesota. Association for Computational Linguistics.

\bibitem[Mishra et~al., 2019]{mishra-etal-2019-modular}
{ Mishra, A.}, { Tater, T.}, {and} { Sankaranarayanan, K.} 2019.
\newblock A modular architecture for unsupervised sarcasm generation.
\newblock In {\em Proceedings of the 2019 Conference on Empirical Methods in
  Natural Language Processing and the 9th International Joint Conference on
  Natural Language Processing (EMNLP-IJCNLP)}, pp. 6144--6154, Hong Kong,
  China. Association for Computational Linguistics.

\bibitem[Mislove et~al., 2011]{mislove2011understanding}
{ Mislove, A.}, { Lehmann, S.}, { Ahn, Y.-Y.}, { Onnela, J.-P.}, {and} {
  Rosenquist, J.} 2011.
\newblock Understanding the demographics of twitter users.
\newblock {\em Proceedings of the International AAAI Conference on Web and
  Social Media}, 5(1).

\bibitem[Mizumoto et~al., 2011]{mizumoto-etal-2011-mining}
{ Mizumoto, T.}, { Komachi, M.}, { Nagata, M.}, {and} { Matsumoto, Y.} 2011.
\newblock Mining revision log of language learning {SNS} for automated
  {J}apanese error correction of second language learners.
\newblock In {\em Proceedings of 5th International Joint Conference on Natural
  Language Processing}, pp. 147--155, Chiang Mai, Thailand. Asian Federation of
  Natural Language Processing.

\bibitem[Mohammad, 2012]{mohammad_emotional_2012}
{ Mohammad, S.} 2012.
\newblock \#emotional tweets.
\newblock In {\em *{SEM} 2012: The First Joint Conference on Lexical and
  Computational Semantics – Volume 1: Proceedings of the main conference and
  the shared task, and Volume 2: Proceedings of the Sixth International
  Workshop on Semantic Evaluation ({SemEval} 2012)}, pp. 246--255, Montréal,
  Canada. Association for Computational Linguistics.

\bibitem[Mohammad et~al., 2016]{mohammad-etal-2016-metaphor}
{ Mohammad, S.}, { Shutova, E.}, {and} { Turney, P.} 2016.
\newblock Metaphor as a medium for emotion: An empirical study.
\newblock In {\em Proceedings of the Fifth Joint Conference on Lexical and
  Computational Semantics}, pp. 23--33, Berlin, Germany. Association for
  Computational Linguistics.

\bibitem[Mohammad et~al., 2015]{mohammad2015sentiment}
{ Mohammad, S.~M.}, { Zhu, X.}, { Kiritchenko, S.}, {and} { Martin, J.} 2015.
\newblock Sentiment, emotion, purpose, and style in electoral tweets.
\newblock {\em Information Processing \& Management}, 51(4):480--499.

\bibitem[Morreall, 1983]{morreall1983taking}
{ Morreall, J.} 1983.
\newblock {\em Taking Laughter Seriously}.
\newblock Suny Press.

\bibitem[Mueller et~al., 2017]{pmlr-v70-mueller17a}
{ Mueller, J.}, { Gifford, D.}, {and} { Jaakkola, T.} 2017.
\newblock Sequence to better sequence: Continuous revision of combinatorial
  structures.
\newblock In { Precup, D.} {and} { Teh, Y.~W.}, editors, {\em Proceedings of
  the 34th International Conference on Machine Learning}, volume~70 of {\em
  Proceedings of Machine Learning Research}, pp. 2536--2544. PMLR.

\bibitem[Mukherjee, 2005]{mukherjee}
{ Mukherjee, J.} 2005.
\newblock Stylistics.
\newblock {\em Encyclopedia of Linguistics}, pp. 1043--1045.

\bibitem[Myers and Myers, 2010]{myers2010gifts}
{ Myers, I.~B.} {and} { Myers, P.~B.} 2010.
\newblock {\em Gifts differing: Understanding personality type}.
\newblock Nicholas Brealey Publishing.

\bibitem[Nangi et~al., 2021]{nangi-etal-2021-counterfactuals}
{ Nangi, S.~R.}, { Chhaya, N.}, { Khosla, S.}, { Kaushik, N.}, {and} { Nyati,
  H.} 2021.
\newblock Counterfactuals to control latent disentangled text representations
  for style transfer.
\newblock In {\em Proceedings of the 59th Annual Meeting of the Association for
  Computational Linguistics and the 11th International Joint Conference on
  Natural Language Processing (Volume 2: Short Papers)}, pp. 40--48, Online.
  Association for Computational Linguistics.

\bibitem[Newman, 2019]{authenticity-newman}
{ Newman, G.~E.} 2019.
\newblock The psychology of authenticity.
\newblock {\em Review of General Psychology}, 23(1):8--18.

\bibitem[Newman et~al., 2003]{newman2003lying}
{ Newman, M.~L.}, { Pennebaker, J.~W.}, { Berry, D.~S.}, {and} { Richards,
  J.~M.} 2003.
\newblock Lying words: Predicting deception from linguistic styles.
\newblock {\em Personality and social psychology bulletin}, 29(5):665--675.

\bibitem[Nguyen et~al., 2013]{nguyen2013old}
{ Nguyen, D.}, { Gravel, R.}, { Trieschnigg, D.}, {and} { Meder, T.} 2013.
\newblock "how old do you think i am?" a study of language and age in twitter.
\newblock {\em Proceedings of the International AAAI Conference on Web and
  Social Media}, 7(1).

\bibitem[Niculae and Yaneva, 2013]{niculae-yaneva-2013-computational}
{ Niculae, V.} {and} { Yaneva, V.} 2013.
\newblock Computational considerations of comparisons and similes.
\newblock In {\em 51st Annual Meeting of the Association for Computational
  Linguistics Proceedings of the Student Research Workshop}, pp. 89--95, Sofia,
  Bulgaria. Association for Computational Linguistics.

\bibitem[Niu et~al., 2018]{niu-etal-2018-multi-task-formality}
{ Niu, X.}, { Rao, S.}, {and} { Carpuat, M.} 2018.
\newblock Multi-task neural models for translating between styles within and
  across languages.
\newblock In {\em Proceedings of the 27th International Conference on
  Computational Linguistics}, pp. 1008--1021, Santa Fe, New Mexico, USA.
  Association for Computational Linguistics.

\bibitem[Nobata et~al., 2016]{nobata2016abusive}
{ Nobata, C.}, { Tetreault, J.}, { Thomas, A.}, { Mehdad, Y.}, {and} { Chang,
  Y.} 2016.
\newblock Abusive language detection in online user content.
\newblock In {\em Proceedings of the 25th International Conference on World
  Wide Web}, WWW '16,  145–153, Republic and Canton of Geneva, CHE.
  International World Wide Web Conferences Steering Committee.

\bibitem[Oraby et~al., 2018]{oraby-etal-2018-controlling}
{ Oraby, S.}, { Reed, L.}, { Tandon, S.}, { T.S., S.}, { Lukin, S.}, {and} {
  Walker, M.} 2018.
\newblock Controlling personality-based stylistic variation with neural natural
  language generators.
\newblock In {\em Proceedings of the 19th Annual {SIG}dial Meeting on Discourse
  and Dialogue}, pp. 180--190, Melbourne, Australia. Association for
  Computational Linguistics.

\bibitem[Orwell, 1962]{orwell1946politics}
{ Orwell, G.} 1962.
\newblock {\em Inside the Whale and Other Essays}, chapter Politics and the
  English Language, pp. 143--157.
\newblock Penguin Books.

\bibitem[Pang, 2019a]{pang2019daunting}
{ Pang, R.~Y.} 2019a.
\newblock The daunting task of real-world textual style transfer
  auto-evaluation.
\newblock {\em arXiv preprint arXiv:1910.03747}.

\bibitem[Pang, 2019b]{pang-2019-towards}
{ Pang, R.~Y.} 2019b.
\newblock Towards actual (not operational) textual style transfer
  auto-evaluation.
\newblock In {\em Proceedings of the 5th Workshop on Noisy User-generated Text
  (W-NUT 2019)}, pp. 444--445, Hong Kong, China. Association for Computational
  Linguistics.

\bibitem[Pang and Gimpel, 2019]{pang-gimpel-2019-unsupervised}
{ Pang, R.~Y.} {and} { Gimpel, K.} 2019.
\newblock Unsupervised evaluation metrics and learning criteria for
  non-parallel textual transfer.
\newblock In {\em Proceedings of the 3rd Workshop on Neural Generation and
  Translation}, pp. 138--147, Hong Kong. Association for Computational
  Linguistics.

\bibitem[Papineni et~al., 2002]{papineni2002bleu}
{ Papineni, K.}, { Roukos, S.}, { Ward, T.}, {and} { Zhu, W.-J.} 2002.
\newblock Bleu: a method for automatic evaluation of machine translation.
\newblock In {\em Proceedings of the 40th annual meeting of the Association for
  Computational Linguistics}, pp. 311--318.

\bibitem[Parker et~al., 2011]{parker2011english}
{ Parker, R.}, { Graff, D.}, { Kong, J.}, { Chen, K.}, {and} { Maeda, K.} 2011.
\newblock English gigaword fifth edition.

\bibitem[Paul, 1970]{paul-1970-figurative-language}
{ Paul, A.~M.} 1970.
\newblock Figurative language.
\newblock {\em Philosophy {\&} Rhetoric}, 3(4):225--248.

\bibitem[Peled and Reichart, 2017]{peled-reichart-2017-sarcasm}
{ Peled, L.} {and} { Reichart, R.} 2017.
\newblock Sarcasm {SIGN}: Interpreting sarcasm with sentiment based monolingual
  machine translation.
\newblock In {\em Proceedings of the 55th Annual Meeting of the Association for
  Computational Linguistics (Volume 1: Long Papers)}, pp. 1690--1700,
  Vancouver, Canada. Association for Computational Linguistics.

\bibitem[Pennacchiotti and Popescu, 2011]{pennacchiotti2011machine}
{ Pennacchiotti, M.} {and} { Popescu, A.-M.} 2011.
\newblock A machine learning approach to twitter user classification.
\newblock {\em Proceedings of the International AAAI Conference on Web and
  Social Media}, 5(1).

\bibitem[Pennebaker and Stone, 2003]{pennebaker2003words}
{ Pennebaker, J.~W.} {and} { Stone, L.~D.} 2003.
\newblock Words of wisdom: language use over the life span.
\newblock {\em Journal of personality and social psychology}, 85(2):291--301.

\bibitem[Plank and Hovy, 2015]{plank-hovy-2015-personality}
{ Plank, B.} {and} { Hovy, D.} 2015.
\newblock Personality traits on {T}witter{---}or{---}{H}ow to get 1,500
  personality tests in a week.
\newblock In {\em Proceedings of the 6th Workshop on Computational Approaches
  to Subjectivity, Sentiment and Social Media Analysis}, pp. 92--98, Lisboa,
  Portugal. Association for Computational Linguistics.

\bibitem[Plaza-del Arco et~al., 2021]{Plazadelarco2021}
{ Plaza-del Arco, F.~M.}, { Halat, S.}, { Pad\'o, S.}, {and} { Klinger, R.}
  2021.
\newblock Multi-task learning with sentiment, emotion, and target detection to
  recognize hate speech and offensive language.
\newblock In {\em Forum for Information Retrieval Evaluation}, Virtual
  Event/India.

\bibitem[Popu{\c{s}}oi et~al., 2018]{popucsoi2018get}
{ Popu{\c{s}}oi, S.~A.}, { Hav{\^a}rneanu, G.~M.}, {and} { Hav{\^a}rneanu,
  C.~E.} 2018.
\newblock ``{G}et the f\#{*}k out of my way!" {E}xploring the cathartic effect
  of swear words in coping with driving anger.
\newblock {\em Transportation Research Part F: Traffic Psychology and
  Behaviour}, 56:215--226.

\bibitem[Poqu{\'e}russe et~al., 2018]{poquerusse2018alexithymia}
{ Poqu{\'e}russe, J.}, { Pastore, L.}, { Dellantonio, S.}, {and} { Esposito,
  G.} 2018.
\newblock Alexithymia and autism spectrum disorder: a complex relationship.
\newblock {\em Frontiers in psychology}, 9:1196.

\bibitem[Prabhumoye et~al., 2020]{prabhumoye-etal-2020-exploring}
{ Prabhumoye, S.}, { Black, A.~W.}, {and} { Salakhutdinov, R.} 2020.
\newblock Exploring controllable text generation techniques.
\newblock In {\em Proceedings of the 28th International Conference on
  Computational Linguistics}, pp. 1--14, Barcelona, Spain (Online).
  International Committee on Computational Linguistics.

\bibitem[Prabhumoye et~al.,
  2018a]{prabhumoye-etal-2018-multilingual-back-translation}
{ Prabhumoye, S.}, { Tsvetkov, Y.}, { Black, A.~W.}, {and} { Salakhutdinov, R.}
  2018a.
\newblock {Style Transfer Through Multilingual and Feedback-Based
  Back-Translation}.
\newblock {\em arXiv}.

\bibitem[Prabhumoye et~al., 2018b]{prabhumoye-etal-2018-style}
{ Prabhumoye, S.}, { Tsvetkov, Y.}, { Salakhutdinov, R.}, {and} { Black, A.~W.}
  2018b.
\newblock Style transfer through back-translation.
\newblock In {\em Proceedings of the 56th Annual Meeting of the Association for
  Computational Linguistics (Volume 1: Long Papers)}, pp. 866--876, Melbourne,
  Australia. Association for Computational Linguistics.

\bibitem[Preo\c{t}iuc-Pietro et~al., 2016]{preotiuc2016discovering}
{ Preo\c{t}iuc-Pietro, D.}, { Xu, W.}, {and} { Ungar, L.} 2016.
\newblock Discovering user attribute stylistic differences via paraphrasing.
\newblock {\em Proceedings of the AAAI Conference on Artificial Intelligence},
  30(1).

\bibitem[Preo{\c{t}}iuc-Pietro et~al., 2016]{Preotiuc2016}
{ Preo{\c{t}}iuc-Pietro, D.}, { Schwartz, H.~A.}, { Park, G.}, { Eichstaedt,
  J.}, { Kern, M.}, { Ungar, L.}, {and} { Shulman, E.} 2016.
\newblock Modelling valence and arousal in {F}acebook posts.
\newblock In {\em Proceedings of the 7th Workshop on Computational Approaches
  to Subjectivity, Sentiment and Social Media Analysis}, pp. 9--15, San Diego,
  California. Association for Computational Linguistics.

\bibitem[Pryzant et~al., 2020]{pryzant2020automatically}
{ Pryzant, R.}, { Martinez, R.~D.}, { Dass, N.}, { Kurohashi, S.}, { Jurafsky,
  D.}, {and} { Yang, D.} 2020.
\newblock Automatically neutralizing subjective bias in text.
\newblock In {\em Proceedings of the AAAI Conference on Artificial
  Intelligence}, volume~34, pp. 480--489.

\bibitem[Rabinovich et~al., 2017]{rabinovich-etal-2017-personalized}
{ Rabinovich, E.}, { Patel, R.~N.}, { Mirkin, S.}, { Specia, L.}, {and} {
  Wintner, S.} 2017.
\newblock Personalized machine translation: Preserving original author traits.
\newblock In {\em Proceedings of the 15th Conference of the {E}uropean Chapter
  of the Association for Computational Linguistics: Volume 1, Long Papers}, pp.
  1074--1084, Valencia, Spain. Association for Computational Linguistics.

\bibitem[Radford et~al., 2019]{Radford2019}
{ Radford, A.}, { Wu, J.}, { Child, R.}, { Luan, D.}, { Amodei, D.}, {
  Sutskever, I.}, {and} { others} 2019.
\newblock Language models are unsupervised multitask learners.
\newblock {\em Technical Report. OpenAI blog}.

\bibitem[Raffel et~al., 2020]{2020t5}
{ Raffel, C.}, { Shazeer, N.}, { Roberts, A.}, { Lee, K.}, { Narang, S.}, {
  Matena, M.}, { Zhou, Y.}, { Li, W.}, {and} { Liu, P.~J.} 2020.
\newblock Exploring the limits of transfer learning with a unified text-to-text
  transformer.
\newblock {\em Journal of Machine Learning Research}, 21(140):1--67.

\bibitem[Rangel et~al., 2015]{rangel2015overview}
{ Rangel, F.}, { Rosso, P.}, { Potthast, M.}, { Stein, B.}, {and} { Daelemans,
  W.} 2015.
\newblock Overview of the 3rd author profiling task at {PAN} 2015.
\newblock In {\em Conference and Labs of the Evaluation Forum (CLEF)},  2015.
  sn.

\bibitem[Rank, 1980]{rank1980}
{ Rank, H.} 1980.
\newblock Analyzing political rhetoric.
\newblock {\em The English Journal}, 69(9):38--43.

\bibitem[Rao and Tetreault, 2018]{rao-tetreault-2018-dear}
{ Rao, S.} {and} { Tetreault, J.} 2018.
\newblock Dear sir or madam, may {I} introduce the {GYAFC} dataset: Corpus,
  benchmarks and metrics for formality style transfer.
\newblock In {\em Proceedings of the 2018 Conference of the North {A}merican
  Chapter of the Association for Computational Linguistics: Human Language
  Technologies, Volume 1 (Long Papers)}, pp. 129--140, New Orleans, Louisiana.
  Association for Computational Linguistics.

\bibitem[Rashkin et~al., 2019]{rashkin-etal-2019-towards}
{ Rashkin, H.}, { Smith, E.~M.}, { Li, M.}, {and} { Boureau, Y.-L.} 2019.
\newblock Towards empathetic open-domain conversation models: A new benchmark
  and dataset.
\newblock In {\em Proceedings of the 57th Annual Meeting of the Association for
  Computational Linguistics}, pp. 5370--5381, Florence, Italy. Association for
  Computational Linguistics.

\bibitem[Raskin, 1979]{raskin-1979-semantic}
{ Raskin, V.} 1979.
\newblock Semantic mechanisms of humor.
\newblock In {\em Proceedings of the Fifth Annual Meeting of the Berkeley
  Linguistics Society}, volume~5, pp. 325--335.

\bibitem[Razavi et~al., 2010]{razavi-2010-offensive}
{ Razavi, A.~H.}, { Inkpen, D.}, { Uritsky, S.}, {and} { Matwin, S.} 2010.
\newblock Offensive language detection using multi-level classification.
\newblock In {\em Canadian Conference on Artificial Intelligence}, pp. 16--27.
  Springer.

\bibitem[Reddy and Knight, 2016]{reddy2016obfuscating}
{ Reddy, S.} {and} { Knight, K.} 2016.
\newblock Obfuscating gender in social media writing.
\newblock In {\em Proceedings of 2016 EMNLP Workshop on Natural Language
  Processing and Computational Social Science}, pp. 17--26.

\bibitem[Reid and Zhong, 2021]{reid-zhong-2021-lewis}
{ Reid, M.} {and} { Zhong, V.} 2021.
\newblock {LEWIS}: {L}evenshtein editing for unsupervised text style transfer.
\newblock In {\em Findings of the Association for Computational Linguistics:
  ACL-IJCNLP 2021}, pp. 3932--3944, Online. Association for Computational
  Linguistics.

\bibitem[Reif et~al., 2022]{reif2021recipe}
{ Reif, E.}, { Ippolito, D.}, { Yuan, A.}, { Coenen, A.}, { Callison-Burch,
  C.}, {and} { Wei, J.} 2022.
\newblock A recipe for arbitrary text style transfer with large language
  models.
\newblock In {\em Proceedings of the 60th Annual Meeting of the Association for
  Computational Linguistics (Volume 2: Short Papers)}, pp. 837--848, Dublin,
  Ireland. Association for Computational Linguistics.

\bibitem[Reisigl, 2008]{reisigl2008}
{ Reisigl, M.} 2008.
\newblock {\em 11. Rhetoric of political speeches}, pp. 243--270.
\newblock De Gruyter Mouton.

\bibitem[Riley et~al., 2021]{riley-etal-2021-textsettr}
{ Riley, P.}, { Constant, N.}, { Guo, M.}, { Kumar, G.}, { Uthus, D.}, {and} {
  Parekh, Z.} 2021.
\newblock {T}ext{SETTR}: Few-shot text style extraction and tunable targeted
  restyling.
\newblock In {\em Proceedings of the 59th Annual Meeting of the Association for
  Computational Linguistics and the 11th International Joint Conference on
  Natural Language Processing (Volume 1: Long Papers)}, pp. 3786--3800, Online.
  Association for Computational Linguistics.

\bibitem[Riloff et~al., 2013]{riloff-etal-2013-sarcasm}
{ Riloff, E.}, { Qadir, A.}, { Surve, P.}, { De~Silva, L.}, { Gilbert, N.},
  {and} { Huang, R.} 2013.
\newblock Sarcasm as contrast between a positive sentiment and negative
  situation.
\newblock In {\em Proceedings of the 2013 Conference on Empirical Methods in
  Natural Language Processing}, pp. 704--714, Seattle, Washington, USA.
  Association for Computational Linguistics.

\bibitem[Ritchie, 1999]{ritchie-1999-incongruity-resolution}
{ Ritchie, G.} 1999.
\newblock Developing the incongruity-resolution theory.
\newblock In {\em In Proceedings of the AISB Symposium on Creative Language:
  Stories and Humour}, pp. 78--85.

\bibitem[Romanov et~al., 2019]{romanov-etal-2019-adversarial}
{ Romanov, A.}, { Rumshisky, A.}, { Rogers, A.}, {and} { Donahue, D.} 2019.
\newblock Adversarial decomposition of text representation.
\newblock In {\em Proceedings of the 2019 Conference of the North {A}merican
  Chapter of the Association for Computational Linguistics: Human Language
  Technologies, Volume 1 (Long and Short Papers)}, pp. 815--825, Minneapolis,
  Minnesota. Association for Computational Linguistics.

\bibitem[Rosenthal and McKeown, 2011]{rosenthal-mckeown-2011-age}
{ Rosenthal, S.} {and} { McKeown, K.} 2011.
\newblock Age prediction in blogs: A study of style, content, and online
  behavior in pre- and post-social media generations.
\newblock In {\em Proceedings of the 49th Annual Meeting of the Association for
  Computational Linguistics: Human Language Technologies}, pp. 763--772,
  Portland, Oregon, USA. Association for Computational Linguistics.

\bibitem[Rubner et~al., 1998]{rubner1998metric}
{ Rubner, Y.}, { Tomasi, C.}, {and} { Guibas, L.~J.} 1998.
\newblock A metric for distributions with applications to image databases.
\newblock In {\em Sixth International Conference on Computer Vision (IEEE Cat.
  No. 98CH36271)}, pp. 59--66. IEEE.

\bibitem[Rude et~al., 2004]{rude2004language}
{ Rude, S.}, { Gortner, E.-M.}, {and} { Pennebaker, J.} 2004.
\newblock Language use of depressed and depression-vulnerable college students.
\newblock {\em Cognition \& Emotion}, 18(8):1121--1133.

\bibitem[Rutter, 1997]{rutter1997stand}
{ Rutter, J.} 1997.
\newblock {\em Stand-up as Interaction: Performance and Audience in Comedy
  Venues}.
\newblock University of Salford (United Kingdom).

\bibitem[Sarawgi et~al., 2011]{sarawgi2011gender}
{ Sarawgi, R.}, { Gajulapalli, K.}, {and} { Choi, Y.} 2011.
\newblock Gender attribution: tracing stylometric evidence beyond topic and
  genre.
\newblock In {\em Proceedings of the fifteenth conference on computational
  natural language learning}, pp. 78--86.

\bibitem[Scherer, 2005]{scherer2005emotions}
{ Scherer, K.~R.} 2005.
\newblock What are emotions? and how can they be measured?
\newblock {\em Social science information}, 44(4):695--729.

\bibitem[Scherer and Wallbott, 1994]{scherer1994evidence}
{ Scherer, K.~R.} {and} { Wallbott, H.~G.} 1994.
\newblock Evidence for universality and cultural variation of differential
  emotion response patterning.
\newblock {\em Journal of personality and social psychology}, 66(2):310.

\bibitem[Schmidt and Wiegand, 2017]{schmidt-wiegand-2017-survey}
{ Schmidt, A.} {and} { Wiegand, M.} 2017.
\newblock A survey on hate speech detection using natural language processing.
\newblock In {\em Proceedings of the Fifth International Workshop on Natural
  Language Processing for Social Media}, pp. 1--10, Valencia, Spain.
  Association for Computational Linguistics.

\bibitem[Schuff et~al., 2017]{Schuff2017}
{ Schuff, H.}, { Barnes, J.}, { Mohme, J.}, { Pad{\'o}, S.}, {and} { Klinger,
  R.} 2017.
\newblock Annotation, modelling and analysis of fine-grained emotions on a
  stance and sentiment detection corpus.
\newblock In {\em Proceedings of the 8th Workshop on Computational Approaches
  to Subjectivity, Sentiment and Social Media Analysis}, pp. 13--23,
  Copenhagen, Denmark. Association for Computational Linguistics.

\bibitem[Segarra, 2007]{segarra2007become}
{ Segarra, C.} 2007.
\newblock {\em How to Become a True Professional}.
\newblock Xulon Press, Incorporated.

\bibitem[Shang et~al., 2019]{shang-etal-2019-semi}
{ Shang, M.}, { Li, P.}, { Fu, Z.}, { Bing, L.}, { Zhao, D.}, { Shi, S.}, {and}
  { Yan, R.} 2019.
\newblock Semi-supervised text style transfer: Cross projection in latent
  space.
\newblock In {\em Proceedings of the 2019 Conference on Empirical Methods in
  Natural Language Processing and the 9th International Joint Conference on
  Natural Language Processing (EMNLP-IJCNLP)}, pp. 4937--4946, Hong Kong,
  China. Association for Computational Linguistics.

\bibitem[Shapiro, 1986]{shapiro1986language}
{ Shapiro, M.~J.} 1986.
\newblock Language and politics.
\newblock {\em Annual review of applied linguistics}, 7:74--85.

\bibitem[Shardlow and Nawaz, 2019]{shardlow-nawaz-2019-neural}
{ Shardlow, M.} {and} { Nawaz, R.} 2019.
\newblock Neural text simplification of clinical letters with a domain specific
  phrase table.
\newblock In {\em Proceedings of the 57th Annual Meeting of the Association for
  Computational Linguistics}, pp. 380--389, Florence, Italy. Association for
  Computational Linguistics.

\bibitem[Shen et~al., 2017]{shen2017style}
{ Shen, T.}, { Lei, T.}, { Barzilay, R.}, {and} { Jaakkola, T.} 2017.
\newblock Style transfer from non-parallel text by cross-alignment.
\newblock In {\em Advances in neural information processing systems}, pp.
  6830--6841.

\bibitem[Shetty et~al., 2018]{shetty2018a4nt}
{ Shetty, R.}, { Schiele, B.}, {and} { Fritz, M.} 2018.
\newblock A4nt: Author attribute anonymity by adversarial training of neural
  machine translation.
\newblock In {\em 27th {USENIX} Security Symposium ({USENIX} Security 18)}, pp.
  1633--1650, Baltimore, MD. {USENIX} Association.

\bibitem[Shuster et~al., 2019]{shuster2019engaging}
{ Shuster, K.}, { Humeau, S.}, { Hu, H.}, { Bordes, A.}, {and} { Weston, J.}
  2019.
\newblock Engaging image captioning via personality.
\newblock In {\em Proceedings of the IEEE Conference on Computer Vision and
  Pattern Recognition}, pp. 12516--12526.

\bibitem[Singh and Palod, 2018]{singh2018sentiment}
{ Singh, A.} {and} { Palod, R.} 2018.
\newblock Sentiment transfer using seq2seq adversarial autoencoders.
\newblock {\em arXiv preprint arXiv:1804.04003}.

\bibitem[Singh et~al., 2021]{singh-etal-2021-drag}
{ Singh, H.}, { Verma, G.}, { Garimella, A.}, {and} { Srinivasan, B.~V.} 2021.
\newblock {DRAG}: Director-generator language modelling framework for
  non-parallel author stylized rewriting.
\newblock In {\em Proceedings of the 16th Conference of the European Chapter of
  the Association for Computational Linguistics: Main Volume}, pp. 863--873,
  Online. Association for Computational Linguistics.

\bibitem[Smith et~al., 2019]{smith2019zero}
{ Smith, E.~M.}, { Gonzalez{-}Rico, D.}, { Dinan, E.}, {and} { Boureau, Y.}
  2019.
\newblock Zero-shot fine-grained style transfer: Leveraging distributed
  continuous style representations to transfer to unseen styles.
\newblock {\em CoRR}, abs/1911.03914.

\bibitem[Song et~al., 2019]{song-etal-2019-generating}
{ Song, Z.}, { Zheng, X.}, { Liu, L.}, { Xu, M.}, {and} { Huang, X.} 2019.
\newblock Generating responses with a specific emotion in dialog.
\newblock In {\em Proceedings of the 57th Annual Meeting of the Association for
  Computational Linguistics}, pp. 3685--3695, Florence, Italy. Association for
  Computational Linguistics.

\bibitem[Sonnemans and Frijda, 1994]{sonnemans1994structure}
{ Sonnemans, J.} {and} { Frijda, N.~H.} 1994.
\newblock The structure of subjective emotional intensity.
\newblock {\em Cognition \& Emotion}, 8(4):329--350.

\bibitem[Speer et~al., 2017]{speer-conceptnet}
{ Speer, R.}, { Chin, J.}, {and} { Havasi, C.} 2017.
\newblock Conceptnet 5.5: An open multilingual graph of general knowledge.
\newblock In {\em Proceedings of the Thirty-First AAAI Conference on Artificial
  Intelligence}, AAAI'17,  4444–4451. AAAI Press.

\bibitem[Spencer-Bennett, 2018]{spencer2018moral}
{ Spencer-Bennett, J.} 2018.
\newblock {\em Moral talk: Stance and evaluation in political discourse}.
\newblock Routledge.

\bibitem[Stamatatos, 2017]{stamatatos-2017-authorship}
{ Stamatatos, E.} 2017.
\newblock Authorship attribution using text distortion.
\newblock In {\em Proceedings of the 15th Conference of the {E}uropean Chapter
  of the Association for Computational Linguistics: Volume 1, Long Papers}, pp.
  1138--1149, Valencia, Spain. Association for Computational Linguistics.

\bibitem[Stranisci et~al., 2022]{Stranisci2022}
{ Stranisci, M.~A.}, { Frenda, S.}, { Ceccaldi, E.}, { Basile, V.}, { Damiano,
  R.}, {and} { Patti, V.} 2022.
\newblock {APPReddit}: a corpus of reddit posts annotated for appraisal.
\newblock In {\em Proceedings of The 13th Language Resources and Evaluation
  Conference}, Marseille, France. European Language Resources Association.

\bibitem[Su et~al., 2017]{su-etal-2017-rephrasing}
{ Su, H.-P.}, { Huang, Z.-J.}, { Chang, H.-T.}, {and} { Lin, C.-J.} 2017.
\newblock Rephrasing profanity in {C}hinese text.
\newblock In {\em Proceedings of the First Workshop on Abusive Language
  Online}, pp. 18--24, Vancouver, BC, Canada. Association for Computational
  Linguistics.

\bibitem[Sudhakar et~al., 2019]{sudhakar-etal-2019-transforming}
{ Sudhakar, A.}, { Upadhyay, B.}, {and} { Maheswaran, A.} 2019.
\newblock {``}transforming{''} delete, retrieve, generate approach for
  controlled text style transfer.
\newblock In {\em Proceedings of the 2019 Conference on Empirical Methods in
  Natural Language Processing and the 9th International Joint Conference on
  Natural Language Processing (EMNLP-IJCNLP)}, pp. 3269--3279, Hong Kong,
  China. Association for Computational Linguistics.

\bibitem[Sue et~al., 2007]{sue2007racial}
{ Sue, D.~W.}, { Capodilupo, C.~M.}, { Torino, G.~C.}, { Bucceri, J.~M.}, {
  Holder, A.}, { Nadal, K.~L.}, {and} { Esquilin, M.} 2007.
\newblock {Racial microaggressions in everyday life: Implications for clinical
  practice.}
\newblock {\em American psychologist}, 62(4):271--286.

\bibitem[Sulis et~al., 2016]{SULIS2016132}
{ Sulis, E.}, { {Irazú Hernández Farías}, D.}, { Rosso, P.}, { Patti, V.},
  {and} { Ruffo, G.} 2016.
\newblock Figurative messages and affect in twitter: Differences between
  \#irony, \#sarcasm and \#not.
\newblock {\em Knowledge-Based Systems}, 108:132--143.
\newblock New Avenues in Knowledge Bases for Natural Language Processing.

\bibitem[Surya et~al., 2019]{surya-etal-2019-unsupervised}
{ Surya, S.}, { Mishra, A.}, { Laha, A.}, { Jain, P.}, {and} {
  Sankaranarayanan, K.} 2019.
\newblock Unsupervised neural text simplification.
\newblock In {\em Proceedings of the 57th Annual Meeting of the Association for
  Computational Linguistics}, pp. 2058--2068, Florence, Italy. Association for
  Computational Linguistics.

\bibitem[Sydorenko, 2018]{sydorenko2018notion}
{ Sydorenko, I.} 2018.
\newblock The notion of idiostyle in linguistic studies of literary texts.
\newblock {\em Advanced Linguistics}, pp. 15--20.

\bibitem[Syed et~al., 2020]{syed2020adapting}
{ Syed, B.}, { Verma, G.}, { Srinivasan, B.~V.}, { Natarajan, A.}, {and} {
  Varma, V.} 2020.
\newblock Adapting language models for non-parallel author-stylized rewriting.
\newblock In {\em Proceedings of the AAAI Conference on Artificial
  Intelligence}, volume~34, pp. 9008--9015.

\bibitem[Tajfel, 1974]{tajfel1974social}
{ Tajfel, H.} 1974.
\newblock Social identity and intergroup behaviour.
\newblock {\em Social science information}, 13(2):65--93.

\bibitem[Tajiri et~al., 2012]{tajiri-etal-2012-tense}
{ Tajiri, T.}, { Komachi, M.}, {and} { Matsumoto, Y.} 2012.
\newblock Tense and aspect error correction for {ESL} learners using global
  context.
\newblock In {\em Proceedings of the 50th Annual Meeting of the Association for
  Computational Linguistics (Volume 2: Short Papers)}, pp. 198--202, Jeju
  Island, Korea. Association for Computational Linguistics.

\bibitem[Terkourafi, 2008]{terkourafi2008toward}
{ Terkourafi, M.} 2008.
\newblock {\em Toward a unified theory of politeness, impoliteness, and
  rudeness}, pp. 45--76.
\newblock De Gruyter Mouton.

\bibitem[Tian et~al., 2018]{tian2018structured}
{ Tian, Y.}, { Hu, Z.}, {and} { Yu, Z.} 2018.
\newblock Structured content preservation for unsupervised text style transfer.
\newblock {\em arXiv preprint arXiv:1810.06526}.

\bibitem[Tikhonov et~al., 2019]{tikhonov-etal-2019-style}
{ Tikhonov, A.}, { Shibaev, V.}, { Nagaev, A.}, { Nugmanova, A.}, {and} {
  Yamshchikov, I.~P.} 2019.
\newblock Style transfer for texts: Retrain, report errors, compare with
  rewrites.
\newblock In {\em Proceedings of the 2019 Conference on Empirical Methods in
  Natural Language Processing and the 9th International Joint Conference on
  Natural Language Processing (EMNLP-IJCNLP)}, pp. 3936--3945, Hong Kong,
  China. Association for Computational Linguistics.

\bibitem[Tikhonov and Yamshchikov, 2018]{tikhonov2018wrong}
{ Tikhonov, A.} {and} { Yamshchikov, I.~P.} 2018.
\newblock What is wrong with style transfer for texts?
\newblock {\em arXiv preprint arXiv:1808.04365}.

\bibitem[Toshevska and Gievska, 2021]{toshevska2021review}
{ Toshevska, M.} {and} { Gievska, S.} 2021.
\newblock A review of text style transfer using deep learning.
\newblock {\em IEEE Transactions on Artificial Intelligence}.

\bibitem[Tran et~al., 2020]{tran-etal-2020-towards}
{ Tran, M.}, { Zhang, Y.}, {and} { Soleymani, M.} 2020.
\newblock Towards a friendly online community: An unsupervised style transfer
  framework for profanity redaction.
\newblock In {\em Proceedings of the 28th International Conference on
  Computational Linguistics}, pp. 2107--2114, Barcelona, Spain (Online).
  International Committee on Computational Linguistics.

\bibitem[Troiano et~al., 2020]{Troiano2020}
{ Troiano, E.}, { Klinger, R.}, {and} { Pad{\'o}, S.} 2020.
\newblock Lost in back-translation: Emotion preservation in neural machine
  translation.
\newblock In {\em Proceedings of the 28th International Conference on
  Computational Linguistics}, pp. 4340--4354, Barcelona, Spain (Online).
  International Committee on Computational Linguistics.

\bibitem[Troiano et~al., 2022]{Troiano2022}
{ Troiano, E.}, { Oberl\"ander, L.}, { Wegge, M.}, {and} { Klinger, R.} 2022.
\newblock {x-enVENT}: A corpus of event descriptions with experiencer-specific
  emotion and appraisal annotations.
\newblock In {\em Proceedings of The 13th Language Resources and Evaluation
  Conference}, Marseille, France. European Language Resources Association.

\bibitem[Utsumi, 2000]{UTSUMI20001777}
{ Utsumi, A.} 2000.
\newblock Verbal irony as implicit display of ironic environment:
  Distinguishing ironic utterances from nonirony.
\newblock {\em Journal of Pragmatics}, 32(12):1777--1806.

\bibitem[van Aken et~al., 2018]{van-aken-etal-2018-challenges}
{ van Aken, B.}, { Risch, J.}, { Krestel, R.}, {and} { L{\"o}ser, A.} 2018.
\newblock Challenges for toxic comment classification: An in-depth error
  analysis.
\newblock In {\em Proceedings of the 2nd Workshop on Abusive Language Online
  ({ALW}2)}, pp. 33--42, Brussels, Belgium. Association for Computational
  Linguistics.

\bibitem[Vanecek and Dressler, 1975]{vanecek1975bericht}
{ Vanecek, E.} {and} { Dressler, W.} 1975.
\newblock Bericht uber psycholinguistische experimente zur sprechvariation.
\newblock {\em Weiner Linguistische Gazette}, 9:17--38.

\bibitem[Verhoeven et~al., 2016]{verhoeven-etal-2016-twisty}
{ Verhoeven, B.}, { Daelemans, W.}, {and} { Plank, B.} 2016.
\newblock {T}wi{S}ty: A multilingual {T}witter stylometry corpus for gender and
  personality profiling.
\newblock In {\em Proceedings of the Tenth International Conference on Language
  Resources and Evaluation ({LREC}'16)}, pp. 1632--1637, Portoro{\v{z}},
  Slovenia. European Language Resources Association (ELRA).

\bibitem[Voigt et~al., 2018]{voigt-etal-2018-rtgender}
{ Voigt, R.}, { Jurgens, D.}, { Prabhakaran, V.}, { Jurafsky, D.}, {and} {
  Tsvetkov, Y.} 2018.
\newblock {R}t{G}ender: A corpus for studying differential responses to gender.
\newblock In {\em Proceedings of the Eleventh International Conference on
  Language Resources and Evaluation ({LREC} 2018)}, Miyazaki, Japan. European
  Language Resources Association (ELRA).

\bibitem[Wang et~al., 2019a]{wang_2019_latent_edit}
{ Wang, K.}, { Hua, H.}, {and} { Wan, X.} 2019a.
\newblock Controllable unsupervised text attribute transfer via editing
  entangled latent representation.
\newblock In {\em Advances in Neural Information Processing Systems},
  volume~32.

\bibitem[Wang et~al., 2019b]{wang-etal-2019-harnessing}
{ Wang, Y.}, { Wu, Y.}, { Mou, L.}, { Li, Z.}, {and} { Chao, W.} 2019b.
\newblock Harnessing pre-trained neural networks with rules for formality style
  transfer.
\newblock In {\em Proceedings of the 2019 Conference on Empirical Methods in
  Natural Language Processing and the 9th International Joint Conference on
  Natural Language Processing (EMNLP-IJCNLP)}, pp. 3573--3578, Hong Kong,
  China. Association for Computational Linguistics.

\bibitem[Wang et~al., 2020]{wang-etal-2020-formality}
{ Wang, Y.}, { Wu, Y.}, { Mou, L.}, { Li, Z.}, {and} { Chao, W.} 2020.
\newblock Formality style transfer with shared latent space.
\newblock In {\em Proceedings of the 28th International Conference on
  Computational Linguistics}, pp. 2236--2249, Barcelona, Spain (Online).
  International Committee on Computational Linguistics.

\bibitem[Waseem et~al., 2017]{waseem-etal-2017-understanding}
{ Waseem, Z.}, { Davidson, T.}, { Warmsley, D.}, {and} { Weber, I.} 2017.
\newblock Understanding abuse: A typology of abusive language detection
  subtasks.
\newblock In {\em Proceedings of the First Workshop on Abusive Language
  Online}, pp. 78--84, Vancouver, BC, Canada. Association for Computational
  Linguistics.

\bibitem[Waseem and Hovy, 2016]{waseem-hovy-2016-hateful}
{ Waseem, Z.} {and} { Hovy, D.} 2016.
\newblock Hateful symbols or hateful people? predictive features for hate
  speech detection on {T}witter.
\newblock In {\em Proceedings of the {NAACL} Student Research Workshop}, pp.
  88--93, San Diego, California. Association for Computational Linguistics.

\bibitem[Weller et~al., 2020]{weller-etal-2020-humor}
{ Weller, O.}, { Fulda, N.}, {and} { Seppi, K.} 2020.
\newblock Can humor prediction datasets be used for humor generation? humorous
  headline generation via style transfer.
\newblock In {\em Proceedings of the Second Workshop on Figurative Language
  Processing}, pp. 186--191, Online. Association for Computational Linguistics.

\bibitem[Wen et~al., 2020]{wen-etal-2020-decode}
{ Wen, Z.}, { Cao, J.}, { Yang, R.}, {and} { Wang, S.} 2020.
\newblock Decode with template: Content preserving sentiment transfer.
\newblock In {\em Proceedings of the 12th Language Resources and Evaluation
  Conference}, pp. 4671--4679, Marseille, France. European Language Resources
  Association.

\bibitem[West and Horvitz, 2019]{west-2019-satire}
{ West, R.} {and} { Horvitz, E.} 2019.
\newblock Reverse-engineering satire, or “paper on computational humor
  accepted despite making serious advances”.
\newblock {\em Proceedings of the AAAI Conference on Artificial Intelligence},
  33:7265--7272.

\bibitem[Whitehead and Cavedon, 2010]{whitehead-cavedon-2010-generating}
{ Whitehead, S.} {and} { Cavedon, L.} 2010.
\newblock Generating shifting sentiment for a conversational agent.
\newblock In {\em Proceedings of the {NAACL} {HLT} 2010 Workshop on
  Computational Approaches to Analysis and Generation of Emotion in Text}, pp.
  89--97, Los Angeles, CA. Association for Computational Linguistics.

\bibitem[Wiegand et~al., 2021]{wiegand-etal-2021-implicitly-abusive}
{ Wiegand, M.}, { Ruppenhofer, J.}, {and} { Eder, E.} 2021.
\newblock Implicitly abusive language {--} what does it actually look like and
  why are we not getting there?
\newblock In {\em Proceedings of the 2021 Conference of the North American
  Chapter of the Association for Computational Linguistics: Human Language
  Technologies}, pp. 576--587, Online. Association for Computational
  Linguistics.

\bibitem[Wierzbicka, 1988]{wierzbicka1988semantics}
{ Wierzbicka, A.} 1988.
\newblock {\em The semantics of grammar}, volume~18.
\newblock John Benjamins Publishing.

\bibitem[Wieting and Gimpel, 2018]{wieting-gimpel-2018-paranmt}
{ Wieting, J.} {and} { Gimpel, K.} 2018.
\newblock {P}ara{NMT}-50{M}: Pushing the limits of paraphrastic sentence
  embeddings with millions of machine translations.
\newblock In {\em Proceedings of the 56th Annual Meeting of the Association for
  Computational Linguistics (Volume 1: Long Papers)}, pp. 451--462, Melbourne,
  Australia. Association for Computational Linguistics.

\bibitem[Wu et~al., 2019a]{wu-etal-2019-hierarchical-reinforced}
{ Wu, C.}, { Ren, X.}, { Luo, F.}, {and} { Sun, X.} 2019a.
\newblock A hierarchical reinforced sequence operation method for unsupervised
  text style transfer.
\newblock In {\em Proceedings of the 57th Annual Meeting of the Association for
  Computational Linguistics}, pp. 4873--4883, Florence, Italy. Association for
  Computational Linguistics.

\bibitem[Wu et~al., 2017]{wu2017survey}
{ Wu, X.}, { Xu, K.}, {and} { Hall, P.} 2017.
\newblock A survey of image synthesis and editing with generative adversarial
  networks.
\newblock {\em Tsinghua Science and Technology}, 22(6):660--674.

\bibitem[Wu et~al., 2019b]{ijcai2019-maskinfill}
{ Wu, X.}, { Zhang, T.}, { Zang, L.}, { Han, J.}, {and} { Hu, S.} 2019b.
\newblock Mask and infill: Applying masked language model for sentiment
  transfer.
\newblock In {\em Proceedings of the Twenty-Eighth International Joint
  Conference on Artificial Intelligence, (IJCAI-19)}, pp. 5271--5277.
  International Joint Conferences on Artificial Intelligence Organization.

\bibitem[Wubben et~al., 2012]{wubben-etal-2012-sentence}
{ Wubben, S.}, { van~den Bosch, A.}, {and} { Krahmer, E.} 2012.
\newblock Sentence simplification by monolingual machine translation.
\newblock In {\em Proceedings of the 50th Annual Meeting of the Association for
  Computational Linguistics (Volume 1: Long Papers)}, pp. 1015--1024, Jeju
  Island, Korea. Association for Computational Linguistics.

\bibitem[Xu et~al., 2018]{xu-etal-2018-unpaired}
{ Xu, J.}, { Sun, X.}, { Zeng, Q.}, { Zhang, X.}, { Ren, X.}, { Wang, H.},
  {and} { Li, W.} 2018.
\newblock Unpaired sentiment-to-sentiment translation: A cycled reinforcement
  learning approach.
\newblock In {\em Proceedings of the 56th Annual Meeting of the Association for
  Computational Linguistics (Volume 1: Long Papers)}, pp. 979--988, Melbourne,
  Australia. Association for Computational Linguistics.

\bibitem[Xu et~al., 2019a]{xu-etal-2019-alter}
{ Xu, Q.}, { Xu, C.}, {and} { Qu, L.} 2019a.
\newblock {ALTER}: Auxiliary text rewriting tool for natural language
  generation.
\newblock In {\em Proceedings of the 2019 Conference on Empirical Methods in
  Natural Language Processing and the 9th International Joint Conference on
  Natural Language Processing (EMNLP-IJCNLP): System Demonstrations}, pp.
  13--18, Hong Kong, China. Association for Computational Linguistics.

\bibitem[Xu et~al., 2019b]{xu2019formality}
{ Xu, R.}, { Ge, T.}, {and} { Wei, F.} 2019b.
\newblock Formality style transfer with hybrid textual annotations.
\newblock {\em arXiv preprint arXiv:1903.06353}.

\bibitem[Xu et~al., 2016]{xu-etal-2016-optimizing}
{ Xu, W.}, { Napoles, C.}, { Pavlick, E.}, { Chen, Q.}, {and} { Callison-Burch,
  C.} 2016.
\newblock Optimizing statistical machine translation for text simplification.
\newblock {\em Transactions of the Association for Computational Linguistics},
  4:401--415.

\bibitem[Xu et~al., 2012]{xu-etal-2012-paraphrasing}
{ Xu, W.}, { Ritter, A.}, { Dolan, B.}, { Grishman, R.}, {and} { Cherry, C.}
  2012.
\newblock Paraphrasing for style.
\newblock In {\em Proceedings of {COLING} 2012}, pp. 2899--2914, Mumbai, India.
  The COLING 2012 Organizing Committee.

\bibitem[Yamshchikov et~al., 2019]{yamshchikov-etal-2019-decomposing}
{ Yamshchikov, I.~P.}, { Shibaev, V.}, { Nagaev, A.}, { Jost, J.}, {and} {
  Tikhonov, A.} 2019.
\newblock Decomposing textual information for style transfer.
\newblock In {\em Proceedings of the 3rd Workshop on Neural Generation and
  Translation}, pp. 128--137, Hong Kong. Association for Computational
  Linguistics.

\bibitem[Yang and Klein, 2021]{yang-klein-2021-fudge}
{ Yang, K.} {and} { Klein, D.} 2021.
\newblock {FUDGE}: Controlled text generation with future discriminators.
\newblock In {\em Proceedings of the 2021 Conference of the North American
  Chapter of the Association for Computational Linguistics: Human Language
  Technologies}, pp. 3511--3535, Online. Association for Computational
  Linguistics.

\bibitem[Yang et~al., 2019]{yang-etal-2019-specificity}
{ Yang, P.}, { Lin, J.}, { Xu, J.}, { Xie, J.}, { Su, Q.}, {and} { Sun, X.}
  2019.
\newblock Specificity-driven cascading approach for unsupervised sentiment
  modification.
\newblock In {\em Proceedings of the 2019 Conference on Empirical Methods in
  Natural Language Processing and the 9th International Joint Conference on
  Natural Language Processing (EMNLP-IJCNLP)}, pp. 5508--5517, Hong Kong,
  China. Association for Computational Linguistics.

\bibitem[Yang et~al., 2018]{yang2018unsupervised}
{ Yang, Z.}, { Hu, Z.}, { Dyer, C.}, { Xing, E.~P.}, {and} { Berg-Kirkpatrick,
  T.} 2018.
\newblock Unsupervised text style transfer using language models as
  discriminators.
\newblock In {\em Proceedings of the 32nd International Conference on Neural
  Information Processing Systems}, pp. 7298--7309.

\bibitem[Yao and Yu, 2021]{yao-yu-2021-improving}
{ Yao, Z.} {and} { Yu, H.} 2021.
\newblock Improving formality style transfer with context-aware rule injection.
\newblock In {\em Proceedings of the 59th Annual Meeting of the Association for
  Computational Linguistics and the 11th International Joint Conference on
  Natural Language Processing (Volume 1: Long Papers)}, pp. 1561--1570, Online.
  Association for Computational Linguistics.

\bibitem[Yu et~al., 2018]{yu-etal-2018-neural}
{ Yu, Z.}, { Tan, J.}, {and} { Wan, X.} 2018.
\newblock A neural approach to pun generation.
\newblock In {\em Proceedings of the 56th Annual Meeting of the Association for
  Computational Linguistics (Volume 1: Long Papers)}, pp. 1650--1660,
  Melbourne, Australia. Association for Computational Linguistics.

\bibitem[Zhang et~al., 2020a]{zhang-2020-bert-score}
{ Zhang, T.}, { Kishore, V.}, { Wu, F.}, { Weinberger, K.~Q.}, {and} { Artzi,
  Y.} 2020a.
\newblock Bertscore: Evaluating text generation with bert.
\newblock In {\em International Conference on Learning Representations}.

\bibitem[Zhang and Lapata, 2017]{zhang-lapata-2017-sentence}
{ Zhang, X.} {and} { Lapata, M.} 2017.
\newblock Sentence simplification with deep reinforcement learning.
\newblock In {\em Proceedings of the 2017 Conference on Empirical Methods in
  Natural Language Processing}, pp. 584--594, Copenhagen, Denmark. Association
  for Computational Linguistics.

\bibitem[Zhang et~al., 2018a]{zhang-etal-2018-shaped}
{ Zhang, Y.}, { Ding, N.}, {and} { Soricut, R.} 2018a.
\newblock {SHAPED}: Shared-private encoder-decoder for text style adaptation.
\newblock In {\em Proceedings of the 2018 Conference of the North {A}merican
  Chapter of the Association for Computational Linguistics: Human Language
  Technologies, Volume 1 (Long Papers)}, pp. 1528--1538, New Orleans,
  Louisiana. Association for Computational Linguistics.

\bibitem[Zhang et~al., 2020b]{zhang-etal-2020-parallel}
{ Zhang, Y.}, { Ge, T.}, {and} { Sun, X.} 2020b.
\newblock Parallel data augmentation for formality style transfer.
\newblock In {\em Proceedings of the 58th Annual Meeting of the Association for
  Computational Linguistics}, pp. 3221--3228, Online. Association for
  Computational Linguistics.

\bibitem[Zhang et~al., 2018b]{zhang-etal-2018-learning}
{ Zhang, Y.}, { Xu, J.}, { Yang, P.}, {and} { Sun, X.} 2018b.
\newblock Learning sentiment memories for sentiment modification without
  parallel data.
\newblock In {\em Proceedings of the 2018 Conference on Empirical Methods in
  Natural Language Processing}, pp. 1103--1108, Brussels, Belgium. Association
  for Computational Linguistics.

\bibitem[Zhao et~al., 2018a]{zhao-etal-2018-learning}
{ Zhao, J.}, { Zhou, Y.}, { Li, Z.}, { Wang, W.}, {and} { Chang, K.-W.} 2018a.
\newblock Learning gender-neutral word embeddings.
\newblock In {\em Proceedings of the 2018 Conference on Empirical Methods in
  Natural Language Processing}, pp. 4847--4853, Brussels, Belgium. Association
  for Computational Linguistics.

\bibitem[Zhao et~al., 2018b]{zhao-et-al-2018}
{ Zhao, J.~J.}, { Kim, Y.}, { Zhang, K.}, { Rush, A.~M.}, {and} { LeCun, Y.}
  2018b.
\newblock Adversarially regularized autoencoders.
\newblock In { Dy, J.~G.} {and} { Krause, A.}, editors, {\em Proceedings of the
  35th International Conference on Machine Learning, {ICML} 2018,
  Stockholmsm{\"{a}}ssan, Stockholm, Sweden, July 10-15, 2018}, volume~80 of
  {\em Proceedings of Machine Learning Research}, pp. 5897--5906. {PMLR}.

\bibitem[Zhao et~al., 2018c]{zhao-etal-2018-integrating}
{ Zhao, S.}, { Meng, R.}, { He, D.}, { Saptono, A.}, {and} { Parmanto, B.}
  2018c.
\newblock Integrating transformer and paraphrase rules for sentence
  simplification.
\newblock In {\em Proceedings of the 2018 Conference on Empirical Methods in
  Natural Language Processing}, pp. 3164--3173, Brussels, Belgium. Association
  for Computational Linguistics.

\bibitem[Zhou and Wang, 2018]{zhou-wang-2018-mojitalk}
{ Zhou, X.} {and} { Wang, W.~Y.} 2018.
\newblock {M}oji{T}alk: Generating emotional responses at scale.
\newblock In {\em Proceedings of the 56th Annual Meeting of the Association for
  Computational Linguistics (Volume 1: Long Papers)}, pp. 1128--1137,
  Melbourne, Australia. Association for Computational Linguistics.

\bibitem[Zhu et~al., 2019]{zhu2019neural}
{ Zhu, M.}, { Yu, Z.}, {and} { Wan, X.} 2019.
\newblock A neural approach to irony generation.
\newblock {\em arXiv preprint arXiv:1909.06200}.

\bibitem[Zull, 2006]{zull2006key}
{ Zull, J.~E.} 2006.
\newblock Key aspects of how the brain learns.
\newblock {\em New Directions for Adult and Continuing Education}, 110:1--10.

\end{thebibliography}

\label{lastpage}

\end{document}